\documentclass{article}

\usepackage{microtype}
\usepackage{graphicx}
\usepackage{booktabs} 
\usepackage{tabularx}
\usepackage{enumitem}

\usepackage{multirow}
\usepackage{amsmath}
\usepackage{amssymb}
\usepackage{dsfont} 
\usepackage{xcolor}
\usepackage{lipsum}
\usepackage{colortbl}
\usepackage{tcolorbox}
\usepackage{subcaption}
\usepackage{xspace}
\usepackage[colorinlistoftodos, textsize=small]{todonotes}
\usepackage{moresize}
\usepackage{tikz}
\usetikzlibrary{svg.path}

\usepackage{hyperref}
\expandafter\def\expandafter\UrlBreaks\expandafter{\UrlBreaks\do\-}

\newcommand{\name}{LSA}
\newcommand{\dname}{D-LSA}
\newcommand{\nname}{N-LSA}

\newcommand{\qwen}{\textsc{Qwen-3.8-27B}}
\newcommand{\glm}{\textsc{GLM-5.3-Flash}}
\newcommand{\coll}[1]{\textsc{#1}}
\newcommand{\contribution}[1]{\mbox{\textbf{(contribution \##1)}}}

\definecolor{fullattention}{HTML}{963946}
\definecolor{sparseattention}{HTML}{2251A8}
\definecolor{ours}{HTML}{1E8A3C}
\definecolor{blockflat}{HTML}{2251A8}
\definecolor{blocktree}{HTML}{8EAEE6}
\definecolor{lsaflat}{HTML}{1E8A3C}
\definecolor{lsatree}{HTML}{6CC56A}
\definecolor{lsafree}{HTML}{B9E3AE}
\definecolor{swa}{HTML}{D4848F}
\definecolor{nlsa}{HTML}{D95F0E}
\newcommand{\methodbox}[1]{\tikz[baseline=0pt]{\filldraw[fill=#1, draw=black, line width=0.5pt] (0,0) rectangle (1.3ex,1.3ex);}}
\newcommand{\boxfull}{\methodbox{fullattention}}
\newcommand{\boxblockflat}{\methodbox{blockflat}}
\newcommand{\boxblocktree}{\methodbox{blocktree}}
\newcommand{\boxlsaflat}{\methodbox{lsaflat}}
\newcommand{\boxlsatree}{\methodbox{lsatree}}
\newcommand{\boxlsafree}{\methodbox{lsafree}}
\newcommand{\boxswa}{\methodbox{swa}}
\newcommand{\boxnlsa}{\methodbox{nlsa}}
\definecolor{query}{HTML}{950607}
\definecolor{querylight}{HTML}{D0544C}
\definecolor{centroid}{HTML}{006D2C}
\definecolor{algogreen}{HTML}{EDF8E9}
\definecolor{algogrey}{HTML}{F2F2F2}
\makeatletter
\newcommand{\algophase}[1]{\item[]\hspace{-\ALC@tlm}\makebox[\linewidth][c]{\rule[-0.8ex]{0pt}{3.0ex}\leaders\hrule height 0.55ex depth -0.48ex\hfill\enspace\textbf{#1}\enspace\leaders\hrule height 0.55ex depth -0.48ex\hfill}}
\AtBeginDocument{\renewcommand{\algorithmiccomment}[1]{\hfill$\triangleright$~#1}}
\newcommand{\algolinelabel}[1]{\protected@edef\@currentlabel{\arabic{ALC@line}}\label{#1}}
\newcommand{\algoreset}{\let\ALC@lno\relax\let\ALC@it\relax\let\ALC@com\relax\let\REQUIRE\relax\let\ENSURE\relax\let\STATE\relax\let\COMMENT\relax\let\INPUT\relax\let\OUTPUT\relax\let\SET\relax\let\AND\relax\let\OR\relax\let\IF\relax\let\SHORTIF\relax\let\ELSE\relax\let\ELSIF\relax\let\FOR\relax\let\FORALL\relax\let\SHORTFORALL\relax\let\WHILE\relax\let\LOOP\relax\let\FUNCTION\relax\let\MAIN\relax\let\REPEAT\relax\let\UNTIL\relax\let\ENDIF\relax\let\ENDFOR\relax\let\ENDWHILE\relax\let\ENDLOOP\relax\let\ENDFUNCTION\relax\let\ENDMAIN\relax\let\ALC@g\relax\let\endALC@g\relax\let\ALC@func\relax\let\endALC@func\relax\let\ALC@main\relax\let\endALC@main\relax\let\ALC@if\relax\let\endALC@if\relax\let\ALC@for\relax\let\endALC@for\relax\let\ALC@whl\relax\let\endALC@whl\relax\let\ALC@loop\relax\let\endALC@loop\relax\let\ALC@rpt\relax\let\endALC@rpt\relax}
\newcounter{algoboxline}
\newenvironment{algobox}[2]{\setcounter{algoboxline}{\value{ALC@line}}\item[]\begin{tcolorbox}[colback=#1,colframe=#1,boxrule=0pt,arc=2pt,left=4pt,right=4pt,top=3pt,bottom=3pt,boxsep=0pt,before skip=3pt,after skip=3pt]\textit{#2}\par\algoreset\begin{algorithmic}[1]\setcounter{ALC@line}{\value{algoboxline}}}{\setcounter{algoboxline}{\value{ALC@line}}\end{algorithmic}\end{tcolorbox}\setcounter{ALC@line}{\value{algoboxline}}}
\newenvironment{algoboxwide}[2]{\setcounter{algoboxline}{\value{ALC@line}}\item[]\ifvmode\ifdim\prevdepth>-1000pt\ifdim\prevdepth<\dp\strutbox\vskip\dimexpr\dp\strutbox-\prevdepth\relax\fi\fi\fi\begin{tcolorbox}[enlarge left by=-1.5em,width=\linewidth+1.5em,colback=#1,colframe=#1,boxrule=0pt,arc=2pt,left=4pt,right=4pt,top=3pt,bottom=3pt,boxsep=0pt,before skip=2pt,after skip=0pt]\textit{#2}\par\algoreset\begin{algorithmic}[1]\setcounter{ALC@line}{\value{algoboxline}}}{\setcounter{algoboxline}{\value{ALC@line}}\end{algorithmic}\end{tcolorbox}\prevdepth=0pt\setcounter{ALC@line}{\value{algoboxline}}}
\makeatother
\newcommand{\querymark}{\tikz[baseline=-0.5ex]{\filldraw[fill=querylight, draw=query, line width=0.5pt] (0,0) circle (0.5ex);}}
\newcommand{\centroidmark}{\tikz[baseline=0pt]{\filldraw[fill=centroid, draw=centroid, line width=0.6pt, line join=round] (0,1.1ex) -- (0.6ex,0) -- (-0.6ex,0) -- cycle;}}

\newcommand{\separator}{\color{black!30}\vrule width 0.5pt}
\definecolor{absent}{gray}{0.92}

\newif\ifcircledsymbols
\newcommand{\svgicon}[2]{\tikz[baseline={([yshift=-0.7ex]current bounding box.center)}]{\pgfmathsetmacro{\iconscale}{#1/512}\edef\iconfill{\noexpand\fill[even odd rule, scale=\iconscale, yscale=-1] svg {#2};}\iconfill}}
\newcommand{\iconcheck}{M374 145.7c-10.7-7.8-25.7-5.4-33.5 5.3L221.1 315.2 169 263.1c-9.4-9.4-24.6-9.4-33.9 0s-9.4 24.6 0 33.9l72 72c5 5 11.8 7.5 18.8 7s13.4-4.1 17.5-9.8L379.3 179.2c7.8-10.7 5.4-25.7-5.3-33.5z}
\newcommand{\iconcross}{M167 167c9.4-9.4 24.6-9.4 33.9 0l55 55 55-55c9.4-9.4 24.6-9.4 33.9 0s9.4 24.6 0 33.9l-55 55 55 55c9.4 9.4 9.4 24.6 0 33.9s-24.6 9.4-33.9 0l-55-55-55 55c-9.4 9.4-24.6 9.4-33.9 0s-9.4-24.6 0-33.9l55-55-55-55c-9.4-9.4-9.4-24.6 0-33.9z}
\newcommand{\iconhalf}{M448 256c0-106-86-192-192-192l0 384c106 0 192-86 192-192zM0 256a256 256 0 1 1 512 0 256 256 0 1 1 -512 0z}
\newcommand{\iconcheckcircle}{M256 512a256 256 0 1 1 0-512 256 256 0 1 1 0 512z}
\newcommand{\iconcrosscircle}{M256 512a256 256 0 1 0 0-512 256 256 0 1 0 0 512z}
\newcommand{\supported}{\ifcircledsymbols\svgicon{1.6ex}{\iconcheckcircle\iconcheck}\else\svgicon{2.6ex}{\iconcheck}\fi}
\newcommand{\partlysupported}{\svgicon{1.6ex}{\iconhalf}}
\newcommand{\unsupported}{\ifcircledsymbols\svgicon{1.6ex}{\iconcrosscircle\iconcross}\else\svgicon{3ex}{\iconcross}\fi}
\newcommand{\nodeglyph}[1]{\tikz[baseline={([yshift=-0.7ex]current bounding box.center)}, x=1ex, y=1ex, line width=0.6pt]{#1}}
\newcommand{\flatsearch}{\nodeglyph{\draw (0,0) -- (1.8,0); \foreach \x in {0,0.9,1.8} \fill (\x,0) circle (0.34);}}
\newcommand{\fixedsearch}{\nodeglyph{\draw (0.9,1.3) -- (0,0) (0.9,1.3) -- (0.9,0) (0.9,1.3) -- (1.8,0); \foreach \x in {0,0.9,1.8} \fill (\x,0) circle (0.34); \fill (0.9,1.3) circle (0.34);}}
\newcommand{\treesearch}{\nodeglyph{\draw (0.9,1.4) -- (0,0.7) (0.9,1.4) -- (0.9,0.7) (0.9,1.4) -- (1.8,0.7) (0.9,0.7) -- (0.45,0) (0.9,0.7) -- (1.35,0); \foreach \p in {(0.9,1.4),(0,0.7),(0.9,0.7),(1.8,0.7),(0.45,0),(1.35,0)} \fill \p circle (0.3);}}
\definecolor{ink}{HTML}{1A1A1A}
\makeatletter
\newcommand{\stepbadge}[4]{{%
  \edef\badgesize{\fpeval{#2*\f@size}pt}%
  \edef\badgedrop{\fpeval{-0.27*\f@size}pt}%
  \edef\badgefont{\fpeval{#3*\f@size}}%
  \tikz[baseline={([yshift=\badgedrop]badge.center)}]\node[#1, fill=ink, text=white, inner sep=0pt, minimum size=\badgesize,
    font=\sffamily\bfseries\fontsize{\badgefont}{\badgefont}\selectfont](badge){#4};}}
\makeatother
\newcommand{\keystep}[1]{\stepbadge{circle}{1.0}{0.62}{#1}}
\newcommand{\querystep}[1]{\stepbadge{rectangle, rounded corners=0.16em}{0.92}{0.56}{#1}}
\newcommand{\bigO}[1]{\mathcal{O}(#1)}

\usepackage[accepted]{preprint}

\newcommand{\icongithub}{M216.5 362.5c-66-8-112.5-55.5-112.5-117 0-25 9-52 24-70-6.5-16.5-5.5-51.5 2-66 20-2.5 47 8 63 22.5 19-6 39-9 63.5-9s44.5 3 62.5 8.5c15.5-14 43-24.5 63-22 7 13.5 8 48.5 1.5 65.5 16 19 24.5 44.5 24.5 70.5 0 61.5-46.5 108-113.5 116.5 17 11 28.5 35 28.5 62.5l0 52C323 491.5 335.5 500 350.5 494 441 459.5 512 369 512 257 512 115.5 397 0 255.5 0S0 115.5 0 257c0 111 70.5 203 165.5 237.5 13.5 5 26.5-4 26.5-17.5l0-40c-7 3-16 5-24 5-33 0-52.5-18-66.5-51.5-5.5-13.5-11.5-21.5-23-23-6-.5-8-3-8-6 0-6 10-10.5 20-10.5 14.5 0 27 9 40 27.5 10 14.5 20.5 21 33 21s20.5-4.5 32-16c8.5-8.5 15-16 21-21z}
\newcommand{\githubrepo}[1]{\svgicon{1.8ex}{\icongithub}\,\href{https://github.com/#1}{github.com/#1}}

\papertitlerunning{Latent Sparse Attention: Towards Subquadratic Ultra-Long Context Attention}

\begin{document}

\twocolumn[
\papertitle{Latent Sparse Attention: Towards Subquadratic\texorpdfstring{\\}{ }Ultra-Long Context Attention}

\papersetsymbol{core}{*}
\papersetsymbol{sa}{$^{\dagger}$}

\begin{paperauthorlist}
\paperauthor{Ales Kubicek}{core,pf}
\paperauthor{Andrea Caciolai}{core,pf}
\paperauthor{Liangzu Peng}{core,pf}
\paperauthor{Sukrit Sumant}{pf}
\paperauthor{Benjamin Hoover}{pf}
\paperauthor{Fraser Greenlee}{pf}
\paperauthor{Ruairidh M. Battleday}{pf}
\paperauthor{Radim Urban}{eth}
\paperauthor{Torsten Hoefler}{eth}
\paperauthor{James C.R. Whittington}{sa,pf}
\end{paperauthorlist}

\paperaffiliation{pf}{Prefrontal AI, London, United Kingdom}
\paperaffiliation{eth}{ETH Zurich, Zurich, Switzerland}

\papercorrespondingauthor{Ales Kubicek}{ales@prefrontal.ai}
\papercorrespondingauthor{James C.R. Whittington}{james@prefrontal.ai}

\paperkeywords{Sparse Attention, Long Context, Machine Learning Systems}

\vskip 0.25in

\begin{abstract}
Agentic workflows are increasingly shifting towards long-horizon tasks, such as open-ended scientific work, where models benefit from long context windows.
However, at ultra-long context (above 1M tokens), full attention is prohibitively expensive. A promising approach is sparse attention, in which a dedicated indexer selects a subset of keys for the attention. Yet current methods of indexing either still have quadratic cost in context size or lose accuracy at ultra-long context scale.
We address this with Latent Sparse Attention (\name{}), an efficient attention method that maintains accuracy at ultra-long context. In \name{}, keys are hierarchically clustered in a learned latent space, and queries score the hierarchy of clusters instead of scoring every key, making prefill, decode and training faster at long context.
We propose two variants, Distilled-\name{} (\dname{}), which can be applied to pre-trained models via distillation, and Native-\name{} (\nname{}), which is trained natively on the next-token loss alone.
On \glm{} (\qwen{}) at 10M tokens, \dname{} speeds up one attention layer by $22.7\times$ ($72.4\times$) in prefill and $4.6\times$ ($42.5\times$) in decode over its original sparse attention (full attention), while keeping similar accuracy across a variety of tasks.
\nname{} matches the accuracy of full attention with a training speedup of $1.6\times$ per layer at 128k tokens.
\name{} offers practical and efficient attention for models to benefit from ultra-long context.
\end{abstract}
]

\printAffiliationsAndNotice{\textsuperscript{*}Core contributors. $^{\dagger}$ Senior author. }

\section{Introduction}

\begin{figure}[!t]
\centering
\includegraphics[width=\columnwidth]{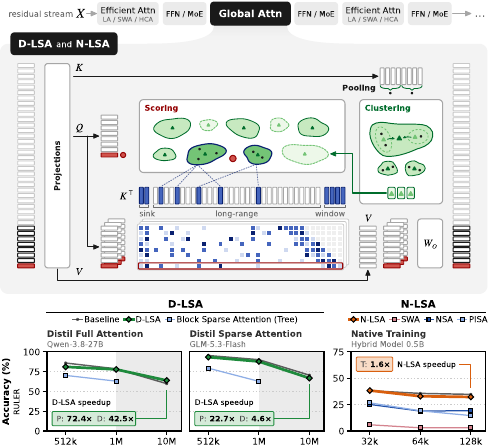}
\caption{\textbf{\dname{} and \nname{} overview.} Top: one attention layer. The keys are pooled and clustered into buckets. A query (\querymark) scores the bucket centroids (\centroidmark) and attends only to the selected long-range keys (plus sink and window). Bottom: accuracy and performance (P: prefill, D: decode, T: training) for both \dname{} and \nname{}. \dname{} is shown with flat search. Its subquadratic variant, \dname{} (Tree), follows a similar trend (Figure~\ref{fig:accuracy}).}
\label{fig:overview}
\end{figure}

Large language models have moved from answering single requests to agentic workflows in long-horizon tasks, such as software engineering and open-ended scientific research~\citep{jimenez2024swebench,lu2026towards}.
These tasks keep growing in length~\citep{kwa2025measuring}, and models with longer context windows tend to score higher on them~\citep{zhu2026edgebench}.
In line with this trend, recent models extend their context windows to 1M tokens and beyond~\citep{gemini2024gemini,minimax2025minimax01,deepseekai2026deepseekv4,qwen2026design}.

At this context size, attention dominates the cost, as its work grows with context~\citep{vaswani2017attention,dao2022flashattention}.
An established technique to reduce such costs is sparse attention~\citep{hoefler2021sparsity}, in which each query attends to only a subset of the keys, chosen by a fixed pattern~\citep{child2019generating,beltagy2020longformer,zaheer2020big}, by their content~\citep{kitaev2020reformer,roy2021efficient}, or by a dedicated indexer~\citep{yuan2025native,deepseekai2025deepseekv32}.
Recent frontier models are increasingly adopting sparse attention, in which keys are selected using a dedicated indexer~\citep{deepseekai2025deepseekv32,qwen2026design}.

However, current indexers still score every preceding key, or block of keys, for each query. This remains expensive (quadratic in prefill, linear in decode), and becomes a bottleneck at long context~\citep{zan2026longcat,xu2026hisa}. For example, at 10M tokens, 98\% of the prefill time of a sparse attention layer is spent on indexing (Figure~\ref{fig:motivation}).
The cost of sparse attention at ultra-long context is therefore dominated by the cost of finding the keys.

To address this, we introduce Latent Sparse Attention (\name{}), which clusters keys in a learned latent space and lets each query search the clusters instead of scoring every key. This can be done using only one level of clustering (\emph{Flat}), or with a hierarchy of clusters (\emph{Tree}), which makes the prefill subquadratic.
Furthermore, we propose two trainable designs: \dname{}, which adds \name{} to pre-trained models and only trains small projection matrices per layer \contribution{1}, and \nname{}, which trains \name{} natively with the Next-Token Prediction (NTP) loss alone \contribution{2}.
We evaluate state-of-the-art models (\qwen{} and \glm{}) beyond their advertised context, up to 10M tokens \contribution{3}, on RULER~\citep{hsieh2024ruler}, which is an established long-context benchmark, and PaperTrail, which we developed to test long-context scientific understanding by finding and connecting facts across unseen scientific papers \contribution{4}.
Finally, we integrate \name{} into SGLang~\citep{zheng2024sglang}\footnote{\githubrepo{Prefrontal-AI/sglang}} with context parallelism, provide GPU kernels for H200 and B300, and release the code\footnote{\githubrepo{Prefrontal-AI/latent-sparse-attention}} with all training and inference details \contribution{5}.
Figure~\ref{fig:overview} outlines our contributions.

\section{Sparse Attention}
\label{sec:background}

In full attention, every query attends to every preceding key.
Prefilling a context of $T$ tokens therefore costs $\bigO{T^2}$, and decoding one token costs $\bigO{T}$. In sparse attention, the work is split between an indexer and the attention. For each query, the indexer selects which keys the attention attends to\footnote{In this view, full attention corresponds to an indexer that returns the indices of all preceding keys.}. With a fixed budget of $k$ keys, the attention then costs $\bigO{Tk}$ to prefill and $\bigO{k}$ per decoded token. If the indexer selects the most relevant keys, this is sufficient to stay close to the accuracy of full attention (Appendix~\ref{app:sparsity}). 

\begin{figure}[t]
\centering
\includegraphics[width=\columnwidth]{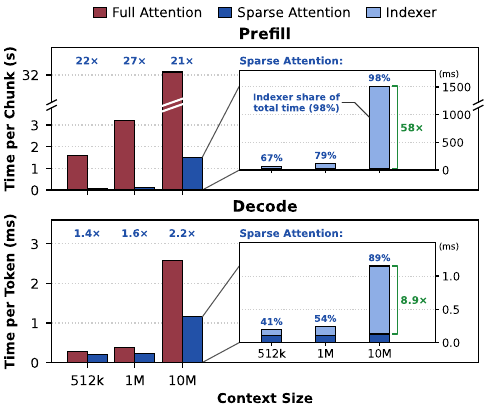}
\caption{\textbf{Traditional sparse attention at ultra-long context.} One global attention layer of \glm{} (512k to 10M tokens) on one H200, prefilling the last chunk of 32,768 queries and decoding one token. Speedup of sparse attention over full attention (\textcolor{sparseattention}{\textbf{blue}}), share of the indexer (\textcolor{sparseattention}{\textbf{\%}}), and potential speedup over sparse attention if the indexer were `free' (\textcolor{ours}{\textbf{green}}).}
\label{fig:motivation}
\end{figure}

However, the indexer has a cost. Since it must score all preceding keys, it typically remains $\bigO{T^2}$ in prefill and $\bigO{T}$ in decode, though with reduced constant factors. This makes ultra-long context still prohibitively expensive. For example, at 10M tokens the indexer takes 98\% of the prefill time of traditional sparse attention (Figure~\ref{fig:motivation}), and a `free' indexer would make sparse attention a further $58\times$ faster in prefill and $8.9\times$ in decode.

\subsection{Design Choices}

Sparse attention methods can be categorised by their indexer designs and how they are trained. 

\textbf{Token Sparse Attention.}
For each query, a lightweight indexer (fewer heads, smaller dimension, lower precision) scores every preceding key and selects which to attend to.

\textbf{Block Sparse Attention.}
Consecutive keys are grouped into blocks, and each query scores blocks instead of every key.
All keys of the selected blocks are attended to.

\textbf{Semantic Sparse Attention.}
Keys are clustered by similarity rather than position, and each query scores one centroid per cluster.
All keys of the selected clusters are attended to.

For the last two categories, a query can search the groups in three ways. \emph{Flat} search scores every group directly, which is $\bigO{T}$ work per query. \emph{Fixed} search adds a shallow hierarchy and refines from coarse to fine groups, which is also $\bigO{T}$ per query. \emph{Tree} search descends a deeper hierarchy and expands only the groups it keeps at each level, which is $\bigO{\log T}$ per query.

An indexer can be training-free, distilled from a pre-trained model~\citep{schmidhuber1992history}, or trained natively on the NTP loss alone.

\subsection{Related Work}
\label{sec:related-work}

\begin{table}[t]
\centering
\caption{\textbf{Sparse attention methods compared.}
Rows are grouped by the unit the indexer scores: single keys, blocks of consecutive keys, or clusters of similar keys.
Design column: compression of the sequence the indexer scans by compressing consecutive keys (C.~Seq.\@), or clustering keys by content (C.~Sem.\@), and its search: flat (\flatsearch), a fixed number of levels (\fixedsearch), or levels efficiently growing as $\log T$ (\treesearch).
\textbf{Evaluation}: the largest context at which accuracy (Acc.\@) and performance (Perf.\@) are reported.
\supported~supported, \partlysupported~partial support, \unsupported~not supported.}
\label{tab:related-work}
\vskip 0.15in
\newcommand{\slot}[2][8pt]{\makebox[#1]{#2}}
\setlength{\tabcolsep}{0pt}
\ssmall
\begin{tabular*}{\columnwidth}{@{\extracolsep{\fill}}lcccc@{}}
\toprule
& \textbf{Training} & \textbf{Speedup} & \textbf{Design} & \textbf{Evaluation} \\
\cmidrule(lr){2-2} \cmidrule(lr){3-3} \cmidrule(lr){4-4} \cmidrule(lr){5-5}
\textbf{Name} & \slot{\rotatebox{90}{\textbf{Free}}}\slot{\rotatebox{90}{\textbf{Distilled}}}\slot{\rotatebox{90}{\textbf{Native}}} & \slot[12pt]{\rotatebox{90}{\textbf{Prefill}}}\slot[12pt]{\rotatebox{90}{\textbf{Decode}}} & \slot{\rotatebox{90}{\textbf{C. Seq.}}}\slot{\rotatebox{90}{\textbf{C. Sem.}}}\slot[10pt]{\rotatebox{90}{\textbf{Search}}} & \slot[16pt]{\rotatebox{90}{\textbf{Acc.}}}\slot[16pt]{\rotatebox{90}{\textbf{Perf.}}} \\
\midrule
\multicolumn{5}{@{}l}{\textit{Token Sparse Attention}} \\
\addlinespace[1.5pt]
DSA~\citep{deepseekai2025deepseekv32} & \slot{\unsupported}\slot{\supported}\slot{\unsupported} & \slot[12pt]{\supported}\slot[12pt]{\supported} & \slot{\unsupported}\slot{\unsupported}\slot[10pt]{\flatsearch} & \slot[16pt]{128k}\slot[16pt]{128k} \\
MiniMax SA~\citep{lai2026minimax} & \slot{\unsupported}\slot{\supported}\slot{\unsupported} & \slot[12pt]{\supported}\slot[12pt]{\supported} & \slot{\unsupported}\slot{\unsupported}\slot[10pt]{\flatsearch} & \slot[16pt]{128k}\slot[16pt]{1M} \\
\midrule
\multicolumn{5}{@{}l}{\textit{Block Sparse Attention}} \\
\addlinespace[1.5pt]
CSA~\citep{deepseekai2026deepseekv4} & \slot{\unsupported}\slot{\supported}\slot{\unsupported} & \slot[12pt]{\supported}\slot[12pt]{\supported} & \slot{\supported}\slot{\unsupported}\slot[10pt]{\flatsearch} & \slot[16pt]{1M}\slot[16pt]{1M} \\
QSA~\citep{qwen2026design} & \slot{\unsupported}\slot{\supported}\slot{\unsupported} & \slot[12pt]{\supported}\slot[12pt]{\supported} & \slot{\supported}\slot{\unsupported}\slot[10pt]{\flatsearch} & \slot[16pt]{1M}\slot[16pt]{1M} \\
\glm{} SA~\citep{zai2026glm53flash} & \slot{\unsupported}\slot{\supported}\slot{\unsupported} & \slot[12pt]{\supported}\slot[12pt]{\supported} & \slot{\supported}\slot{\unsupported}\slot[10pt]{\flatsearch} & \slot[16pt]{1M}\slot[16pt]{1M} \\
NSA~\citep{yuan2025native} & \slot{\unsupported}\slot{\unsupported}\slot{\supported} & \slot[12pt]{\supported}\slot[12pt]{\supported} & \slot{\supported}\slot{\unsupported}\slot[10pt]{\flatsearch} & \slot[16pt]{64k}\slot[16pt]{64k} \\
HSA-UltraLong~\citep{hu2026every} & \slot{\unsupported}\slot{\unsupported}\slot{\supported} & \slot[12pt]{\supported}\slot[12pt]{\supported} & \slot{\supported}\slot{\unsupported}\slot[10pt]{\flatsearch} & \slot[16pt]{16M}\slot[16pt]{256k} \\
MoBA~\citep{lu2025moba} & \slot{\unsupported}\slot{\unsupported}\slot{\supported} & \slot[12pt]{\supported}\slot[12pt]{\supported} & \slot{\supported}\slot{\unsupported}\slot[10pt]{\flatsearch} & \slot[16pt]{1M}\slot[16pt]{10M} \\
SeerAttention~\citep{gao2025seerattention,gao2026sparse} & \slot{\unsupported}\slot{\supported}\slot{\unsupported} & \slot[12pt]{\supported}\slot[12pt]{\supported} & \slot{\supported}\slot{\unsupported}\slot[10pt]{\flatsearch} & \slot[16pt]{128k}\slot[16pt]{128k} \\
LongCat LSA~\citep{zan2026longcat} & \slot{\partlysupported}\slot{\supported}\slot{\unsupported} & \slot[12pt]{\supported}\slot[12pt]{\supported} & \slot{\supported}\slot{\unsupported}\slot[10pt]{\fixedsearch} & \slot[16pt]{1M}\slot[16pt]{1M} \\
HISA~\citep{xu2026hisa} & \slot{\supported}\slot{\unsupported}\slot{\unsupported} & \slot[12pt]{\supported}\slot[12pt]{\supported} & \slot{\supported}\slot{\unsupported}\slot[10pt]{\fixedsearch} & \slot[16pt]{128k}\slot[16pt]{128k} \\
HiP~\citep{lee2025training} & \slot{\supported}\slot{\unsupported}\slot{\unsupported} & \slot[12pt]{\supported}\slot[12pt]{\supported} & \slot{\supported}\slot{\unsupported}\slot[10pt]{\treesearch} & \slot[16pt]{128k}\slot[16pt]{512k} \\
PISA~\citep{tang2026block} & \slot{\unsupported}\slot{\unsupported}\slot{\supported} & \slot[12pt]{\supported}\slot[12pt]{\supported} & \slot{\supported}\slot{\unsupported}\slot[10pt]{\treesearch} & \slot[16pt]{16k}\slot[16pt]{256k} \\
\midrule
\multicolumn{5}{@{}l}{\textit{Semantic Sparse Attention}} \\
\addlinespace[1.5pt]
ClusterKV~\citep{liu2025clusterkv} & \slot{\supported}\slot{\unsupported}\slot{\unsupported} & \slot[12pt]{\unsupported}\slot[12pt]{\supported} & \slot{\unsupported}\slot{\supported}\slot[10pt]{\flatsearch} & \slot[16pt]{32k}\slot[16pt]{32k} \\
RetroInfer~\citep{chen2026retroinfer} & \slot{\supported}\slot{\unsupported}\slot{\unsupported} & \slot[12pt]{\unsupported}\slot[12pt]{\supported} & \slot{\unsupported}\slot{\supported}\slot[10pt]{\flatsearch} & \slot[16pt]{1M}\slot[16pt]{1M} \\
Squeezed Attention~\citep{hooper2025squeezed} & \slot{\supported}\slot{\unsupported}\slot{\unsupported} & \slot[12pt]{\partlysupported}\slot[12pt]{\supported} & \slot{\unsupported}\slot{\supported}\slot[10pt]{\fixedsearch} & \slot[16pt]{32k}\slot[16pt]{512k} \\
Multipole Attention~\citep{hooper2025multipole} & \slot{\supported}\slot{\unsupported}\slot{\unsupported} & \slot[12pt]{\unsupported}\slot[12pt]{\supported} & \slot{\unsupported}\slot{\supported}\slot[10pt]{\fixedsearch} & \slot[16pt]{128k}\slot[16pt]{128k} \\
LycheeCluster~\citep{li2026lycheecluster} & \slot{\supported}\slot{\unsupported}\slot{\unsupported} & \slot[12pt]{\unsupported}\slot[12pt]{\supported} & \slot{\supported}\slot{\supported}\slot[10pt]{\fixedsearch} & \slot[16pt]{128k}\slot[16pt]{72k} \\
IceCache~\citep{mao2026icecache} & \slot{\supported}\slot{\unsupported}\slot{\unsupported} & \slot[12pt]{\unsupported}\slot[12pt]{\supported} & \slot{\unsupported}\slot{\supported}\slot[10pt]{\treesearch} & \slot[16pt]{250k}\slot[16pt]{300k} \\
\midrule
\textbf{\name} & \slot{\supported}\slot{\supported}\slot{\supported} & \slot[12pt]{\supported}\slot[12pt]{\supported} & \slot{\supported}\slot{\supported}\slot[10pt]{\treesearch} & \slot[16pt]{\textbf{10M}}\slot[16pt]{\textbf{10M}} \\
\bottomrule
\end{tabular*}
\end{table}

Table~\ref{tab:related-work} provides an overview of the existing work on sparse attention, as categorised above.
We focus on methods that keep the whole KV cache and select keys per query, unlike fixed sparse patterns~\citep{child2019generating,beltagy2020longformer,zaheer2020big} and KV cache eviction techniques~\citep{zhang2023h2o,li2024snapkv,xiao2024efficient}.

Among distillation methods, DSA~\citep{deepseekai2025deepseekv32} and MiniMax SA~\citep{lai2026minimax} first distil the indexer and then continue training the full model under sparse attention.
Alternatively, SeerAttention~\citep{gao2025seerattention,gao2026sparse} distils a gate over pooled blocks and HashAttention~\citep{desai2025hashattention} learns binary codes of queries and keys, both with the model frozen.
As the context grows, the indexer itself becomes the bottleneck~\citep{xu2026hisa,zan2026longcat,bai2026indexcache}.
To reduce its cost, CSA~\citep{deepseekai2026deepseekv4}, QSA~\citep{qwen2026design}, and \glm{} SA~\citep{zai2026glm53flash} compress every four keys into one before scoring them.
HISA~\citep{xu2026hisa} and LongCat LSA~\citep{zan2026longcat} score blocks first and then only the keys inside the selected blocks.
IndexCache~\citep{bai2026indexcache}, LongCat LSA and DeepSeek-V4.1~\citep{deepseekai2026deepseekv41flash} reuse the selection of one layer in the following layers.
Each improvement divides the cost of the indexer by a constant, and its work per query therefore still grows linearly with the context.

NSA~\citep{yuan2025native} and MoBA~\citep{lu2025moba} instead pre-train the model with sparse attention from scratch, such that the indexer learns from the NTP loss alone.
NSA scores each block by a learned compressed key, and MoBA by the mean of its keys.
InfLLM-V2~\citep{zhao2026infllmv2} and SSA~\citep{mao2026simplified} bring such block selection to existing dense models by fine-tuning.
HSA~\citep{hu2025hardware,leng2026understanding,hu2026hierarchical} and HSA-UltraLong~\citep{hu2026every} also weight the output of each block by its score, such that the loss trains the scores directly.
All of these methods still scan every block.
PISA~\citep{tang2026block} replaces this scan with a tree, but evaluates accuracy only at small scale (up to 16k tokens).

At the other end, Quest~\citep{tang2024quest}, InfLLM~\citep{xiao2024infllm} and MInference~\citep{jiang2024minference} select blocks based on simple summaries of their keys without any training.
HiP~\citep{lee2025training} searches the blocks with a tree, and its successor InfiniteHiP~\citep{lee2025infinitehip} with three fixed stages.
BLASST~\citep{yuan2026blasst} instead skips blocks directly inside the attention kernel, without a dedicated indexer.
Most of these methods summarise blocks of consecutive keys, which may hide a single relevant key~\citep{xiao2024infllm,hu2026hierarchical}.

Closest to \name{} are methods that organise the keys by semantic content rather than by position.
A few early works train at short context, with Reformer~\citep{kitaev2020reformer} hashing keys into buckets, Routing Transformer~\citep{roy2021efficient} learning k-means centroids, and Treeformer~\citep{madaan2023treeformer} routing each query down a learned tree.
Most other works here only operate at inference time: ClusterKV~\citep{liu2025clusterkv} and RetroInfer~\citep{chen2026retroinfer} cluster the keys during prefill and score every centroid at decode (flat search), Squeezed Attention~\citep{hooper2025squeezed}, Multipole Attention~\citep{hooper2025multipole} and LycheeCluster~\citep{li2026lycheecluster} search two levels (fixed search), and IceCache~\citep{mao2026icecache} descends a tree of keys (tree search), while RetrievalAttention~\citep{liu2025retrievalattention} links the keys in a nearest-neighbour graph.
All of these inference methods are training-free, only Squeezed Attention speeds up prefill, and only partially, and none are evaluated at ultra-long context (above 1M tokens).

\begin{figure*}[t]
\centering
\includegraphics[width=\textwidth]{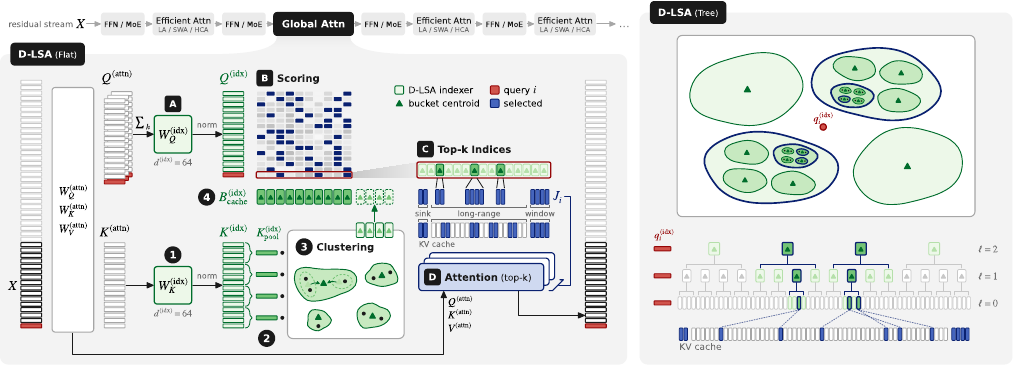}
\caption{\textbf{\dname{} design.} Left: \dname{} (Flat). Indexer keys: \keystep{1} projection and L2 norm, \keystep{2} pooling (4:1), \keystep{3} bottom-up agglomerative clustering (Ward's method), \keystep{4} bucket cache (centroids and index table). Indexer queries: \querystep{A} sum over query heads, projection and L2 norm, \querystep{B} scoring the cached centroids and selecting the top-p buckets per query, \querystep{C} expanding the buckets into indices (sink + top-k long-range + window), \querystep{D} attention over those indices (one index list per query shared by all query heads). Right: \dname{} (Tree) groups the buckets into a hierarchy of clusters, a query descends using beam search (close in the latent space but scattered through the KV cache).}
\label{fig:design}
\end{figure*}

\section{Latent Sparse Attention (\name{})}
\label{sec:lsa}

\name{} is a sparse attention layer for ultra-long context.
It clusters keys by content, in a space that is learned: through index projections in \dname{}, and through the attention's own keys in \nname{}.
\name{} can replace the attention of a pre-trained layer by distillation (\dname{}) or be trained natively with the model (\nname{}).

In both settings, the indices for the attention are selected from the keys and queries by the same process.
We first describe this process, numbered as in Figure~\ref{fig:design}, which consists of semantic clustering of the index keys $k_j^{(\mathrm{idx})}$, and the search with the index queries $q_i^{(\mathrm{idx})}$ so that each query selects its keys.
We then describe the specifics of \dname{} and \nname{}.

\subsection{Semantic Clustering}

\keystep{1} \textbf{Key normalisation.}
Each index key $k_j^{(\mathrm{idx})}$ is first L2 normalised: $k_j^{(\mathrm{idx})} \leftarrow \frac{k_j^{(\mathrm{idx})}}{\lVert k_j^{(\mathrm{idx})} \rVert}$

\keystep{2} \textbf{Pooling.}
Consecutive keys are then compressed, as in CSA~\citep{deepseekai2026deepseekv4}, QSA~\citep{qwen2026design} and \glm{} SA~\citep{zai2026glm53flash}, to improve computational efficiency. Here, every $n_{\mathrm{pool}}$ consecutive index keys are summed into one pool key $k_{\mathrm{pool}}^{(\mathrm{idx})}$, so that clustering runs over $T / n_{\mathrm{pool}}$ pool keys rather than $T$ keys.

\keystep{3} \textbf{Clustering.}
Keys are clustered in cluster blocks of $n_{\mathrm{cluster}}$ consecutive keys.
Once a cluster block contains all its keys, its pool keys are clustered bottom up with Ward's method~\citep{ward1963hierarchical} into $m$ buckets (Algorithm~\ref{alg:ward} in Appendix~\ref{app:method}).
Each pool key starts as a bucket of its own, and each bucket keeps the sum of its keys, whose normalisation is its centroid $c_b$.
In every round, all pairs of buckets that are each other's cheapest merge are merged in parallel.
The cost of a merge is how much it increases the keys’ total squared distance from their centroid (Ward’s criterion).
Merging stops once there are $m = n_{\mathrm{cluster}} / \bar{n}_b$ buckets, where $\bar{n}_b$ is the average number of keys per bucket, such that every cluster block holds the same number of buckets.
A merge is skipped if the merged bucket would hold more than $n_b^{\mathrm{max}}$ keys.

\keystep{4} \textbf{Bucket cache.}
The bucket cache $B_{\mathrm{cache}}^{(\mathrm{idx})}$ stores the centroid $c_b$ of every bucket and an index table of the positions of its keys. This is so later selected buckets can `expand' into their key indices. With one centroid per $\bar{n}_b$ keys, it corresponds to a small fraction of the KV cache.

Once a cluster block is clustered, its buckets never change. In the \emph{Tree} variant, this clustering process is repeated to group the buckets into multiple levels of a hierarchy.

\subsection{Semantic Search}

We now describe how the top-$k$ key indices are extracted for each query, and then submitted to attention. We first describe the process for \emph{Flat}, in which there is only one level of buckets, and then discuss \emph{Tree}.

\querystep{A} \textbf{Query normalisation.}
Each index query $q_i^{(\mathrm{idx})}$ is first L2 normalised: $q_i^{(\mathrm{idx})} \leftarrow \frac{q_i^{(\mathrm{idx})}}{\lVert q_i^{(\mathrm{idx})} \rVert}$

\querystep{B} \textbf{Scoring.}
The index query scores the unit-norm centroid of every bucket in the cluster blocks before its own via $s_{i,b} = c_b \cdot q_i^{(\mathrm{idx})}$, which compares buckets of any size on one scale, and selects the top-$p$ buckets.
Its own cluster block is not clustered yet, and its keys are covered by the window instead.

\querystep{C} \textbf{Expansion.}
The selected $p$ buckets are expanded into their keys, best first, until $k$ `long-range' keys are selected.
As every bucket holds at least $n_{\mathrm{pool}}$ keys, $p$ is chosen such that $p \cdot n_{\mathrm{pool}} \ge k$, and so expansion always fills $k$ keys. 

Besides the $k$ long-range keys, every query attends to $n_{\mathrm{sink}}$ sink keys~\citep{xiao2024efficient} and the $n_w$ most recent keys (window).
A key of a selected bucket that falls in the sink or the window is already attended to, and the expansion moves on to the next key instead.
We set the window to one cluster block, $n_w = n_{\mathrm{cluster}}$, so that the window covers every key not yet clustered.
This gives each query $i$ a set $J_i$ of at most $n_{\mathrm{sink}} + k + n_w$ key indices.

\querystep{D} \textbf{Attention.}
Attention is then computed only over the keys indexed by $J_i$. A single $J_i$ is shared across all query heads.

\emph{Tree} search arranges the buckets into a hierarchy (Figure~\ref{fig:design}, right).
Level 0 holds the buckets, and each higher level $\ell \ge 1$ holds buckets we term `node buckets', each made of $n_{b,\ell}$ buckets of the level below, with a centroid over all the keys it contains.
As at level 0, buckets are merged one cluster block at a time.
Buckets that are not merged yet are defined as open.
Once a level holds $n_{\mathrm{cluster},\ell}$ open buckets, they are merged into node buckets as in Algorithm~\ref{alg:ward}.

Beam search is used to select the buckets through the hierarchy (Algorithm~\ref{alg:beam} in Appendix~\ref{app:method}).
On each level, the query is scored against the centroids of the candidate node buckets, and the best $p_\ell$ are kept.
The candidates on the level below are the children of the kept node buckets and the open buckets of that level.
At level 0, the top-$p$ buckets are selected, as in the flat search.
With a constant beam per level, a query scores $\bigO{\log T}$ centroids, which makes the prefill of the indexer subquadratic, $\bigO{T \log T}$, where the flat search is quadratic.

\subsection{\dname{}}

\dname{} (Figure~\ref{fig:design}) follows the process outlined above, but since it distils from a pre-trained model which already has key and query projections ($k_j^{(\mathrm{attn})}$ and $q_i^{(\mathrm{attn})}$), we additionally learn separate key and query projections for \name{}. In particular, each key $k_j^{(\mathrm{attn})}$ and query $q_i^{(\mathrm{attn})}$ is projected to a smaller dimension $d^{(\mathrm{idx})}$ to keep the scoring, clustering, and bucket cache cheap:
\begin{equation}
\begin{aligned}
k_j^{(\mathrm{idx})} &= k_j^{(\mathrm{attn})} W_K^{(\mathrm{idx})} \\
q_i^{(\mathrm{idx})} &= \sum_h q_{i,h}^{(\mathrm{attn})} W_Q^{(\mathrm{idx})}
\end{aligned}
\label{eq:projection}
\end{equation}
The query heads are summed such that one index query serves every head, which yields the single index list $J_i$ per query.
On \qwen{}, whose keys and queries carry RoPE~\citep{su2024roformer} on a quarter of their dimensions, the projections read only the RoPE-free part.
These keys and queries then go into the normalisation process as before.
Trainable projections can also be replaced by the leading $d^{(\mathrm{idx})}$ principal directions of the keys in the layer, which project both keys and queries into one index space with no training.
We call this variant \dname{} (Training-Free).

The base model's attention is otherwise unchanged: its own queries attend only to the keys and values indexed by $J_i$.

\textbf{Distillation.}
\dname{} adds \name{} to a pre-trained model by a short distillation that trains only $W_{Q,K}^{(\mathrm{idx})}$ and a temperature $\tau$ per layer, with the base model frozen.
This mirrors the dense warm-up of DSA~\citep{deepseekai2025deepseekv32}, which also trains only the indexer against the head-aggregated attention of the frozen model.
The distillation scores every key directly, without clustering.
For a query $i$ of a layer, let $\mathcal{R}_i$ be the keys it can select (all preceding keys outside the sink and the window).
The target $A_i$ is the attention of that frozen layer over $\mathcal{R}_i$, the softmax of each head's logits over $\mathcal{R}_i$ averaged over its $n_h$ heads of dimension $d_h$, and $\hat{A}_i$ is the softmax of the \dname{} scores over the same keys.
The distillation minimises the forward KL divergence between the two:
\begin{equation}
\begin{aligned}
\mathcal{L}_i &= \mathrm{KL}\big(A_i \,\Vert\, \hat{A}_i\big) \\
A_{i,j} &= \tfrac{1}{n_h} \textstyle\sum_h \mathrm{softmax}_{j \in \mathcal{R}_i}\big(q_{i,h}^{(\mathrm{attn})} \cdot k_j^{(\mathrm{attn})} / \sqrt{d_h}\big) \\
\hat{A}_{i,j} &= \mathrm{softmax}_{j \in \mathcal{R}_i}\big(q_i^{(\mathrm{idx})} \cdot k_j^{(\mathrm{idx})} / \tau\big)
\end{aligned}
\label{eq:distillation}
\end{equation}
The temperature $\tau$ sharpens the softmax of the \dname{} scores to match the peaked attention of the model, and is used only in this loss.
Both projections are initialised with these principal directions.

\textbf{Kernels.}
We write GPU kernels for the clustering, scoring, selection and expansion of the indexer, in CuTe DSL for H200 and B300, with a Triton version of each, which runs on any GPU.
For the attention, we reuse existing attention kernels, such as FlashAttention-3~\citep{shah2024flashattention}, which read the index list $J_i$ as a page table.

\textbf{Parallelism.}
We integrate \name{} into SGLang with support for tensor, expert, and pipeline parallelism, and for context parallelism in both prefill and decode.
Context parallelism splits the KV cache across ranks by token, but \name{} replicates its bucket cache on every rank.
Every rank thus builds the same index and selects the same index list. The ranks only exchange the index keys of the newest tokens.
This design choice trades memory for communication.
At 10M tokens on \glm{} with context parallelism over 8 GPUs, each rank holds a 1.25\,GiB share of the 10\,GiB KV cache of a layer and the full 246\,MiB bucket cache.

We provide further details on kernels in Appendix~\ref{app:kernels} and on parallelism in Appendix~\ref{app:parallelism}.

\subsection{\nname{}}

\begin{figure}[t]
\centering
\includegraphics[width=\columnwidth]{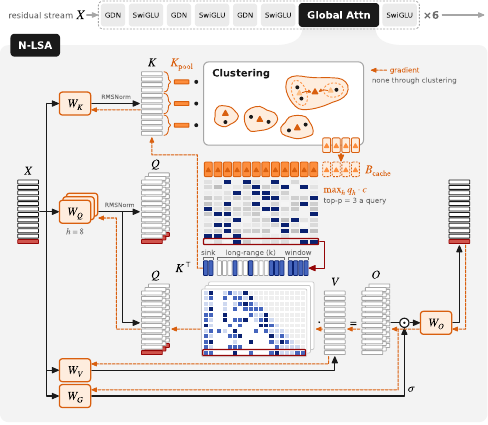}
\caption{\textbf{\nname{} design.} The attention layer of \nname{}, which can be trained from scratch on the next-token loss alone.}
\label{fig:nlsa}
\end{figure}

\nname{} (Figure~\ref{fig:nlsa}) trains \name{} natively with the model, using only the NTP loss.
As the model learns with the selection from the start, we cluster the keys of the attention itself, without dedicated index projections.
We use a single key-value head, such that one key per token serves every query head, and score each bucket by the best similarity any query head gives its centroid:
\begin{equation}
s_{i,b} = \max_h \hat{q}_{i,h} \cdot c_b
\label{eq:native-score}
\end{equation}
where $\hat{q}_{i,h}$ is the normalised query of head $h$.
We stop the gradient through the selection, and the NTP loss trains the attention projections only through the keys each query attends to, the selected keys plus the sink and the window.
\nname{} therefore adds no parameters to the model and needs no teacher, no index projections and no auxiliary loss.

The attention layers of \nname{} use no positional encoding, as is becoming more common in hybrid models, and RoPE could be supported by clustering only the RoPE-free part of the keys, as \dname{} does on \qwen{}.

\section{Evaluation}
\label{sec:evaluation}

In this section, we describe the models, the methods we compare, and their settings.
Then, we evaluate the accuracy of \dname{} up to 10M tokens, and its performance per attention layer and end to end.
Finally, we show that \nname{} can be trained natively at an accuracy close to full attention.

\subsection{Setup}
\label{sec:setup}

\textbf{Models.}
\dname{} aims to keep the accuracy of a model while making its attention significantly cheaper. We compare each model with \dname{} against its baseline attention, in accuracy and in performance.
We demonstrate \dname{} on both a model with full attention (\qwen{}~\citep{qwen38,qwen38_27b}) and one with sparse attention (\glm{}~\citep{zai2026glm53flash}). On \qwen{}, we also replace the attention with block sparse attention, inspired by QSA~\citep{qwen2026design}, as a second sparse baseline. We also compare flat to tree search where possible.

\textbf{Methods.}
We compare the following methods:
\begin{itemize}[leftmargin=*,nosep]
\item \boxfull~\textbf{Full Attention}: every query attends to every preceding key, the baseline attention of \qwen{}.
\item \boxblockflat~\textbf{Block Sparse Attention (Flat)}: sparse attention whose indexer scores every block of four consecutive keys, as in QSA~\citep{qwen2026design} and \glm{}~\citep{zai2026glm53flash}.
\item \boxblocktree~\textbf{Block Sparse Attention (Tree)}: sparse attention whose indexer searches a tree over blocks of the keys of the attention itself, as in HiP~\citep{lee2025training} and PISA~\citep{tang2026block}. To represent these methods, we search a tree over blocks of keys directly on the KV cache, without training, which uses the full keys and favours accuracy over performance. We use blocks of eight keys, the most accurate of the sizes we compare (Appendix~\ref{app:block-size}).
\item \boxlsaflat~\textbf{\dname{} (Flat)} and \boxlsatree~\textbf{\dname{} (Tree)}: \dname{} with a flat search over every bucket centroid and with a tree search over the hierarchy of buckets.
In decode, each request processes one token per step, and the tree search runs its levels one after another, which pays off only when many requests decode together.
As we evaluate one request at a time, \dname{} (Tree) decodes with the flat search (Appendix~\ref{app:kernels}).
\item \boxlsafree~\textbf{\dname{} (Training-Free)}: \dname{} (Flat) with the principal directions of the keys as both projections instead of trained ones.
\end{itemize}
Every trained indexer we compare is distilled in the same way to ensure fairness.

\textbf{Design parameters.}
Table~\ref{tab:methods} lists the parameters of every method per model.
For \dname{}, we found that a budget of $k = 4096$ long-range keys holds accuracy past 1M tokens better than smaller budgets, and these extra keys cost little next to the scan of the indexer they replace.
On \glm{}, \dname{} (Flat) selects $k = 2048$ long-range keys for queries before position 1M and $k = 4096$ beyond.
We found that the smaller budget keeps accuracy below 1M tokens while making the attention faster.
Tuning the budget further, for example per layer or per request, is left to future work.
A small index space of $d^{(\mathrm{idx})} = 64$ dimensions keeps the scoring and the clustering cheap, and we found wider spaces to add little accuracy.

\begin{table}[t]
\centering
\caption{\textbf{Parameters.}
\boxblockflat~Block Sparse Attention (Flat), \boxblocktree~Block Sparse Attention (Tree), \boxlsaflat~\dname{} (Flat) and \boxlsatree~\dname{} (Tree) per model (symbols in Table~\ref{tab:notation}).
The search is flat (\flatsearch) or tree (\treesearch), over blocks of $n_{\mathrm{pool}}$ consecutive keys or over buckets.
Block Sparse Attention (Tree) always keeps the first, previous and current blocks among its $p$ blocks.
On \glm{}, \dname{} (Flat) uses the first $k$ and $p$ below 1M tokens and the second beyond.}
\label{tab:methods}
\vskip 0.15in
\ssmall
\setlength{\tabcolsep}{1pt}
\begin{tabular*}{\columnwidth}{@{\extracolsep{\fill}}>{\rule[-1.1ex]{0pt}{3.3ex}}l!{\separator}cccc!{\separator}cccc@{}}
\toprule
 & \multicolumn{4}{c!{\separator}}{\textbf{\qwen{}}} & \multicolumn{4}{c}{\textbf{\glm{}}} \\
 & \boxblockflat & \boxblocktree & \boxlsaflat & \boxlsatree & \boxblockflat & \boxblocktree & \boxlsaflat & \boxlsatree \\
\midrule
Unit & block & block & bucket & bucket & block & block & bucket & bucket \\
Search & \flatsearch & \treesearch & \flatsearch & \treesearch & \flatsearch & \treesearch & \flatsearch & \treesearch \\
$k$ & 2,048 & 4,608 & 4,096 & 4,096 & 2,048 & 4,608 & 2,048/4,096 & 4,096 \\
$p$ & 512 & 576 & 1,024 & 1,024 & 512 & 576 & 512/1,024 & 1,024 \\
$n_{\mathrm{sink}}$, $n_w$ & -, - & 8, 16 & 4, 512 & 4, 512 & -, - & 8, 16 & 4, 512 & 4, 512 \\
$d^{(\mathrm{idx})}$ & 128 & 256 & 64 & 64 & 128 & 512 & 64 & 64 \\
$n_{\mathrm{pool}}$ & 4 & 8 & 4 & 4 & 4 & 8 & 4 & 4 \\
$n_{\mathrm{cluster}}$ & - & - & 512 & 512 & - & - & 512 & 512 \\
$\bar{n}_b$, $n_b^{\mathrm{max}}$ & - & - & 8, 16 & 8, 16 & - & - & 8, 16 & 8, 16 \\
$n_{b,\ell}$ & - & 2 & - & 4 & - & 2 & - & 4 \\
$n_{\mathrm{cluster},\ell}$ & - & - & - & 4,096 & - & - & - & 4,096 \\
$p_\ell$ & - & 576 & - & 2,048 & - & 576 & - & 2,048 \\
\bottomrule
\end{tabular*}
\end{table}

\textbf{Indexer distillation.}
We train on 352M tokens of documents of 16k to 1M tokens from seven domains, for 48k steps, each on 128 queries sampled from one document, with a learning rate of $3 \cdot 10^{-4}$.
This is a fraction of the 2.1B tokens of the dense warm-up of DSA.
In Equation~\ref{eq:distillation}, the target $A_i$ keeps only the 2,048 keys with the most attention weight per query, renormalised over them, while $\hat{A}_i$ still spans all of $\mathcal{R}_i$, and $\tau$ is calibrated per layer before the first step.
Unlike DSA, we do not follow the distillation by training the whole model under sparse attention (sparse training), a step \dname{} also supports but we omit here due to its cost.
We provide further details about the composition and construction of the dataset in Appendix~\ref{sec:training-data}.

\textbf{Benchmarks.}
We evaluate accuracy in three categories, each with a representative set of benchmarks.
For knowledge, we use MMLU-Pro~\citep{wang2024mmlupro} (a stratified sample of 2,000 questions) and GPQA-Diamond~\citep{rein2024gpqa}.
For reasoning, we use AIME25 (four repeats) and MATH500~\citep{lightman2024lets} (100 problems).
For long context, we use LongBench v2~\citep{bai2025longbench} (60 questions) and the 13 tasks of RULER~\citep{hsieh2024ruler} from 4k to 10M tokens (100 prompts per task up to 1M and 20 beyond).
Evaluating each benchmark in full at ultra-long context is prohibitively expensive, and we therefore use the subsets given in parentheses.

\textbf{PaperTrail.}
\label{sec:papertrail-intro}
To better understand model capabilities in a less synthetic ultra-long context setting, we extend RULER with a set of new tasks called PaperTrail.
The tasks are constructed by embedding a trail of fabricated facts in a large collection of scientific papers.
All papers are recent, almost all published after the release date of the models, such that no model could have seen them in training.
A question can only be answered by finding the facts of its trail and connecting them, as in a literature search.
Samples are built in three domains: exoplanets, long-context machine learning and quantum error correction.
The questions come in three tiers of increasing difficulty, where tier 1 recalls a fact from one paper, tier 2 links facts across two papers and tier 3 traces a trail across three or more.
We provide further details about the construction of PaperTrail in Appendix~\ref{sec:papertrail}.

\textbf{Model sampling.}
We configure each model's sampling as recommended by its model card with generation up to 163,840 tokens (Appendix~\ref{app:accuracy}).
\qwen{} reaches 10M tokens with YaRN~\citep{peng2024yarn} at factor 40 (42 for PaperTrail at 10M), while \glm{} needs no extension as it has no explicit positional encoding.

\textbf{Hardware and serving.}
We serve all models in SGLang~\citep{zheng2024sglang} on H200 and B300 GPUs.
To show that \dname{} can be integrated with the main parallelism strategies, we evaluate it with tensor parallelism (TP8) and with context parallelism in both prefill and decode (CP8). We show more details in Appendix~\ref{app:parallelism}.
All accuracy measurements use context parallelism wherever the KV cache of a request outgrows the memory of a single GPU.

\subsection{Accuracy}
\label{sec:accuracy}

\begin{figure*}[t]
\centering
\includegraphics[width=\textwidth]{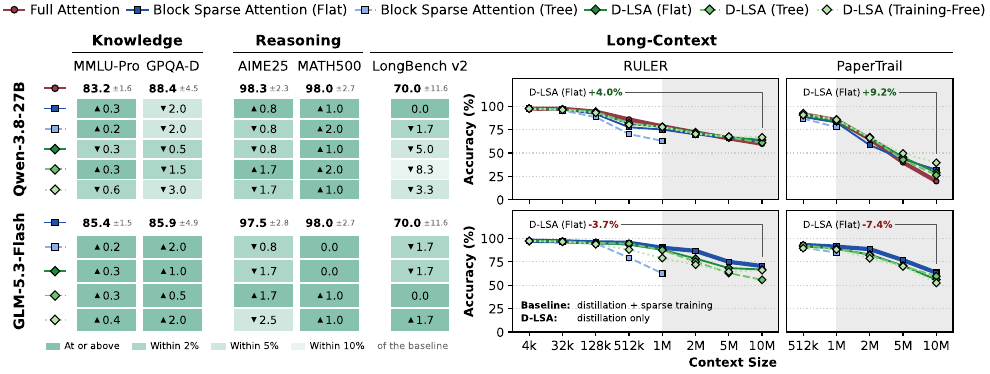}
\caption{\textbf{\dname{} accuracy is on par with full attention and sparse attention baselines on downstream tasks.} Shown on knowledge, reasoning and long-context tasks up to 10M tokens. Baseline scores are given with their 95\% confidence interval.}
\label{fig:accuracy}
\end{figure*}

Figure~\ref{fig:accuracy} shows that on knowledge, reasoning and long-context tasks, \dname{} stays close to the accuracy of baseline Full Attention on \qwen{} and of baseline sparse attention on \glm{}.
PaperTrail further shows how well the models find and connect facts across large collections of scientific papers.


\textbf{Knowledge and reasoning.}
On knowledge, reasoning and LongBench v2, \dname{} stays within the 95\% confidence interval of both baselines, even with reasoning growing to tens of thousands of tokens (Appendix~\ref{app:accuracy}).
We further show that \dname{} does not measurably affect the length of the reasoning.

\textbf{Long context.}
On \qwen{}, \dname{} (Flat) stays within 5 points of baseline Full Attention up to 1M tokens and exceeds it at 10M tokens, at 63.9 against 59.9 on RULER and 29.1 against 19.8 on PaperTrail.
It also stays above Block Sparse Attention (Flat) on RULER from 128k tokens on, in a fair comparison, as both indexers are distilled the same way.

On \glm{}, \dname{} (Flat) stays within 3 points of baseline sparse attention up to 1M tokens, trails it by up to 8 points at 2M to 5M tokens, and is within 4 points on RULER at 10M tokens.
This comparison favours the baseline, which was trained together with the model, while \dname{} (Flat) is distilled from attention that is already sparse (with no finetuning of the base model itself).
On both models, MultiKey 2, MultiKey 3 and CWE fall the most for every method, the baselines included. We break down RULER by task and PaperTrail by tier in Appendix~\ref{app:accuracy}.

\textbf{Design comparison.}
\dname{} (Tree) follows \dname{} (Flat) on \qwen{}, but loses accuracy past 1M tokens on \glm{}, at 55.8 against 66.8 on RULER at 10M tokens.
Block Sparse Attention (Tree) falls behind from 512k tokens, to about 63 on RULER at 1M tokens on both models, against 78 and 88 for \dname{} (Tree), and we therefore do not evaluate it beyond 1M tokens.
Large blocks fall earlier (Appendix~\ref{app:block-size}).
\dname{} (Training-Free), without any training, stays close to \dname{} (Flat) on \qwen{}, but trails it on \glm{}.

\dname{} follows the accuracy of the baselines, showing that our semantic clustering can preserve accuracy up to 10M tokens. However, we do observe a steep drop in accuracy beyond 512k tokens for Block Sparse Attention (Tree).

\subsection{Performance}
\label{sec:performance}

\begin{figure*}[t]
\centering
\includegraphics[width=\textwidth]{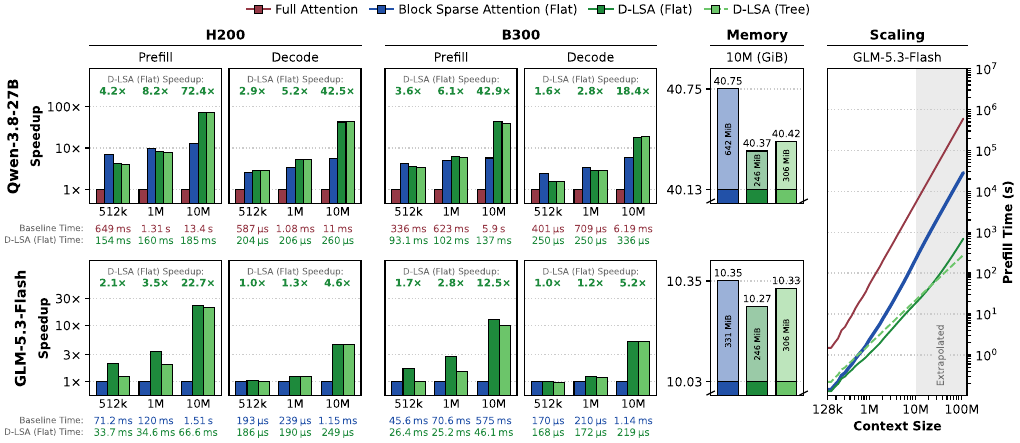}
\caption{\textbf{\dname{} speedup of one global attention layer.}
Prefill and decode speedup over the baseline attention (log scale), Full Attention for \qwen{} and Block Sparse Attention (Flat) for \glm{}, required memory at 10M with one index key per block of four keys for Block Sparse Attention (Flat), and prefill scaling up to 100M (log-log), extrapolated beyond 10M (shaded).}
\label{fig:layer-performance}
\end{figure*}

Next, we evaluate the performance of \dname{}.
We first measure a single attention layer and then the whole model in production serving.

\textbf{Methodology.}
For one layer, we time the whole attention module, including its projections, the indexer and the attention, on a single GPU.
In prefill, we time the last chunk of 32,768 queries before the given context.
In decode, we time one step of one request with CUDA graph enabled (clustering is amortised across the steps).
For the whole model, we serve each method in SGLang with prefill chunks of 32,768 tokens, and measure the time to first token (TTFT) and the time per output token (TPOT) of one request at a time, the median of three requests of 256 output tokens.

Figure~\ref{fig:layer-performance} shows that the prefill speedup of \dname{} grows with the context.
On \glm{}, it rises from 3.5$\times$ at 1M to 22.7$\times$ at 10M tokens over baseline sparse attention, and on \qwen{} to 72.4$\times$ at 10M tokens over baseline Full Attention.
Below 1M tokens, the gain is smaller, at 2.1$\times$ at 512k tokens, as the scan of every block is still cheap.
Extrapolated beyond 10M tokens, the prefill of \dname{} (Tree) grows subquadratically and overtakes the flat search, and at 100M tokens it is predicted to be more than 90$\times$ faster than baseline sparse attention.

Decode follows the same trend, with a speedup up to 4.6$\times$ over baseline sparse attention and 42.5$\times$ over baseline Full Attention at 10M tokens, with \dname{} (Tree) decoding with the flat search.
On B300, \dname{} reaches 12.5$\times$ and 42.9$\times$ in prefill and 5.2$\times$ and 18.4$\times$ in decode at 10M tokens.

\dname{} also requires less memory for its indexer.
Baseline sparse attention caches an index key for every block of four keys, while \dname{} caches one centroid in its smaller index space for every bucket of about eight keys, together with the id of each key.
At 10M tokens, the index of \dname{} (Flat) takes 246\,MiB per layer against 331\,MiB in \glm{} and 642\,MiB in QSA.
Larger buckets would shrink the index of \dname{} further.

\begin{table}[!t]
\centering
\caption{\textbf{\dname{} speedup in production setting.}
Time to first token (TTFT) and time per output token (TPOT) of one request under SGLang.
\boxfull~Full Attention, \boxblockflat~Block Sparse Attention (Flat), \boxlsaflat~\dname{} (Flat) and \boxlsatree~\dname{} (Tree).
Each block's first row gives the baseline's times and the others their speedup over it, with \dname{} at 10M in bold.
\qwen{} is served with tensor parallelism (TP8) and decode context parallelism (DCP8) on H200 and with TP8 on B300, \glm{} with either context parallelism (CP8, with DCP8 and expert parallelism EP8) or TP8.
Without DCP, Block Sparse Attention's prefill indexer runs whole on every TP rank.
All models run on a single node, except TP8 for \glm{} at 10M on H200 (2 nodes with PP2); a B300 node holds it on one.}
\label{tab:production}
\vskip 0.15in
\setlength{\tabcolsep}{8.6pt}
\ssmall
\begin{tabular}{@{}lc!{\separator}rrr!{\separator}rrr@{}}
\toprule
 & & \multicolumn{3}{c}{\textbf{TTFT (s)}} & \multicolumn{3}{c}{\textbf{TPOT (ms)}} \\
\cmidrule(lr){3-5} \cmidrule(l){6-8}
 & & 512k & 1M & 10M & 512k & 1M & 10M \\
\midrule
\multicolumn{8}{@{}l}{\textit{H200}} \\
\addlinespace[1.5pt]
\multicolumn{8}{@{}l}{\textbf{\qwen{}}} \\
\addlinespace[1.5pt]
\multirow{4}{*}{\rotatebox{90}{\shortstack{TP8\\DCP8}}} & \boxfull & 21.7 & 65.3 & 4582.2 & 7.8 & 9.1 & 32.1 \\
 & \boxblockflat & 1.61$\times$ & 2.30$\times$ & 8.02$\times$ & 0.89$\times$ & 0.91$\times$ & 0.89$\times$ \\
 & \boxlsaflat & 1.03$\times$ & 1.51$\times$ & \textbf{9.21$\times$} & 0.87$\times$ & 1.00$\times$ & \textbf{2.87$\times$} \\
 & \boxlsatree & 0.96$\times$ & 1.41$\times$ & \textbf{8.83$\times$} & 0.87$\times$ & 1.00$\times$ & \textbf{2.87$\times$} \\
\arrayrulecolor{black!30}\cmidrule[0.5pt]{1-8}\arrayrulecolor{black}
\multicolumn{8}{@{}l}{\textbf{\glm{}}} \\
\addlinespace[1.5pt]
\multirow{3}{*}{\rotatebox{90}{\shortstack{CP8\\DCP8}}} & \boxblockflat & 8.5 & 18.1 & 451.5 & 7.6 & 8.2 & 17.9 \\
 & \boxlsaflat & 1.03$\times$ & 1.09$\times$ & \textbf{2.24$\times$} & 1.01$\times$ & 1.06$\times$ & \textbf{2.10$\times$} \\
 & \boxlsatree & 0.93$\times$ & 0.98$\times$ & \textbf{2.14$\times$} & 0.97$\times$ & 1.03$\times$ & \textbf{2.10$\times$} \\
\addlinespace[4pt]
\multirow{3}{*}{\rotatebox{90}{\shortstack{TP8}}} & \boxblockflat & 17.9 & 45.1 & 1300.9 & 5.8 & 6.3 & 18.9 \\
 & \boxlsaflat & 1.22$\times$ & 1.51$\times$ & \textbf{5.83$\times$} & 0.99$\times$ & 1.06$\times$ & \textbf{1.91$\times$} \\
 & \boxlsatree & 0.98$\times$ & 1.21$\times$ & \textbf{5.62$\times$} & 0.95$\times$ & 1.03$\times$ & \textbf{1.89$\times$} \\
\midrule
\multicolumn{8}{@{}l}{\textit{B300}} \\
\addlinespace[1.5pt]
\multicolumn{8}{@{}l}{\textbf{\qwen{}}} \\
\addlinespace[1.5pt]
\multirow{4}{*}{\rotatebox{90}{\shortstack{TP8}}} & \boxfull & 10.6 & 31.6 & 2172.2 & 4.9 & 6.0 & 27.3 \\
 & \boxblockflat & 0.79$\times$ & 0.82$\times$ & 0.84$\times$ & 0.82$\times$ & 0.90$\times$ & 1.33$\times$ \\
 & \boxlsaflat & 0.68$\times$ & 0.96$\times$ & \textbf{5.73$\times$} & 0.63$\times$ & 0.73$\times$ & \textbf{1.46$\times$} \\
 & \boxlsatree & 0.63$\times$ & 0.91$\times$ & \textbf{5.40$\times$} & 0.63$\times$ & 0.73$\times$ & \textbf{1.46$\times$} \\
\arrayrulecolor{black!30}\cmidrule[0.5pt]{1-8}\arrayrulecolor{black}
\multicolumn{8}{@{}l}{\textbf{\glm{}}} \\
\addlinespace[1.5pt]
\multirow{3}{*}{\rotatebox{90}{\shortstack{CP8\\DCP8}}} & \boxblockflat & 6.5 & 13.5 & 273.1 & 6.9 & 7.4 & 17.3 \\
 & \boxlsaflat & 0.96$\times$ & 0.99$\times$ & \textbf{1.56$\times$} & 0.97$\times$ & 1.03$\times$ & \textbf{2.21$\times$} \\
 & \boxlsatree & 0.78$\times$ & 0.81$\times$ & \textbf{1.48$\times$} & 0.96$\times$ & 1.03$\times$ & \textbf{2.22$\times$} \\
\addlinespace[4pt]
\multirow{3}{*}{\rotatebox{90}{\shortstack{TP8}}} & \boxblockflat & 11.1 & 27.2 & 1186.4 & 5.2 & 5.7 & 15.6 \\
 & \boxlsaflat & 1.15$\times$ & 1.39$\times$ & \textbf{4.76$\times$} & 0.96$\times$ & 1.03$\times$ & \textbf{2.56$\times$} \\
 & \boxlsatree & 0.91$\times$ & 1.09$\times$ & \textbf{4.28$\times$} & 0.95$\times$ & 1.03$\times$ & \textbf{2.56$\times$} \\
\bottomrule
\end{tabular}
\end{table}

\textbf{End-to-end.}
Table~\ref{tab:production} shows that the speedup of one layer carries over to the whole model at ultra-long context.
On \glm{}, \dname{} (Flat) is on par with baseline sparse attention up to 1M tokens with CP8 and already 1.5$\times$ faster to first token at 1M with TP8, and at 10M tokens reduces the TTFT by 2.2$\times$ and the TPOT by 2.1$\times$ with CP8, and by 5.8$\times$ and 1.9$\times$ with TP8.
On \qwen{}, it reduces them by 9.2$\times$ and 2.9$\times$ at 10M tokens against baseline Full Attention.
On B300, \dname{} (Flat) reduces the TTFT and the TPOT at 10M tokens by 1.6$\times$ and 2.2$\times$ with CP8 and by 4.8$\times$ and 2.6$\times$ with TP8 on \glm{}, and by 5.7$\times$ and 1.5$\times$ on \qwen{}.
The speedup of the whole model is expected to be smaller than that of one layer, as only the global attention layers become faster, while the other layers keep their cost.

\subsection{Native Training}
\label{sec:native}

\begin{figure}[t]
\centering
\includegraphics[width=\columnwidth]{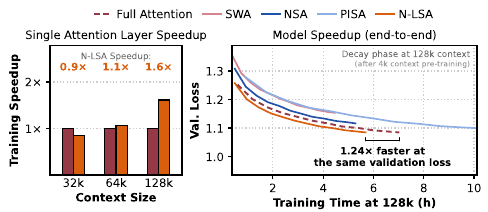}
\caption{\textbf{\nname{} training speedup.} Models trained natively on the NTP loss with Full Attention, SWA, NSA, PISA and \nname{}. Left: one attention layer, forward and backward, on one H200. Right: validation loss over the training time of the whole model in the decay phase (all on the same data and number of tokens).}
\label{fig:native}
\end{figure}

\begin{table}[t]
\centering
\caption{\textbf{Native training accuracy.}
Accuracy (\%) of models trained from scratch with \boxfull~Full Attention, \boxswa~SWA, \boxblockflat~NSA, \boxblocktree~PISA and \boxnlsa~\nname{}.
Every sparse method attends to about 2,560 keys per query.}
\label{tab:native}
\vskip 0.15in
\setlength{\tabcolsep}{1.6pt}
\ssmall
\begin{tabular}{@{}l!{\separator}ccc!{\separator}cc!{\separator}cccccc@{}}
\toprule
& \multicolumn{3}{c}{\textbf{Commonsense}} & \multicolumn{2}{c}{\textbf{Recall}} & \multicolumn{6}{c}{\textbf{RULER}} \\
\cmidrule(lr){2-4} \cmidrule(lr){5-6} \cmidrule(l){7-12}
& ARC-e & ARC-c & HellaSwag & SWDE & FDA & 4k & 8k & 16k & 32k & 64k & 128k \\
\midrule
\boxfull{} Full Attention & 50.0 & 30.0 & 45.6 & 67.8 & 79.3 & 53.9 & 46.0 & 42.5 & 38.1 & 35.6 & 34.7 \\
\arrayrulecolor{black!30}\cmidrule[0.5pt]{1-12}\arrayrulecolor{black}
\boxswa{} SWA & 53.9 & 31.4 & 45.2 & 64.6 & 69.6 & 26.7 & 15.6 & 10.1 & 5.9 & 3.0 & 3.0 \\
\boxblockflat{} NSA & 53.8 & 30.1 & 45.8 & 62.2 & 72.2 & 39.4 & 34.6 & 31.6 & 25.2 & 18.9 & 19.2 \\
\boxblocktree{} PISA & 52.2 & 28.7 & 45.5 & 63.7 & 78.6 & 56.2 & 48.4 & 40.4 & 26.6 & 19.1 & 14.6 \\
\arrayrulecolor{black!30}\cmidrule[0.5pt]{1-12}\arrayrulecolor{black}
\boxnlsa{} \nname{} & 54.6 & 30.8 & 45.5 & 66.4 & 79.2 & 52.2 & 42.5 & 41.5 & 38.4 & 32.9 & 32.2 \\
\bottomrule
\end{tabular}
\end{table}

Finally, we evaluate \nname{}, which is trained natively on the NTP loss alone.

\textbf{Setup.}
We train a hybrid model of about 0.5B parameters with 24 layers, in which every fourth layer is an attention layer and the others are Gated DeltaNet layers~\citep{yang2025gated}, a form of linear attention~\citep{schmidhuber1992fastweight,schlag2021linear}.
Each attention layer has 8 query heads over a single key-value head of 128 dimensions, without positional encoding.
We pre-train with a standard Warmup-Stable-Decay (WSD) learning rate schedule~\citep{hu2024minicpmunveilingpotentialsmall}, on 100B tokens of FineWeb-Edu~\citep{penedo2024fineweb} at a context of 4k tokens in the Warmup and Stable phases, followed by 12B tokens of LongDocs in the Decay phase, at 128k tokens (Appendix~\ref{sec:training-data}). 
We compare \nname{} with Full Attention, Sliding Window Attention (SWA), and the native forms of block sparse attention, NSA~\citep{yuan2025native} for Block Sparse Attention (Flat) and PISA~\citep{tang2026block} for Block Sparse Attention (Tree), all in the same model architecture and on the same data.
Every sparse method attends to about 2,560 keys per query.
\nname{} selects $k = 2048$ long-range keys with flat search, next to a window of 512 keys and 4 sink keys.
SWA attends to a window of 2,560 keys and 4 sink keys, NSA attends to 32 selected blocks of 64 keys and a window of 512 keys, and PISA attends to 40 blocks of 64 keys, which always include the first block and the two most recent blocks (window).
We train \nname{} with flat search only, and leave native training with tree search to future work.

Figure~\ref{fig:native} shows that \nname{} reaches the validation loss of Full Attention throughout the decay phase at 128k token context, at 1.085 against 1.085, while PISA, NSA and SWA trail at 1.100, 1.117 and 1.154.
During pre-training at 4k tokens, all methods stay within 0.004 of each other, as their keys cover most of the context.
Table~\ref{tab:native} confirms this on downstream tasks.
On commonsense reasoning~\citep{clark2018think,zellers2019hellaswag}, all models are on par.
On recall~\citep{arora2024simple}, \nname{} stays close to Full Attention and ahead of the other sparse methods.
On RULER, \nname{} stays within 3.5 points of Full Attention up to 128k tokens, at 32.2 against 34.7 at 128k tokens, while NSA and PISA fall to 19.2 and 14.6, and SWA to 3.0 once the context outgrows its window.

\nname{} also trains faster than Full Attention at long context.
For one attention layer on a single GPU, the forward and backward pass of \nname{} is 1.1$\times$ faster than Full Attention with FlashAttention-3~\citep{shah2024flashattention} at 64k tokens and 1.6$\times$ faster at 128k tokens (Figure~\ref{fig:native}, left).
End to end, the whole model completes its decay phase 1.24$\times$ faster at the same validation loss (Figure~\ref{fig:native}, right), as only every fourth layer is an attention layer.
At shorter contexts, the selection of the keys costs more than it saves.
Its peak memory is higher, at 6.8\,GiB against 5.8\,GiB per layer at 128k tokens, partly because \nname{} keeps the indices of the selected keys of every query for the backward pass, which grow linearly with the context.

Overall, \nname{} trains natively at the accuracy of Full Attention and becomes faster as the context grows.

\section{Discussion}

\name{} reduces the cost of finding the keys, but not of storing them, and the KV cache still grows linearly with the context.
It is orthogonal to methods that compress the KV cache~\citep{deepseekai2026deepseekv4}, share the index across layers~\citep{bai2026indexcache,deepseekai2026deepseekv41flash}, or predict several tokens per step~\citep{gloeckle2024better,deepseekai2024deepseekv3}, and could be combined with each.

We observed tree search to lose accuracy faster than flat search past 1M tokens on \glm{}, while it overtakes flat search in decode only from about 5M tokens with one request, and is predicted to do so in prefill only beyond 10M tokens.
It remains open whether sparse training would close this gap beyond indexer distillation.

We train \nname{} at a small scale, with about 0.5B parameters, up to 128k tokens and with flat search only.
Scaling native training to larger models, longer contexts and tree search remains open.

\section{Conclusion}

At ultra-long context, the cost of sparse attention lies in finding which keys to attend to.
We introduce \name{}, which clusters the keys in a learned latent space and replaces the scan of every key with a search over the clusters, flat or along a tree.
\dname{} distils \name{} into pre-trained models and stays close to the accuracy of their baseline attention up to 10M tokens.
At 10M tokens, one attention layer of \dname{} prefills 22.7$\times$ and decodes 4.6$\times$ faster than sparse attention in \glm{}, and 72.4$\times$ and 42.5$\times$ faster than Full Attention in \qwen{}.
End to end, this reduces the time to first token by 2.2$\times$ and 9.2$\times$, respectively.
\nname{} trains \name{} natively on the next-token loss alone, at the accuracy of Full Attention and 1.6$\times$ faster per layer at 128k tokens.
With tree search, the cost of finding the keys grows subquadratically, which makes \name{} a step towards attention that scales well beyond 10M tokens.
\name{} thus provides a practical way forward for ultra-long context attention, both for distilling pre-trained models and for native training.

\section*{Acknowledgements}
We thank Tri Dao for his valuable feedback and support.
We also thank Daniel Woods, Roman Iurkov, Cyril Kondratenko, Manuel Gomez, and the rest of the Nebius team for excellent support with the compute platform.

\bibliography{main}

\begin{thebibliography}{98}
\providecommand{\natexlab}[1]{#1}
\providecommand{\url}[1]{\texttt{#1}}
\expandafter\ifx\csname urlstyle\endcsname\relax
  \providecommand{\doi}[1]{doi: #1}\else
  \providecommand{\doi}{doi: \begingroup \urlstyle{rm}\Url}\fi

\bibitem[Alabi et~al.(2012)Alabi, Blanchard, Gordon, and Steinbach]{alabi2012fast}
Alabi, T., Blanchard, J.~D., Gordon, B., and Steinbach, R.
\newblock Fast $k$-selection algorithms for graphics processing units.
\newblock \emph{ACM Journal of Experimental Algorithmics}, 17, 2012.
\newblock \doi{10.1145/2133803.2345676}.

\bibitem[Arora et~al.(2024)Arora, Eyuboglu, Zhang, Timalsina, Alberti, Zinsley, Zou, Rudra, and R{\'e}]{arora2024simple}
Arora, S., Eyuboglu, S., Zhang, M., Timalsina, A., Alberti, S., Zinsley, D., Zou, J., Rudra, A., and R{\'e}, C.
\newblock Simple linear attention language models balance the recall-throughput tradeoff.
\newblock In \emph{Proceedings of the 41st International Conference on Machine Learning}, 2024.

\bibitem[Bai et~al.(2025)Bai, Tu, Zhang, Peng, Wang, Lv, Cao, Xu, Hou, Dong, Tang, and Li]{bai2025longbench}
Bai, Y., Tu, S., Zhang, J., Peng, H., Wang, X., Lv, X., Cao, S., Xu, J., Hou, L., Dong, Y., Tang, J., and Li, J.
\newblock {LongBench} v2: Towards deeper understanding and reasoning on realistic long-context multitasks.
\newblock In \emph{Proceedings of the 63rd Annual Meeting of the Association for Computational Linguistics}, 2025.

\bibitem[Bai et~al.(2026)Bai, Dong, Jiang, Lv, Du, Zeng, Tang, and Li]{bai2026indexcache}
Bai, Y., Dong, Q., Jiang, T., Lv, X., Du, Z., Zeng, A., Tang, J., and Li, J.
\newblock {IndexCache}: Accelerating sparse attention via cross-layer index reuse.
\newblock In \emph{Third Conference on Language Modeling}, 2026.
\newblock URL \url{https://openreview.net/forum?id=74886M5htU}.

\bibitem[Beltagy et~al.(2020)Beltagy, Peters, and Cohan]{beltagy2020longformer}
Beltagy, I., Peters, M.~E., and Cohan, A.
\newblock Longformer: The long-document transformer.
\newblock \emph{arXiv preprint arXiv:2004.05150}, 2020.

\bibitem[Bhatia et~al.(2025)Bhatia, More, Borkar, Mitra, Matas, Zhao, Golub, Mudigere, Pharris, and Darvish~Rouhani]{bhatia2025helix}
Bhatia, N., More, A., Borkar, R., Mitra, T., Matas, R., Zhao, R., Golub, M., Mudigere, D., Pharris, B., and Darvish~Rouhani, B.
\newblock Helix parallelism: Rethinking sharding strategies for interactive multi-million-token {LLM} decoding.
\newblock \emph{arXiv preprint arXiv:2507.07120}, 2025.

\bibitem[Brandon et~al.(2023)Brandon, Nrusimha, Qian, Ankner, Jin, Song, and Ragan-Kelley]{brandon2023striped}
Brandon, W., Nrusimha, A., Qian, K., Ankner, Z., Jin, T., Song, Z., and Ragan-Kelley, J.
\newblock Striped attention: Faster ring attention for causal transformers.
\newblock \emph{arXiv preprint arXiv:2311.09431}, 2023.

\bibitem[Chen et~al.(2026)Chen, Zhang, Lu, Zhang, Zhang, Liu, Luo, Liu, Jiang, Chen, Ding, Yan, Jiang, Chen, Zhang, Li, Yang, Yang, and Yang]{chen2026retroinfer}
Chen, Y., Zhang, J., Lu, B., Zhang, Q., Zhang, C., Liu, J., Luo, J., Liu, D., Jiang, H., Chen, Q., Ding, B., Yan, X., Jiang, J., Chen, C., Zhang, M., Li, C., Yang, Y., Yang, F., and Yang, M.
\newblock {RetroInfer}: A vector storage engine for scalable long-context {LLM} inference.
\newblock \emph{Proceedings of the VLDB Endowment}, 19\penalty0 (5):\penalty0 1016--1031, 2026.
\newblock \doi{10.14778/3796195.3796212}.

\bibitem[Child et~al.(2019)Child, Gray, Radford, and Sutskever]{child2019generating}
Child, R., Gray, S., Radford, A., and Sutskever, I.
\newblock Generating long sequences with sparse transformers.
\newblock \emph{arXiv preprint arXiv:1904.10509}, 2019.

\bibitem[Clark et~al.(2018)Clark, Cowhey, Etzioni, Khot, Sabharwal, Schoenick, and Tafjord]{clark2018think}
Clark, P., Cowhey, I., Etzioni, O., Khot, T., Sabharwal, A., Schoenick, C., and Tafjord, O.
\newblock Think you have solved question answering? {T}ry {ARC}, the {AI2} reasoning challenge.
\newblock \emph{arXiv preprint arXiv:1803.05457}, 2018.

\bibitem[Dao et~al.(2022)Dao, Fu, Ermon, Rudra, and R\'{e}]{dao2022flashattention}
Dao, T., Fu, D., Ermon, S., Rudra, A., and R\'{e}, C.
\newblock {FlashAttention}: Fast and memory-efficient exact attention with {IO}-awareness.
\newblock In \emph{Advances in Neural Information Processing Systems}, volume~35, pp.\  16344--16359. Curran Associates, Inc., 2022.
\newblock \doi{10.52202/068431-1189}.

\bibitem[Dao et~al.(2023)Dao, Haziza, Massa, and Sizov]{dao2023flashdecoding}
Dao, T., Haziza, D., Massa, F., and Sizov, G.
\newblock {Flash-Decoding} for long-context inference, 2023.
\newblock URL \url{https://pytorch.org/blog/flash-decoding/}.

\bibitem[{DeepSeek-AI}(2024)]{deepseekai2024deepseekv3}
{DeepSeek-AI}.
\newblock {DeepSeek-V3} technical report.
\newblock \emph{arXiv preprint arXiv:2412.19437}, 2024.

\bibitem[{DeepSeek-AI}(2025)]{deepseekai2025deepseekv32}
{DeepSeek-AI}.
\newblock {DeepSeek-V3.2}: Pushing the frontier of open large language models.
\newblock \emph{arXiv preprint arXiv:2512.02556}, 2025.

\bibitem[{DeepSeek-AI}(2026{\natexlab{a}})]{deepseekai2026deepseekv4}
{DeepSeek-AI}.
\newblock {DeepSeek-V4}: Towards highly efficient million-token context intelligence.
\newblock \emph{arXiv preprint arXiv:2606.19348}, 2026{\natexlab{a}}.

\bibitem[{DeepSeek-AI}(2026{\natexlab{b}})]{deepseekai2026deepseekv41flash}
{DeepSeek-AI}.
\newblock {DeepSeek-V4.1-Flash}: Pushing the limits of {KV} cache compression.
\newblock \emph{arXiv preprint arXiv:2609.19969}, 2026{\natexlab{b}}.

\bibitem[Desai et~al.(2025)Desai, Yang, Cuadron, Zaharia, Gonzalez, and Stoica]{desai2025hashattention}
Desai, A., Yang, S., Cuadron, A., Zaharia, M., Gonzalez, J.~E., and Stoica, I.
\newblock {HashAttention}: Semantic sparsity for faster inference.
\newblock In \emph{Proceedings of the 42nd International Conference on Machine Learning}, volume 267 of \emph{Proceedings of Machine Learning Research}, pp.\  13402--13418. PMLR, 2025.

\bibitem[Floyd \& Rivest(1975)Floyd and Rivest]{floyd1975expected}
Floyd, R.~W. and Rivest, R.~L.
\newblock Expected time bounds for selection.
\newblock \emph{Communications of the ACM}, 18\penalty0 (3):\penalty0 165--172, 1975.
\newblock \doi{10.1145/360680.360691}.

\bibitem[Gao et~al.(2025)Gao, Zeng, Du, Cao, Zhou, Qi, Lai, So, Cao, Yang, and Yang]{gao2025seerattention}
Gao, Y., Zeng, Z., Du, D., Cao, S., Zhou, P., Qi, J., Lai, J., So, H., Cao, T., Yang, F., and Yang, M.
\newblock {SeerAttention}: Self-distilled attention gating for efficient long-context prefilling.
\newblock In \emph{Advances in Neural Information Processing Systems}, volume~38, pp.\  62230--62253. Curran Associates, Inc., 2025.
\newblock \doi{10.52202/085713-1869}.

\bibitem[Gao et~al.(2026)Gao, Guo, Cao, Xia, Cheng, Wang, Ma, Sun, Ye, Dong, So, Hua, Cao, Yang, and Yang]{gao2026sparse}
Gao, Y., Guo, S., Cao, S., Xia, Y., Cheng, Y., Wang, L., Ma, L., Sun, Y., Ye, T., Dong, L., So, H. K.-H., Hua, Y., Cao, T., Yang, F., and Yang, M.
\newblock Sparse attention adaptation for long reasoning.
\newblock In \emph{The Fourteenth International Conference on Learning Representations}, 2026.
\newblock URL \url{https://openreview.net/forum?id=c5BOcHM6J8}.

\bibitem[{Gemini Team}(2024)]{gemini2024gemini}
{Gemini Team}.
\newblock {Gemini 1.5}: Unlocking multimodal understanding across millions of tokens of context.
\newblock \emph{arXiv preprint arXiv:2403.05530}, 2024.

\bibitem[Gloeckle et~al.(2024)Gloeckle, Youbi~Idrissi, Rozi{\`e}re, Lopez-Paz, and Synnaeve]{gloeckle2024better}
Gloeckle, F., Youbi~Idrissi, B., Rozi{\`e}re, B., Lopez-Paz, D., and Synnaeve, G.
\newblock Better \& faster large language models via multi-token prediction.
\newblock In \emph{Proceedings of the 41st International Conference on Machine Learning}, 2024.

\bibitem[Hoefler et~al.(2021)Hoefler, Alistarh, Ben-Nun, Dryden, and Peste]{hoefler2021sparsity}
Hoefler, T., Alistarh, D., Ben-Nun, T., Dryden, N., and Peste, A.
\newblock Sparsity in deep learning: Pruning and growth for efficient inference and training in neural networks.
\newblock \emph{Journal of Machine Learning Research}, 22\penalty0 (241):\penalty0 1--124, 2021.
\newblock URL \url{http://jmlr.org/papers/v22/21-0366.html}.

\bibitem[Hooper et~al.(2025{\natexlab{a}})Hooper, Zhao, Manolache, Kim, Mahoney, Shao, Keutzer, and Gholami]{hooper2025multipole}
Hooper, C., Zhao, S., Manolache, L., Kim, S., Mahoney, M., Shao, S., Keutzer, K., and Gholami, A.
\newblock Multipole attention for efficient long context reasoning.
\newblock In \emph{Advances in Neural Information Processing Systems}, volume~38, pp.\  41954--41977. Curran Associates, Inc., 2025{\natexlab{a}}.
\newblock \doi{10.52202/085713-1255}.

\bibitem[Hooper et~al.(2025{\natexlab{b}})Hooper, Kim, Mohammadzadeh, Maheswaran, Zhao, Paik, Mahoney, Keutzer, and Gholami]{hooper2025squeezed}
Hooper, C. R.~C., Kim, S., Mohammadzadeh, H., Maheswaran, M., Zhao, S., Paik, J., Mahoney, M.~W., Keutzer, K., and Gholami, A.
\newblock Squeezed attention: Accelerating long context length {LLM} inference.
\newblock In \emph{Proceedings of the 63rd Annual Meeting of the Association for Computational Linguistics (Volume 1: Long Papers)}, pp.\  32631--32652, Vienna, Austria, 2025{\natexlab{b}}. Association for Computational Linguistics.
\newblock \doi{10.18653/v1/2025.acl-long.1568}.

\bibitem[Hsieh et~al.(2024)Hsieh, Sun, Kriman, Acharya, Rekesh, Jia, Zhang, and Ginsburg]{hsieh2024ruler}
Hsieh, C.-P., Sun, S., Kriman, S., Acharya, S., Rekesh, D., Jia, F., Zhang, Y., and Ginsburg, B.
\newblock {RULER}: What's the real context size of your long-context language models?
\newblock In \emph{First Conference on Language Modeling}, 2024.

\bibitem[Hu et~al.(2024)Hu, Tu, Han, He, Cui, Long, Zheng, Fang, Huang, Zhao, Zhang, Thai, Zhang, Wang, Yao, Zhao, Zhou, Cai, Zhai, Ding, Jia, Zeng, Li, Liu, and Sun]{hu2024minicpmunveilingpotentialsmall}
Hu, S., Tu, Y., Han, X., He, C., Cui, G., Long, X., Zheng, Z., Fang, Y., Huang, Y., Zhao, W., Zhang, X., Thai, Z.~L., Zhang, K., Wang, C., Yao, Y., Zhao, C., Zhou, J., Cai, J., Zhai, Z., Ding, N., Jia, C., Zeng, G., Li, D., Liu, Z., and Sun, M.
\newblock {MiniCPM}: Unveiling the potential of small language models with scalable training strategies.
\newblock \emph{arXiv preprint arXiv:2404.06395}, 2024.

\bibitem[Hu et~al.(2025)Hu, Leng, Zhao, Tu, and Wu]{hu2025hardware}
Hu, X., Leng, J., Zhao, J., Tu, K., and Wu, W.
\newblock Hardware-aligned hierarchical sparse attention for efficient long-term memory access.
\newblock In \emph{Advances in Neural Information Processing Systems}, volume~38, pp.\  98788--98813. Curran Associates, Inc., 2025.
\newblock \doi{10.52202/085713-2974}.

\bibitem[Hu et~al.(2026{\natexlab{a}})Hu, Wei, Gu, Zhang, Liang, Li, Zhu, Wang, Han, Bai, Tu, Mi, and Liang]{hu2026hierarchical}
Hu, X., Wei, X., Gu, H., Zhang, M., Liang, T., Li, H., Zhu, L., Wang, Y., Han, S., Bai, Y., Tu, K., Mi, H., and Liang, L.
\newblock Hierarchical sparse attention done right: Toward infinite context modeling.
\newblock \emph{arXiv preprint arXiv:2607.02980}, 2026{\natexlab{a}}.

\bibitem[Hu et~al.(2026{\natexlab{b}})Hu, Zhou, Liang, Li, Wu, and Li]{hu2026every}
Hu, X., Zhou, Z., Liang, R., Li, Z., Wu, W., and Li, J.
\newblock Every token counts: Generalizing {16M} ultra-long context in large language models.
\newblock In \emph{Proceedings of the 64th Annual Meeting of the Association for Computational Linguistics (Volume 1: Long Papers)}, pp.\  10208--10220, San Diego, California, United States, 2026{\natexlab{b}}. Association for Computational Linguistics.
\newblock \doi{10.18653/v1/2026.acl-long.464}.

\bibitem[J{\'e}gou et~al.(2011)J{\'e}gou, Tavenard, Douze, and Amsaleg]{jegou2011searching}
J{\'e}gou, H., Tavenard, R., Douze, M., and Amsaleg, L.
\newblock Searching in one billion vectors: Re-rank with source coding.
\newblock In \emph{2011 IEEE International Conference on Acoustics, Speech and Signal Processing (ICASSP)}, pp.\  861--864. IEEE, 2011.
\newblock \doi{10.1109/ICASSP.2011.5946540}.

\bibitem[Jiang et~al.(2024)Jiang, Li, Zhang, Wu, Luo, Ahn, Han, Abdi, Li, Lin, Yang, and Qiu]{jiang2024minference}
Jiang, H., Li, Y., Zhang, C., Wu, Q., Luo, X., Ahn, S., Han, Z., Abdi, A.~H., Li, D., Lin, C.-Y., Yang, Y., and Qiu, L.
\newblock {MInference} 1.0: Accelerating pre-filling for long-context {LLMs} via dynamic sparse attention.
\newblock In \emph{Advances in Neural Information Processing Systems}, volume~37, pp.\  52481--52515. Curran Associates, Inc., 2024.
\newblock \doi{10.52202/079017-1663}.

\bibitem[Jimenez et~al.(2024)Jimenez, Yang, Wettig, Yao, Pei, Press, and Narasimhan]{jimenez2024swebench}
Jimenez, C.~E., Yang, J., Wettig, A., Yao, S., Pei, K., Press, O., and Narasimhan, K.
\newblock {SWE}-bench: Can language models resolve real-world {GitHub} issues?
\newblock In \emph{The Twelfth International Conference on Learning Representations}, 2024.
\newblock URL \url{https://openreview.net/forum?id=VTF8yNQM66}.

\bibitem[Johnson et~al.(2021)Johnson, Douze, and J{\'e}gou]{johnson2021billion}
Johnson, J., Douze, M., and J{\'e}gou, H.
\newblock Billion-scale similarity search with {GPUs}.
\newblock \emph{IEEE Transactions on Big Data}, 7\penalty0 (3):\penalty0 535--547, 2021.
\newblock \doi{10.1109/TBDATA.2019.2921572}.

\bibitem[Kandpal et~al.(2025)Kandpal, Lester, Raffel, Majstorovic, Biderman, Abbasi, Soldaini, Shippole, Cooper, Skowron, Kirchenbauer, Longpre, Sutawika, Albalak, Xu, Penedo, Ben~Allal, Bakouch, Pressman, Fan, Stander, Song, Gokaslan, Goldstein, Bartoldson, Kailkhura, and Murray]{kandpal2025common}
Kandpal, N., Lester, B., Raffel, C., Majstorovic, S., Biderman, S., Abbasi, B., Soldaini, L., Shippole, E., Cooper, A.~F., Skowron, A., Kirchenbauer, J., Longpre, S., Sutawika, L., Albalak, A., Xu, Z., Penedo, G., Ben~Allal, L., Bakouch, E., Pressman, J.~D., Fan, H., Stander, D., Song, G., Gokaslan, A., Goldstein, T., Bartoldson, B.~R., Kailkhura, B., and Murray, T.
\newblock The {Common Pile} v0.1: An 8{TB} dataset of public domain and openly licensed text.
\newblock \emph{arXiv preprint arXiv:2506.05209}, 2025.

\bibitem[Kitaev et~al.(2020)Kitaev, Kaiser, and Levskaya]{kitaev2020reformer}
Kitaev, N., Kaiser, {\L}., and Levskaya, A.
\newblock Reformer: The efficient transformer.
\newblock In \emph{The Eighth International Conference on Learning Representations}, 2020.
\newblock URL \url{https://openreview.net/forum?id=rkgNKkHtvB}.

\bibitem[Kocetkov et~al.(2023)Kocetkov, Li, Ben~Allal, Li, Mou, Mu{\~n}oz~Ferrandis, Jernite, Mitchell, Hughes, Wolf, Bahdanau, von Werra, and de~Vries]{kocetkov2023stack}
Kocetkov, D., Li, R., Ben~Allal, L., Li, J., Mou, C., Mu{\~n}oz~Ferrandis, C., Jernite, Y., Mitchell, M., Hughes, S., Wolf, T., Bahdanau, D., von Werra, L., and de~Vries, H.
\newblock The {Stack}: 3 {TB} of permissively licensed source code.
\newblock \emph{Transactions on Machine Learning Research}, 2023.

\bibitem[Kwa et~al.(2025)Kwa, West, Becker, Deng, Garcia, Hasin, Jawhar, Kinniment, Rush, Von~Arx, Bloom, Broadley, Du, Goodrich, Jurkovic, Miles, Nix, Lin, Parikh, Rein, Koba~Sato, Wijk, Ziegler, Barnes, and Chan]{kwa2025measuring}
Kwa, T., West, B., Becker, J., Deng, A., Garcia, K., Hasin, M., Jawhar, S., Kinniment, M., Rush, N., Von~Arx, S., Bloom, R., Broadley, T., Du, H., Goodrich, B., Jurkovic, N., Miles, L., Nix, S., Lin, T., Parikh, N., Rein, D., Koba~Sato, L.~J., Wijk, H., Ziegler, D., Barnes, E., and Chan, L.
\newblock Measuring {AI} ability to complete long software tasks.
\newblock In \emph{Advances in Neural Information Processing Systems}, volume~38, pp.\  102382--102435. Curran Associates, Inc., 2025.
\newblock \doi{10.52202/085713-3086}.

\bibitem[Lai et~al.(2026)Lai, Xu, Yang, Chen, Xu, Zeng, Li, Sun, Zhu, Zhang, Hu, Li, Gao, Li, Zhu, Zhou, and Zhao]{lai2026minimax}
Lai, X., Xu, W., Yang, Y., Chen, Q., Xu, Y., Zeng, L., Li, X., Sun, H., Zhu, H., Zhang, V., Hu, J., Li, J., Gao, R., Li, Z., Zhu, S., Zhou, J., and Zhao, P.
\newblock {MiniMax} sparse attention.
\newblock \emph{arXiv preprint arXiv:2606.13392}, 2026.

\bibitem[Lee et~al.(2025{\natexlab{a}})Lee, Park, Lee, Suh, Kim, Jeong, Kim, Lee, Jeon, and Hwang]{lee2025training}
Lee, H., Park, G., Lee, Y., Suh, J., Kim, J., Jeong, W., Kim, B., Lee, H., Jeon, M., and Hwang, S.~J.
\newblock A training-free sub-quadratic cost transformer model serving framework with hierarchically pruned attention.
\newblock In \emph{The Thirteenth International Conference on Learning Representations}, 2025{\natexlab{a}}.
\newblock URL \url{https://openreview.net/forum?id=PTcMzQgKmn}.

\bibitem[Lee et~al.(2025{\natexlab{b}})Lee, Park, Suh, and Hwang]{lee2025infinitehip}
Lee, H., Park, G., Suh, J., and Hwang, S.~J.
\newblock {InfiniteHiP}: Extending language model context up to 3 million tokens on a single {GPU}.
\newblock \emph{arXiv preprint arXiv:2502.08910}, 2025{\natexlab{b}}.

\bibitem[Leng et~al.(2026)Leng, Hu, Wang, Li, Wu, and Lu]{leng2026understanding}
Leng, J., Hu, X., Wang, J., Li, J., Wu, W., and Lu, Y.
\newblock Understanding and improving length generalization in hierarchical sparse attention models.
\newblock In \emph{The Fourteenth International Conference on Learning Representations}, 2026.
\newblock URL \url{https://openreview.net/forum?id=iHqdSQk6qc}.

\bibitem[Li et~al.(2026)Li, Liu, Lin, Hu, and Zhang]{li2026lycheecluster}
Li, D., Liu, Z., Lin, G., Hu, B., and Zhang, M.
\newblock {LycheeCluster}: Efficient long-context inference with structure-aware chunking and hierarchical {KV} indexing.
\newblock In \emph{Findings of the Association for Computational Linguistics: ACL 2026}, pp.\  7607--7623, San Diego, California, United States, 2026. Association for Computational Linguistics.
\newblock \doi{10.18653/v1/2026.findings-acl.376}.

\bibitem[Li et~al.(2025)Li, Liu, and Sun]{flashmla2025}
Li, J., Liu, S., and Sun, Y.
\newblock {FlashMLA}: Efficient multi-head latent attention kernels, 2025.
\newblock URL \url{https://github.com/deepseek-ai/FlashMLA}.

\bibitem[Li et~al.(2024)Li, Huang, Yang, Venkitesh, Locatelli, Ye, Cai, Lewis, and Chen]{li2024snapkv}
Li, Y., Huang, Y., Yang, B., Venkitesh, B., Locatelli, A., Ye, H., Cai, T., Lewis, P., and Chen, D.
\newblock {SnapKV}: {LLM} knows what you are looking for before generation.
\newblock In \emph{Advances in Neural Information Processing Systems}, volume~37, pp.\  22947--22970. Curran Associates, Inc., 2024.
\newblock \doi{10.52202/079017-0722}.

\bibitem[Lightman et~al.(2024)Lightman, Kosaraju, Burda, Edwards, Baker, Lee, Leike, Schulman, Sutskever, and Cobbe]{lightman2024lets}
Lightman, H., Kosaraju, V., Burda, Y., Edwards, H., Baker, B., Lee, T., Leike, J., Schulman, J., Sutskever, I., and Cobbe, K.
\newblock Let's verify step by step.
\newblock In \emph{The Twelfth International Conference on Learning Representations}, 2024.

\bibitem[Liu et~al.(2025{\natexlab{a}})Liu, Chen, Lu, Jiang, Han, Zhang, Chen, Zhang, Ding, Zhang, Chen, Yang, Yang, and Qiu]{liu2025retrievalattention}
Liu, D., Chen, M., Lu, B., Jiang, H., Han, Z., Zhang, Q., Chen, Q., Zhang, C., Ding, B., Zhang, K., Chen, C., Yang, F., Yang, Y., and Qiu, L.
\newblock {RetrievalAttention}: Accelerating long-context {LLM} inference via vector retrieval.
\newblock In \emph{Advances in Neural Information Processing Systems}, volume~38, pp.\  60539--60566. Curran Associates, Inc., 2025{\natexlab{a}}.
\newblock \doi{10.52202/085713-1816}.

\bibitem[Liu et~al.(2025{\natexlab{b}})Liu, Li, Zhao, Zhang, and Guo]{liu2025clusterkv}
Liu, G., Li, C., Zhao, J., Zhang, C., and Guo, M.
\newblock {ClusterKV}: Manipulating {LLM} {KV} cache in semantic space for recallable compression.
\newblock In \emph{2025 62nd ACM/IEEE Design Automation Conference (DAC)}, pp.\  1--7. IEEE, 2025{\natexlab{b}}.
\newblock \doi{10.1109/DAC63849.2025.11132479}.

\bibitem[Liu et~al.(2024)Liu, Zaharia, and Abbeel]{liu2024ring}
Liu, H., Zaharia, M., and Abbeel, P.
\newblock Ring attention with blockwise transformers for near-infinite context.
\newblock In \emph{The Twelfth International Conference on Learning Representations}, 2024.

\bibitem[Lu et~al.(2026)Lu, Lu, Lange, Yamada, Hu, Foerster, Ha, and Clune]{lu2026towards}
Lu, C., Lu, C., Lange, R.~T., Yamada, Y., Hu, S., Foerster, J., Ha, D., and Clune, J.
\newblock Towards end-to-end automation of {AI} research.
\newblock \emph{Nature}, 651\penalty0 (8107):\penalty0 914--919, 2026.
\newblock \doi{10.1038/s41586-026-10265-5}.

\bibitem[Lu et~al.(2025)Lu, Jiang, Liu, Du, Jiang, Hong, Liu, He, Yuan, Wang, Huang, Yuan, Xu, Xu, Lai, Chen, Zheng, Yan, Su, Wu, Zhang, Yang, Zhou, Zhang, and Qiu]{lu2025moba}
Lu, E., Jiang, Z., Liu, J., Du, Y., Jiang, T., Hong, C., Liu, S., He, W., Yuan, E., Wang, Y., Huang, Z., Yuan, H., Xu, S., Xu, X., Lai, G., Chen, Y., Zheng, H., Yan, J., Su, J., Wu, Y., Zhang, Y., Yang, Z., Zhou, X., Zhang, M., and Qiu, J.
\newblock {MoBA}: Mixture of block attention for long-context {LLMs}.
\newblock In \emph{Advances in Neural Information Processing Systems}, volume~38, pp.\  20267--20292. Curran Associates, Inc., 2025.
\newblock \doi{10.52202/085713-0602}.

\bibitem[Madaan et~al.(2023)Madaan, Bhojanapalli, Jain, and Jain]{madaan2023treeformer}
Madaan, L., Bhojanapalli, S., Jain, H., and Jain, P.
\newblock Treeformer: Dense gradient trees for efficient attention computation.
\newblock In \emph{The Eleventh International Conference on Learning Representations}, 2023.
\newblock URL \url{https://openreview.net/forum?id=DWn1TEb2fK}.

\bibitem[Mao et~al.(2026{\natexlab{a}})Mao, Li, and Fox]{mao2026simplified}
Mao, Y., Li, M.~Y., and Fox, E.~B.
\newblock Simplified sparse attention via gist tokens.
\newblock \emph{arXiv preprint arXiv:2604.20920}, 2026{\natexlab{a}}.

\bibitem[Mao et~al.(2026{\natexlab{b}})Mao, Wang, Ester, and Li]{mao2026icecache}
Mao, Y., Wang, Q., Ester, M., and Li, K.
\newblock {IceCache}: Memory-efficient {KV}-cache management for long-sequence {LLMs}.
\newblock In \emph{The Fourteenth International Conference on Learning Representations}, 2026{\natexlab{b}}.
\newblock URL \url{https://openreview.net/forum?id=yHxSKM9kdr}.

\bibitem[{MiniMax}(2025)]{minimax2025minimax01}
{MiniMax}.
\newblock {MiniMax-01}: Scaling foundation models with lightning attention.
\newblock \emph{arXiv preprint arXiv:2501.08313}, 2025.

\bibitem[{NVIDIA}(2026{\natexlab{a}})]{nvidia2026blackwellultra}
{NVIDIA}.
\newblock Inside {NVIDIA Blackwell Ultra}: The chip powering the {AI} factory era.
\newblock \url{https://developer.nvidia.com/blog/inside-nvidia-blackwell-ultra-the-chip-powering-the-ai-factory-era/}, 2026{\natexlab{a}}.
\newblock Accessed 2026-10-28.

\bibitem[{NVIDIA}(2026{\natexlab{b}})]{nvidia2026computecapabilities}
{NVIDIA}.
\newblock {CUDA} programming guide, appendix: Compute capabilities.
\newblock \url{https://docs.nvidia.com/cuda/cuda-programming-guide/05-appendices/compute-capabilities.html}, 2026{\natexlab{b}}.
\newblock Accessed 2026-10-28.

\bibitem[{NVIDIA}(2026{\natexlab{c}})]{nvidia2026h200}
{NVIDIA}.
\newblock {NVIDIA H200} tensor core {GPU}.
\newblock \url{https://www.nvidia.com/en-us/data-center/h200/}, 2026{\natexlab{c}}.
\newblock Accessed 2026-10-28.

\bibitem[{NVIDIA}(2026{\natexlab{d}})]{nvidia2026hgx}
{NVIDIA}.
\newblock {NVIDIA HGX} platform.
\newblock \url{https://www.nvidia.com/en-us/data-center/hgx/}, 2026{\natexlab{d}}.
\newblock Accessed 2026-10-28.

\bibitem[{NVIDIA}(2026{\natexlab{e}})]{nvidia2026openswe}
{NVIDIA}.
\newblock {Open-SWE-Traces}.
\newblock \url{https://huggingface.co/datasets/nvidia/Open-SWE-Traces}, 2026{\natexlab{e}}.
\newblock Accessed 2026-10-30.

\bibitem[{NVIDIA}(2026{\natexlab{f}})]{nvidia2026pdl}
{NVIDIA}.
\newblock {CUDA} programming guide, section 4.5: Programmatic dependent launch and synchronization.
\newblock \url{https://docs.nvidia.com/cuda/cuda-programming-guide/04-special-topics/programmatic-dependent-launch.html}, 2026{\natexlab{f}}.
\newblock Accessed 2026-10-30.

\bibitem[Penedo et~al.(2024)Penedo, Kydl{\'\i}{\v{c}}ek, Ben~Allal, Lozhkov, Mitchell, Raffel, Von~Werra, and Wolf]{penedo2024fineweb}
Penedo, G., Kydl{\'\i}{\v{c}}ek, H., Ben~Allal, L., Lozhkov, A., Mitchell, M., Raffel, C., Von~Werra, L., and Wolf, T.
\newblock The {FineWeb} datasets: Decanting the web for the finest text data at scale.
\newblock In \emph{Advances in Neural Information Processing Systems (Datasets and Benchmarks Track)}, 2024.

\bibitem[Peng et~al.(2024)Peng, Quesnelle, Fan, and Shippole]{peng2024yarn}
Peng, B., Quesnelle, J., Fan, H., and Shippole, E.
\newblock {YaRN}: Efficient context window extension of large language models.
\newblock In \emph{The Twelfth International Conference on Learning Representations}, 2024.

\bibitem[{Qwen Team}(2026{\natexlab{a}})]{qwen2026design}
{Qwen Team}.
\newblock On the design of {Qwen3.8-Next} architecture: Evaluation, efficiency, and training stability.
\newblock \emph{arXiv preprint arXiv:2608.30320}, 2026{\natexlab{a}}.

\bibitem[{Qwen Team}(2026{\natexlab{b}})]{qwen38}
{Qwen Team}.
\newblock {Qwen3.8-Max}: A new bar for coding and cowork, August 2026{\natexlab{b}}.
\newblock URL \url{https://qwen.ai/blog?id=qwen3.8}.

\bibitem[{Qwen Team}(2026{\natexlab{c}})]{qwen38_27b}
{Qwen Team}.
\newblock {Qwen3.8-27B}.
\newblock Hugging Face model card, \url{https://huggingface.co/Qwen/Qwen3.8-27B}, 2026{\natexlab{c}}.
\newblock Accessed 2026-10-30.

\bibitem[Rein et~al.(2024)Rein, Hou, Stickland, Petty, Pang, Dirani, Michael, and Bowman]{rein2024gpqa}
Rein, D., Hou, B.~L., Stickland, A.~C., Petty, J., Pang, R.~Y., Dirani, J., Michael, J., and Bowman, S.~R.
\newblock {GPQA}: A graduate-level {Google}-proof {Q\&A} benchmark.
\newblock In \emph{First Conference on Language Modeling}, 2024.

\bibitem[Roy et~al.(2021)Roy, Saffar, Vaswani, and Grangier]{roy2021efficient}
Roy, A., Saffar, M., Vaswani, A., and Grangier, D.
\newblock Efficient content-based sparse attention with routing transformers.
\newblock \emph{Transactions of the Association for Computational Linguistics}, 9:\penalty0 53--68, 2021.
\newblock \doi{10.1162/tacl_a_00353}.

\bibitem[Schlag et~al.(2021)Schlag, Irie, and Schmidhuber]{schlag2021linear}
Schlag, I., Irie, K., and Schmidhuber, J.
\newblock Linear transformers are secretly fast weight programmers.
\newblock In \emph{Proceedings of the 38th International Conference on Machine Learning}, pp.\  9355--9366, 2021.

\bibitem[Schmidhuber(1992{\natexlab{a}})]{schmidhuber1992fastweight}
Schmidhuber, J.
\newblock Learning to control fast-weight memories: An alternative to dynamic recurrent networks.
\newblock \emph{Neural Computation}, 4\penalty0 (1):\penalty0 131--139, 1992{\natexlab{a}}.
\newblock \doi{10.1162/neco.1992.4.1.131}.

\bibitem[Schmidhuber(1992{\natexlab{b}})]{schmidhuber1992history}
Schmidhuber, J.
\newblock Learning complex, extended sequences using the principle of history compression.
\newblock \emph{Neural Computation}, 4\penalty0 (2):\penalty0 234--242, 1992{\natexlab{b}}.
\newblock \doi{10.1162/neco.1992.4.2.234}.

\bibitem[Shah et~al.(2024)Shah, Bikshandi, Zhang, Thakkar, Ramani, and Dao]{shah2024flashattention}
Shah, J., Bikshandi, G., Zhang, Y., Thakkar, V., Ramani, P., and Dao, T.
\newblock {FlashAttention-3}: Fast and accurate attention with asynchrony and low-precision.
\newblock In \emph{Advances in Neural Information Processing Systems}, volume~37, pp.\  68658--68685. Curran Associates, Inc., 2024.
\newblock \doi{10.52202/079017-2193}.

\bibitem[Su et~al.(2024)Su, Ahmed, Lu, Pan, Bo, and Liu]{su2024roformer}
Su, J., Ahmed, M., Lu, Y., Pan, S., Bo, W., and Liu, Y.
\newblock {RoFormer}: Enhanced transformer with rotary position embedding.
\newblock \emph{Neurocomputing}, 568:\penalty0 127063, 2024.

\bibitem[Tang et~al.(2026)Tang, Qin, Pan, Li, and Liu]{tang2026block}
Tang, B., Qin, Z., Pan, Y., Li, Z., and Liu, P.
\newblock Block sparse attention with log-linear complexity.
\newblock \emph{arXiv preprint arXiv:2609.31093}, 2026.

\bibitem[Tang et~al.(2024)Tang, Zhao, Zhu, Xiao, Kasikci, and Han]{tang2024quest}
Tang, J., Zhao, Y., Zhu, K., Xiao, G., Kasikci, B., and Han, S.
\newblock {QUEST}: Query-aware sparsity for efficient long-context {LLM} inference.
\newblock In \emph{Proceedings of the 41st International Conference on Machine Learning}, volume 235 of \emph{Proceedings of Machine Learning Research}, pp.\  47901--47911. PMLR, 2024.

\bibitem[Thakkar et~al.(2023)Thakkar, Ramani, Cecka, Shivam, Lu, Yan, Kosaian, Hoemmen, Wu, Kerr, Nicely, Merrill, Blasig, Atluri, Qiao, Majcher, Springer, Hohnerbach, Wang, and Gupta]{thakkar2023cutlass}
Thakkar, V., Ramani, P., Cecka, C., Shivam, A., Lu, H., Yan, E., Kosaian, J., Hoemmen, M., Wu, H., Kerr, A., Nicely, M., Merrill, D., Blasig, D., Atluri, A., Qiao, F., Majcher, P., Springer, P., Hohnerbach, M., Wang, J., and Gupta, M.
\newblock {CUTLASS}, 2023.
\newblock URL \url{https://github.com/NVIDIA/cutlass}.

\bibitem[Tillet et~al.(2019)Tillet, Kung, and Cox]{tillet2019triton}
Tillet, P., Kung, H.~T., and Cox, D.
\newblock {Triton}: An intermediate language and compiler for tiled neural network computations.
\newblock In \emph{Proceedings of the 3rd ACM SIGPLAN International Workshop on Machine Learning and Programming Languages}, pp.\  10--19. Association for Computing Machinery, 2019.
\newblock \doi{10.1145/3315508.3329973}.

\bibitem[Vaswani et~al.(2017)Vaswani, Shazeer, Parmar, Uszkoreit, Jones, Gomez, Kaiser, and Polosukhin]{vaswani2017attention}
Vaswani, A., Shazeer, N., Parmar, N., Uszkoreit, J., Jones, L., Gomez, A.~N., Kaiser, {\L}., and Polosukhin, I.
\newblock Attention is all you need.
\newblock In \emph{Advances in Neural Information Processing Systems}, volume~30. Curran Associates, Inc., 2017.

\bibitem[Wang et~al.(2025)Wang, Cheng, Shi, Tang, Mo, Xie, Ma, Xia, Xue, Yang, and Yang]{wang2025tilelang}
Wang, L., Cheng, Y., Shi, Y., Tang, Z., Mo, Z., Xie, W., Ma, L., Xia, Y., Xue, J., Yang, F., and Yang, Z.
\newblock {TileLang}: A composable tiled programming model for {AI} systems.
\newblock \emph{arXiv preprint arXiv:2504.17577}, 2025.

\bibitem[Wang et~al.(2024)Wang, Ma, Zhang, Ni, Chandra, Guo, Ren, Arulraj, He, Jiang, Li, Ku, Wang, Zhuang, Fan, Yue, and Chen]{wang2024mmlupro}
Wang, Y., Ma, X., Zhang, G., Ni, Y., Chandra, A., Guo, S., Ren, W., Arulraj, A., He, X., Jiang, Z., Li, T., Ku, M., Wang, K., Zhuang, A., Fan, R., Yue, X., and Chen, W.
\newblock {MMLU-Pro}: A more robust and challenging multi-task language understanding benchmark.
\newblock In \emph{Advances in Neural Information Processing Systems (Datasets and Benchmarks Track)}, 2024.

\bibitem[Ward(1963)]{ward1963hierarchical}
Ward, Jr., J.~H.
\newblock Hierarchical grouping to optimize an objective function.
\newblock \emph{Journal of the American Statistical Association}, 58\penalty0 (301):\penalty0 236--244, 1963.
\newblock \doi{10.1080/01621459.1963.10500845}.

\bibitem[Xiao et~al.(2024{\natexlab{a}})Xiao, Zhang, Han, Xiao, Lin, Zhang, Liu, and Sun]{xiao2024infllm}
Xiao, C., Zhang, P., Han, X., Xiao, G., Lin, Y., Zhang, Z., Liu, Z., and Sun, M.
\newblock {InfLLM}: Training-free long-context extrapolation for {LLMs} with an efficient context memory.
\newblock In \emph{Advances in Neural Information Processing Systems}, volume~37, pp.\  119638--119661. Curran Associates, Inc., 2024{\natexlab{a}}.
\newblock \doi{10.52202/079017-3801}.

\bibitem[Xiao et~al.(2024{\natexlab{b}})Xiao, Tian, Chen, Han, and Lewis]{xiao2024efficient}
Xiao, G., Tian, Y., Chen, B., Han, S., and Lewis, M.
\newblock Efficient streaming language models with attention sinks.
\newblock In \emph{The Twelfth International Conference on Learning Representations}, 2024{\natexlab{b}}.
\newblock URL \url{https://openreview.net/forum?id=NG7sS51zVF}.

\bibitem[Xu et~al.(2026)Xu, Meng, Jiang, Wang, Zhou, Wang, Wu, Pan, Tang, Pei, Liu, Yin, Sun, and Zhang]{xu2026hisa}
Xu, Y., Meng, F., Jiang, F., Wang, Y., Zhou, R., Wang, Z., Wu, J., Pan, Z., Tang, X., Pei, W., Liu, T., Yin, D., Sun, X., and Zhang, M.
\newblock {HISA}: Efficient hierarchical indexing for fine-grained sparse attention.
\newblock In \emph{Third Conference on Language Modeling}, 2026.
\newblock URL \url{https://openreview.net/forum?id=GnWqiwzNzK}.

\bibitem[Yang et~al.(2025{\natexlab{a}})Yang, Yang, Ibrahim, Xie, Tang, Sizov, Park, and Huang]{yang2025context}
Yang, A., Yang, J., Ibrahim, A., Xie, X., Tang, B., Sizov, G., Park, J., and Huang, J.
\newblock Context parallelism for scalable million-token inference.
\newblock In \emph{Proceedings of Machine Learning and Systems}, volume~7. MLSys, 2025{\natexlab{a}}.

\bibitem[Yang et~al.(2025{\natexlab{b}})Yang, Kautz, and Hatamizadeh]{yang2025gated}
Yang, S., Kautz, J., and Hatamizadeh, A.
\newblock Gated delta networks: Improving {Mamba2} with delta rule.
\newblock In \emph{The Thirteenth International Conference on Learning Representations}, 2025{\natexlab{b}}.

\bibitem[Ye et~al.(2025)Ye, Chen, Lai, Lin, Zhang, Wang, Chen, Kasikci, Grover, Krishnamurthy, and Ceze]{ye2025flashinfer}
Ye, Z., Chen, L., Lai, R., Lin, W., Zhang, Y., Wang, S., Chen, T., Kasikci, B., Grover, V., Krishnamurthy, A., and Ceze, L.
\newblock {FlashInfer}: Efficient and customizable attention engine for {LLM} inference serving.
\newblock In \emph{Proceedings of Machine Learning and Systems}, volume~7, 2025.

\bibitem[Yuan et~al.(2025)Yuan, Gao, Dai, Luo, Zhao, Zhang, Xie, Wei, Wang, Xiao, Wang, Ruan, Zhang, Liang, and Zeng]{yuan2025native}
Yuan, J., Gao, H., Dai, D., Luo, J., Zhao, L., Zhang, Z., Xie, Z., Wei, Y., Wang, L., Xiao, Z., Wang, Y., Ruan, C., Zhang, M., Liang, W., and Zeng, W.
\newblock Native sparse attention: Hardware-aligned and natively trainable sparse attention.
\newblock In \emph{Proceedings of the 63rd Annual Meeting of the Association for Computational Linguistics (Volume 1: Long Papers)}, pp.\  23078--23097, Vienna, Austria, 2025. Association for Computational Linguistics.
\newblock \doi{10.18653/v1/2025.acl-long.1126}.

\bibitem[Yuan et~al.(2026)Yuan, Shinn, Xu, Cui, Klimiashvili, Xiao, Zheng, Li, Zhou, Ye, You, Zheng, Brown, Wang, Hoehnerbach, Cai, Demouth, Owens, Hu, Han, Liu, and Mao]{yuan2026blasst}
Yuan, J., Shinn, C., Xu, K., Cui, J., Klimiashvili, G., Xiao, G., Zheng, P., Li, B., Zhou, Y., Ye, Z., You, W., Zheng, T., Brown, D., Wang, P., Hoehnerbach, M., Cai, R., Demouth, J., Owens, J.~D., Hu, X., Han, S., Liu, T., and Mao, H.
\newblock {BLASST}: Dynamic {BL}ocked attention sparsity via softmax thresholding.
\newblock In \emph{Proceedings of Machine Learning and Systems}, volume~8, pp.\  843--859. MLSys, 2026.

\bibitem[Zaheer et~al.(2020)Zaheer, Guruganesh, Dubey, Ainslie, Alberti, Ontanon, Pham, Ravula, Wang, Yang, and Ahmed]{zaheer2020big}
Zaheer, M., Guruganesh, G., Dubey, K.~A., Ainslie, J., Alberti, C., Ontanon, S., Pham, P., Ravula, A., Wang, Q., Yang, L., and Ahmed, A.
\newblock Big bird: Transformers for longer sequences.
\newblock In \emph{Advances in Neural Information Processing Systems}, volume~33, pp.\  17283--17297. Curran Associates, Inc., 2020.

\bibitem[{Z.ai}(2026)]{zai2026glm53flash}
{Z.ai}.
\newblock {GLM-5.3-Flash}, 2026.
\newblock URL \url{https://z.ai/blog/glm-5.3-flash}.

\bibitem[Zan et~al.(2026)Zan, Zhang, Tan, Liu, Wang, Li, Ma, Wu, Lu, Li, Sun, Pei, Xie, and Cai]{zan2026longcat}
Zan, W., Zhang, J., Tan, J., Liu, H., Wang, C., Li, X., Ma, D., Wu, G., Lu, Y., Li, F., Sun, Y., Pei, P., Xie, Y., and Cai, X.
\newblock {LongCat} sparse attention: Taming the lightning via streaming-aware hierarchical cross-layer indexing.
\newblock \emph{arXiv preprint arXiv:2608.01662}, 2026.

\bibitem[Zellers et~al.(2019)Zellers, Holtzman, Bisk, Farhadi, and Choi]{zellers2019hellaswag}
Zellers, R., Holtzman, A., Bisk, Y., Farhadi, A., and Choi, Y.
\newblock {HellaSwag}: Can a machine really finish your sentence?
\newblock In \emph{Proceedings of the 57th Annual Meeting of the Association for Computational Linguistics}, pp.\  4791--4800, 2019.

\bibitem[Zhang et~al.(2023{\natexlab{a}})Zhang, Naruse, Li, and Wang]{zhang2023parallel}
Zhang, J., Naruse, A., Li, X., and Wang, Y.
\newblock Parallel top-{K} algorithms on {GPU}: A comprehensive study and new methods.
\newblock In \emph{Proceedings of the International Conference for High Performance Computing, Networking, Storage and Analysis}, pp.\  1--13. ACM, 2023{\natexlab{a}}.
\newblock \doi{10.1145/3581784.3607062}.

\bibitem[Zhang et~al.(2023{\natexlab{b}})Zhang, Sheng, Zhou, Chen, Zheng, Cai, Song, Tian, R\'{e}, Barrett, Wang, and Chen]{zhang2023h2o}
Zhang, Z., Sheng, Y., Zhou, T., Chen, T., Zheng, L., Cai, R., Song, Z., Tian, Y., R\'{e}, C., Barrett, C., Wang, Z., and Chen, B.
\newblock {H2O}: Heavy-hitter oracle for efficient generative inference of large language models.
\newblock In \emph{Advances in Neural Information Processing Systems}, volume~36, pp.\  34661--34710. Curran Associates, Inc., 2023{\natexlab{b}}.
\newblock \doi{10.52202/075280-1506}.

\bibitem[Zhao et~al.(2026)Zhao, Zhou, Su, Xiao, Li, Li, Zhang, Zhao, Li, Huang, Sun, Han, and Liu]{zhao2026infllmv2}
Zhao, W., Zhou, Z., Su, Z., Xiao, C., Li, Y., Li, Y., Zhang, Y., Zhao, W., Li, Z., Huang, Y., Sun, A., Han, X., and Liu, Z.
\newblock {InfLLM-V2}: Dense-sparse switchable attention for seamless short-to-long adaptation.
\newblock In \emph{The Fourteenth International Conference on Learning Representations}, 2026.
\newblock URL \url{https://openreview.net/forum?id=ZzF9V0H6Vi}.

\bibitem[Zheng et~al.(2024)Zheng, Yin, Xie, Sun, Huang, Yu, Cao, Kozyrakis, Stoica, Gonzalez, Barrett, and Sheng]{zheng2024sglang}
Zheng, L., Yin, L., Xie, Z., Sun, C., Huang, J., Yu, C.~H., Cao, S., Kozyrakis, C., Stoica, I., Gonzalez, J.~E., Barrett, C., and Sheng, Y.
\newblock {SGLang}: Efficient execution of structured language model programs.
\newblock In \emph{Advances in Neural Information Processing Systems}, 2024.

\bibitem[Zhu et~al.(2026)Zhu, Zhou, Qin, Zhu, Ding, Zhong, Wen, Xie, Gou, Ren, Wang, Zhong, Liu, Gao, Lin, Zhang, Song, Qi, Wu, Zhang, Piao, Niu, Lin, Meng, Tang, Tang, Wu, Zheng, Liu, Zhu, Wang, Ding, Wan, Liu, Wang, Zhu, Zhang, Chai, Liu, Lai, Yuan, Su, Zhang, Zhou, Du, Huang, and Shi]{zhu2026edgebench}
Zhu, D., Zhou, X., Qin, S., Zhu, X., Ding, H., Zhong, S., Wen, Z., Xie, Z., Gou, C., Ren, L., Wang, Y., Zhong, J., Liu, R., Gao, T., Lin, Y., Zhang, J., Song, M., Qi, X., Wu, J., Zhang, C., Piao, Y., Niu, Z., Lin, H., Meng, L., Tang, P., Tang, C., Wu, S., Zheng, H., Liu, Y., Zhu, L., Wang, H., Ding, M., Wan, Z., Liu, H., Wang, S., Zhu, H., Zhang, X., Chai, N., Liu, Y., Lai, P., Yuan, S., Su, Z., Zhang, G., Zhou, W., Du, Y., Huang, W., and Shi, G.
\newblock {EdgeBench}: Unveiling scaling laws of learning from real-world environments.
\newblock \emph{arXiv preprint arXiv:2607.05155}, 2026.

\end{thebibliography}
\bibliographystyle{preprint}

\clearpage
\appendix
\renewcommand{\topfraction}{0.9}
\renewcommand{\bottomfraction}{0.8}
\renewcommand{\textfraction}{0.07}
\setcounter{totalnumber}{4}
\makeatletter
\setlength{\@fptop}{0pt}
\makeatother

\twocolumn[{%
\centering
\includegraphics[width=\textwidth]{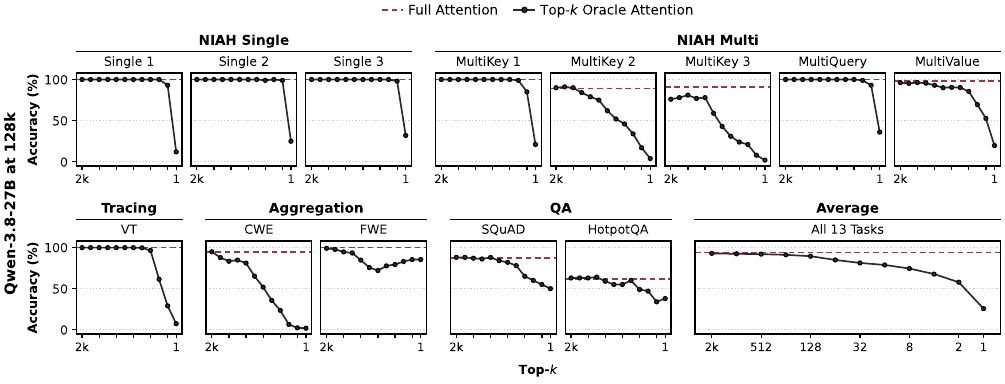}
\captionof{figure}{\textbf{Top-$k$ Oracle Attention on RULER.} \qwen{} at 128k tokens. Each query attends to a 4-token sink, a 512-token window and the top-$k$ long-range keys, one list shared by all heads. Process: sum over all heads, pick top-$k$, renormalise per head over the shared selection.}
\label{fig:topk-oracle}
\vskip 0.25in
}]

\section{Attention Sparsity}
\label{app:sparsity}

To test how many keys a query needs to keep accuracy, we let each query of \qwen{} attend only to the sink, the window and its true top-$k$ keys (Figure~\ref{fig:topk-oracle}).
At 128k tokens, Top-$k$ Oracle Attention keeps 98\% of the RULER accuracy of Full Attention down to $k = 512$, while each query attends to 0.8\% of the keys.
Single-needle retrieval holds down to $k = 16$, while MultiKey 2, MultiKey 3 and CWE fall first in accuracy.
Therefore, sparse attention loses little if it finds the most relevant keys, which is what \name{} is designed to do without scoring every key.

\section{Method Details}
\label{app:method}

Table~\ref{tab:notation} lists the symbols of \name{}.
Algorithm~\ref{alg:ward} clusters one cluster block into buckets, and Algorithm~\ref{alg:beam} is the beam search of \dname{} (Tree).

\begin{table}[t]
\centering
\caption{\textbf{Notation.} \dname{} symbols and their description.}
\label{tab:notation}
\vskip 0.15in
\ssmall
\setlength{\tabcolsep}{6pt}
\begin{tabularx}{\columnwidth}{@{}>{\rule[-1.7ex]{0pt}{4.2ex}}lX@{}}
\toprule
\multicolumn{1}{@{}l}{\textbf{Symbol}} & \textbf{Description} \\
\midrule
$T$ & context size \\
$i$,\enspace $j$,\enspace $h$ & query position, key position, head \\
\midrule
$W_{Q,K,V}^{(\mathrm{attn})}$ & attention projections \\
$q_{i,h}^{(\mathrm{attn})}$,\enspace $k_j^{(\mathrm{attn})}$ & attention query (head $h$) and key \\
$n_h$,\enspace $d_h$ & query heads, head dimension \\
\midrule
$W_{Q,K}^{(\mathrm{idx})}$,\enspace $d^{(\mathrm{idx})}$ & index projections, index dimension \\
$q_i^{(\mathrm{idx})}$,\enspace $k_j^{(\mathrm{idx})}$ & index query and key, unit norm \\
$\mathcal{R}_i$ & long-range keys of query $i$ (without sink and window) \\
$A_{i,j}$,\enspace $\hat{A}_{i,j}$ & attention weight of the model / indexer for query $i$ and key $j$ \\
$\tau$ & temperature of the indexer softmax \\
\midrule
$n_{\mathrm{pool}}$,\enspace $k_{\mathrm{pool}}^{(\mathrm{idx})}$ & keys summed per pool key, pool key \\
$n_{\mathrm{cluster}}$,\enspace $m$ & keys per cluster block, buckets per block \\
$\bar{n}_b$,\enspace $n_b^{\mathrm{max}}$ & average and max number of keys per bucket \\
$k_b^{\mathrm{sum}}$,\enspace $\lvert b \rvert$ & bucket key sum, keys in bucket $b$ \\
$a_b$,\enspace $\Delta(b, b')$ & alignment, merge cost \\
$c_b$,\enspace $B_{\mathrm{cache}}^{(\mathrm{idx})}$ & bucket centroid (unit norm), bucket cache \\
$\ell$,\enspace $n_{b,\ell}$,\enspace $n_{\mathrm{cluster},\ell}$ & level, buckets per node, buckets per node block \\
$p_\ell$,\enspace $n_{\mathrm{start}}$ & node buckets kept per level, at most at the start level \\
\midrule
$s_{i,b}$,\enspace $p$ & bucket score, buckets selected per query \\
$k$,\enspace $n_{\mathrm{sink}}$,\enspace $n_w$ & long-range, sink and window keys \\
$J_i$ & key positions query $i$ attends to \\
\bottomrule
\end{tabularx}
\end{table}

\begin{algorithm}[t]
\caption{Ward clustering of one cluster block}
\label{alg:ward}
\small
\begin{algorithmic}[1]
\REQUIRE \begin{tabular}[t]{@{}l@{}}index keys $k_j^{(\mathrm{idx})}$ of a complete cluster block \\ sizes $n_{\mathrm{cluster}}$, $n_{\mathrm{pool}}$, $m$, $n_b^{\mathrm{max}}$\end{tabular}
\ENSURE $B_{\mathrm{cache}}^{(\mathrm{idx})}$ with $m$ centroids $c_b$ and their key positions
\algophase{Pooling}
\FORALL{pool keys $k_{\mathrm{pool}}^{(\mathrm{idx})}$, in parallel}
\STATE $k_{\mathrm{pool}}^{(\mathrm{idx})} \gets$ the sum of its $n_{\mathrm{pool}}$ consecutive $k_j^{(\mathrm{idx})}$
\ENDFOR
\STATE start one bucket $b$ per pool key, $k_b^{\mathrm{sum}} \gets k_{\mathrm{pool}}^{(\mathrm{idx})}$, $\lvert b \rvert \gets n_{\mathrm{pool}}$
\algophase{Merging}
\WHILE{\algolinelabel{line:ward-while}more than $m$ buckets remain}
\begin{algobox}{algogreen}{every pair of buckets, in parallel}
\STATE\algolinelabel{line:ward-gram} $G \gets [\,k_b^{\mathrm{sum}} \cdot k_{b'}^{\mathrm{sum}}\,]$ \COMMENT{Gram matrix}
\STATE $a_b \gets \sqrt{G_{bb}}$, \ $a_{b \cup b'} \gets \sqrt{G_{bb} + G_{b'b'} + 2 G_{bb'}}$
\STATE $\Delta(b, b') \gets a_b + a_{b'} - a_{b \cup b'}$ \COMMENT{merge cost}
\STATE $\pi(b) \gets \arg\min_{b'} \Delta(b, b')$, \ $\lvert b \rvert + \lvert b' \rvert \le n_b^{\mathrm{max}}$
\end{algobox}
\begin{algobox}{algogreen}{the cheapest mutual pairs $\pi(b) = b'$, $\pi(b') = b$\newline at most as many as buckets above $m$, in parallel}
\STATE\algolinelabel{line:ward-merge} $k_b^{\mathrm{sum}} \gets k_b^{\mathrm{sum}} + k_{b'}^{\mathrm{sum}}$, \ $\lvert b \rvert \gets \lvert b \rvert + \lvert b' \rvert$, \ remove $b'$
\end{algobox}
\ENDWHILE\algolinelabel{line:ward-endwhile}
\algophase{Bucket cache}
\FORALL{buckets $b$, in parallel}
\STATE $c_b \gets k_b^{\mathrm{sum}} / \lVert k_b^{\mathrm{sum}} \rVert$
\STATE store $c_b$ and the positions of the keys of $b$ in $B_{\mathrm{cache}}^{(\mathrm{idx})}$
\ENDFOR
\begin{algoboxwide}{algogrey}{\dname{} (Tree) only}
\STATE\algolinelabel{line:ward-node} merge blocks of $n_{\mathrm{cluster},\ell}$ buckets into node buckets
\end{algoboxwide}
\end{algorithmic}
\end{algorithm}

\begin{algorithm}[t]
\caption{Beam search of one query in \dname{} (Tree)}
\label{alg:beam}
\small
\begin{algorithmic}[1]
\REQUIRE \begin{tabular}[t]{@{}l@{}}index query $q_i^{(\mathrm{idx})}$, levels $\ell = 0, \dots, \ell^{\mathrm{max}}$ of $B_{\mathrm{cache}}^{(\mathrm{idx})}$ \\ sizes $p$, $p_\ell$, $n_{\mathrm{start}}$\end{tabular}
\ENSURE the top-$p$ buckets for query $i$
\STATE\algolinelabel{line:beam-start} $C \gets$ all node buckets of the lowest level $\ell$ with $\le n_{\mathrm{start}}$
\algophase{Descent}
\WHILE{\algolinelabel{line:beam-while}$\ell > 0$}
\begin{algobox}{algogreen}{every candidate in $C$, in parallel}
\STATE\algolinelabel{line:beam-score} $s_{i,b} \gets q_i^{(\mathrm{idx})} \cdot c_b$ \COMMENT{node score}
\end{algobox}
\STATE\algolinelabel{line:beam-keep} $P \gets$ the best $p_\ell$ candidates of $C$
\STATE\algolinelabel{line:beam-children} $C \gets$ the children of $P$ $\cup$ the open nodes of level $\ell - 1$
\STATE $\ell \gets \ell - 1$
\ENDWHILE\algolinelabel{line:beam-endwhile}
\algophase{Selection}
\begin{algoboxwide}{algogreen}{every bucket in $C$, in parallel}
\STATE\algolinelabel{line:beam-level0} $s_{i,b} \gets q_i^{(\mathrm{idx})} \cdot c_b$ \COMMENT{bucket score}
\end{algoboxwide}
\STATE\algolinelabel{line:beam-topp} \textbf{return} the top-$p$ buckets of $C$
\end{algorithmic}
\end{algorithm}

\section{Training Data}
\label{sec:training-data}

In this section, we report additional details on the training dataset we use to train \dname{} under indexer distillation, and also train \nname{} in the decay phase: \emph{LongDocs}.

We use the following terminology:
\begin{itemize}[leftmargin=*,nosep]
\item \textbf{Source}: a collection of documents, such as Project Gutenberg.
\item \textbf{Domain}: a group of sources: books, academic, code, agentic, government, parliament and discussion.
\item \textbf{Document}: one item of a source, such as a book, a paper, a code repository or an agent trace.
\item \textbf{Sequence}: a contiguous part of one document, at a particular sequence length from 16k to 1M tokens.
\end{itemize}

\begin{figure}[t]
\centering
\includegraphics[width=\columnwidth]{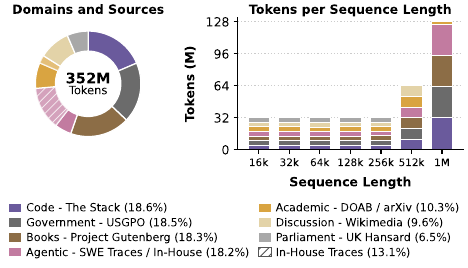}
\caption{\textbf{LongDocs.}
Tokens of the distillation mix by domain and source (left) and by sequence length (right), with the in-house agent traces hatched.}
\label{fig:longdocs}
\end{figure}

\textbf{Sources.}
LongDocs is a collection of long documents from publicly available sources and in-house agent traces.
Eight sources are publicly available: Project Gutenberg, DOAB, arXiv, USGPO, UK Hansard and Wikimedia from the filtered releases of the Common Pile~\citep{kandpal2025common}, The Stack~\citep{kocetkov2023stack}, of which we take the Python and Java repositories, and the software engineering traces of Open-SWE-Traces~\citep{nvidia2026openswe}.
The in-house agent traces are long sessions of agents that play games or solve long-context tasks, and they make up 13.1\% of the training tokens.
We use every public source as released.

\textbf{Sequences.}
Each sequence comes from one document and never joins two documents.
A code document is one repository, its files joined in the order of their paths, and an agent trace is one session rendered as text.
We hold out 2\% of the documents for validation.

\textbf{Distillation training.}
We distil every indexer on a mix of 352M tokens drawn from LongDocs (Figure~\ref{fig:longdocs}). 
The mix is uniformly sampled across domain and sequence length, with a bias towards longer context windows at 512k and 1M. In particular, we give each sequence length 32M tokens, and twice and four times as many to 512k and 1M. Sequences of 1M tokens thus hold about a third of the training tokens. We experimentally determine this choice to be beneficial, with +0.036 accuracy on RULER at 1M (averaged over 3 seeds, notable lifts in \textsc{multikey\_2}, \textsc{multikey\_3}, +0.35 and +0.15 respectively), while accuracy from 4k to 512k stays roughly identical.
Within each length, every domain receives an equal share of the tokens.
A domain with too few documents of that length gives its share to the others (parliament stops at 256k tokens and discussion at 512k).

\textbf{Native training.}
The decay training stage of \nname{} reads the whole collection rather than the mix, without the long-context agent sessions.
The data is tokenised (18.9B tokens over 134,740 documents), and cut into sequence lengths the training requires.

\section{PaperTrail}
\label{sec:papertrail}

\begin{figure}[t]
\centering
\includegraphics[width=\columnwidth]{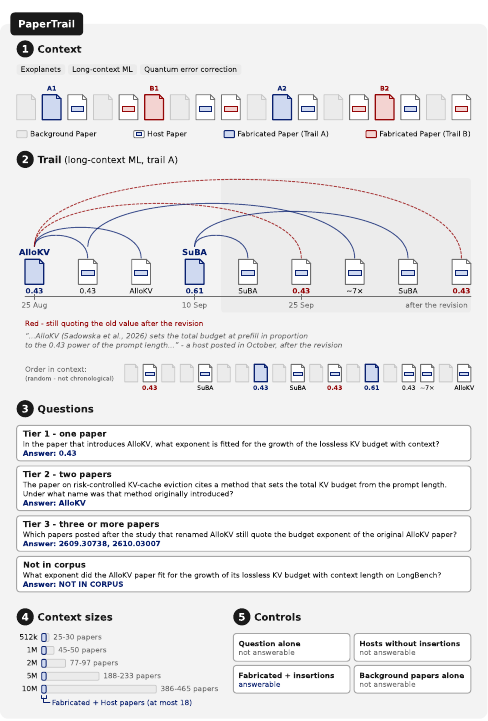}
\caption{\textbf{PaperTrail.}
\keystep{1} A sample, its fabricated, host and background papers.
\keystep{2} One trail, introduced, revised and renamed, and still quoted in its old form after the revision.
\keystep{3} Questions by tier, and one whose answer is not in the papers.
\keystep{4} Context sizes, each holding every paper of the size below.
\keystep{5} Controls.}
\label{fig:papertrail-design}
\end{figure}

As part of this work, we introduce PaperTrail, an extension of the RULER~\citep{hsieh2024ruler} benchmark that tests models at ultra-long context in a more realistic setting, inspired by real-life scientific literature review (Figure~\ref{fig:papertrail-design}).
Instead of finding a needle in synthetic text, a model finds facts spread across real scientific papers and connects them, as in a literature search.
We use the following terminology:
\begin{itemize}[leftmargin=*,nosep]
\item \textbf{Sample}: a collection of real arXiv preprints of 2026 from one domain.
\item \textbf{Trail}: a chain of papers that introduces a result, cites it, and later revises and renames it.
\item \textbf{Fabricated paper}: a paper we write for a trail.
\item \textbf{Host paper}: a real paper into which we insert passages that cite or use facts from a fabricated paper.
\item \textbf{Insertion}: a passage we add to a host paper, which only adds sentences and never changes the text of the authors.
\item \textbf{Background paper}: any other paper of a sample, real and unchanged.
\end{itemize}

\begin{figure*}[t]
\centering
\includegraphics[width=\textwidth]{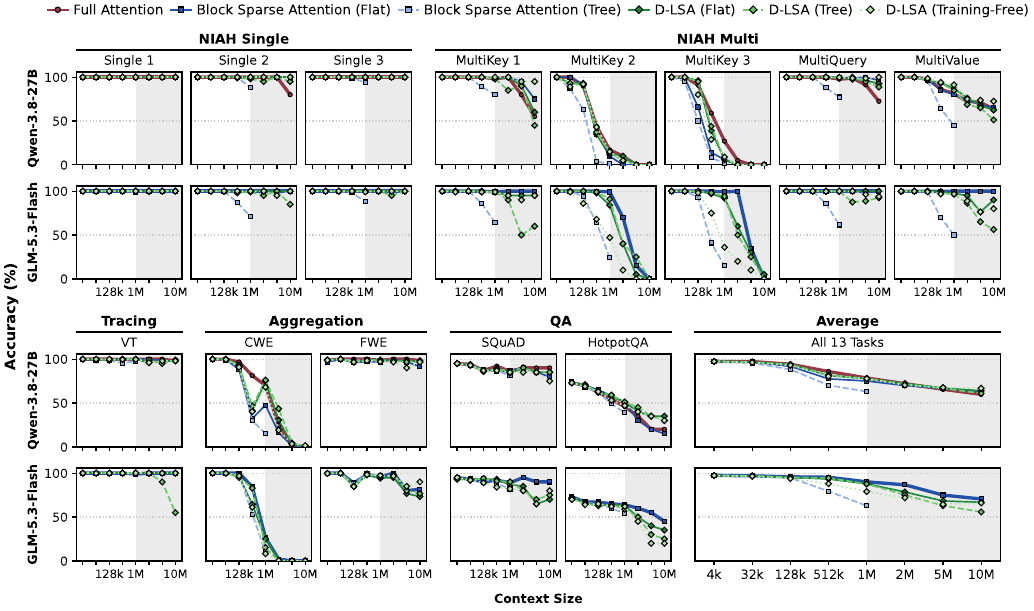}
\caption{\textbf{\dname{} accuracy on RULER individual tasks.} The 13 tasks of RULER from 4k to 10M tokens on \qwen{} and \glm{}, by task group, with their average.}
\label{fig:ruler}
\end{figure*}

\keystep{1} \textbf{Context.}
PaperTrail has three samples, on \emph{exoplanets}, \emph{long-context machine learning} and \emph{quantum error correction}, with 1,246 papers in total.
Each sample hides two trails (Trail A and Trail B).
A trail has two fabricated papers and 8 to 9 insertions in 6 to 7 host papers, and background papers fill the rest of the context.
An insertion places its references where an author would place them, such that a host paper reads naturally as before.
Papers in the context are not presented in chronological order, and a host often comes before the paper it cites.

\keystep{2} \textbf{Trails.}
The first fabricated paper of a trail introduces a result, for example AlloKV, a KV cache eviction method whose budget grows as the 0.43 power of the context length.
Host papers cite and use it.
Two to three weeks later, the second fabricated paper revises the value and renames the method, here to SuBA with an exponent of 0.61.
Host papers posted after the revision either cite the new name or still quote the old value, as real literature often does.

\keystep{3} \textbf{Questions.}
Each sample asks 30 scored questions, 10 per tier.
A tier 1 question recalls a fact from one paper, a tier 2 question links two papers, and a tier 3 question links three or more papers, often the whole trail.
Each sample also asks 6 questions whose answer is not in the papers, about an absent entity, a false premise or a near miss, which we report apart from the score (Table~\ref{tab:papertrail}).
The model ends its reply with the papers it used and its answer.
A number counts within a relative tolerance of 2\% by default, and a set of papers is scored by its F1 against the reference set.

\keystep{4} \textbf{Context sizes.}
We build each sample at 512k, 1M, 2M, 5M and 10M tokens, and every size holds every paper of the size below.
The fabricated and host papers, at most 18 per sample, are the same at every size.
The background papers grow the context from 25 to 30 papers at 512k to 386 to 465 papers at 10M, and spread the trails over the whole context.
The trails are fabricated, and every paper up to 5M was posted after the release of both \qwen{} and \glm{}, guaranteeing that no model has seen these papers in training.
The 10M contexts hold papers posted at most six weeks before its release.

\keystep{5} \textbf{Controls.}
Four controls ask the same questions over parts of a sample: the question alone, the host papers without their insertions, the fabricated papers with the insertions alone, and the background papers alone.
By construction, only the fabricated papers and the insertions hold the answers.
We validate the controls with \glm{} and \dname{} (Training-Free) in Table~\ref{tab:papertrail}.

\section{Accuracy}
\label{app:accuracy}

In this section, we show details of the evaluation setup, compare every method with its baseline on knowledge and reasoning, examine whether \dname{} changes the length of the reasoning, and break down the long-context results, RULER by task and PaperTrail by tier.

\subsection{Evaluation Setup}
\label{app:evaluation-setup}

Table~\ref{tab:evaluation} lists the setup of every benchmark.
We sample \qwen{} at temperature 1.0 with top-p 0.95, top-k 20, min-p 0 and no presence penalty, and \glm{} at temperature 1.0 with top-p 0.95, as their model cards recommend.
On RULER, both models decode greedily with reasoning disabled.
On every benchmark with reasoning, at most 4.6\% of the replies reach the output cap of 163,840 tokens for any model and method.

\begin{table}[t]
\centering
\caption{\textbf{Evaluation setup.}
Questions, repeats, generation and metric of every benchmark, with RULER questions per task.}
\label{tab:evaluation}
\vskip 0.15in
\setlength{\tabcolsep}{2.8pt}
\ssmall
\begin{tabular}{@{}l!{\separator}cccc!{\separator}l@{}}
\toprule
 & Questions & Repeats & Reasoning & Max output & Metric \\
\midrule
MMLU-Pro & 2,000 & 1 & on & 163,840 & Exact match \\
GPQA-Diamond & 198 & 1 & on & 163,840 & Exact match \\
AIME25 & 30 & 4 & on & 163,840 & Math-Verify \\
MATH500 & 100 & 1 & on & 163,840 & Math-Verify \\
LongBench v2 & 60 & 1 & on & 163,840 & Accuracy \\
RULER & 100 ($\leq$1M) / 20 ($>$1M) & 1 & off & 4,096 & String match \\
PaperTrail & 90 & 3 & on & 163,840 & F1 \\
\bottomrule
\end{tabular}
\end{table}

\subsection{Knowledge and Reasoning}
\label{app:knowledge-reasoning}

Table~\ref{tab:accuracy-gaps} compares every method with the baselines on knowledge, reasoning and LongBench v2.
Every score lies within the 95\% confidence interval of the respective baselines.

\begin{table}[t]
\centering
\caption{\textbf{Accuracy compared to the baseline.}
Baseline scores are in bold.
Other rows show the difference to the baseline in points.
Differences smaller than the 95\% confidence interval (CI) of the baseline are not significant.}
\label{tab:accuracy-gaps}
\vskip 0.15in
\setlength{\tabcolsep}{2.6pt}
\ssmall
\begin{tabular}{@{}l!{\separator}ccccc@{}}
\toprule
 & MMLU-Pro & GPQA-D & AIME25 & MATH500 & LongBench v2 \\
\midrule
\multicolumn{6}{@{}l}{\textbf{\qwen{}}} \\
\addlinespace[1.5pt]
\boxfull{} Full Attention & \textbf{83.2} & \textbf{88.4} & \textbf{98.3} & \textbf{98.0} & \textbf{70.0} \\
\boxblockflat{} Block Sparse (Flat) & +0.3 & $-$2.0 & +0.8 & +1.0 & 0.0 \\
\boxblocktree{} Block Sparse (Tree) & +0.2 & $-$2.0 & $-$0.8 & +2.0 & $-$1.7 \\
\boxlsaflat{} \dname{} (Flat) & $-$0.3 & $-$0.5 & $-$0.8 & +1.0 & $-$5.0 \\
\boxlsatree{} \dname{} (Tree) & +0.3 & $-$1.5 & +1.7 & +2.0 & $-$8.3 \\
\boxlsafree{} \dname{} (Training-Free) & $-$0.6 & $-$3.0 & $-$1.7 & +1.0 & $-$3.3 \\
\textit{95\% CI} & \textit{$\pm$1.6} & \textit{$\pm$4.5} & \textit{$\pm$2.3} & \textit{$\pm$2.7} & \textit{$\pm$11.6} \\
\midrule
\multicolumn{6}{@{}l}{\textbf{\glm{}}} \\
\addlinespace[1.5pt]
\boxblockflat{} Block Sparse (Flat) & \textbf{85.4} & \textbf{85.9} & \textbf{97.5} & \textbf{98.0} & \textbf{70.0} \\
\boxblocktree{} Block Sparse (Tree) & +0.2 & +2.0 & $-$0.8 & 0.0 & $-$1.7 \\
\boxlsaflat{} \dname{} (Flat) & +0.3 & +1.0 & $-$1.7 & 0.0 & $-$1.7 \\
\boxlsatree{} \dname{} (Tree) & +0.3 & +0.5 & +1.7 & +1.0 & 0.0 \\
\boxlsafree{} \dname{} (Training-Free) & +0.4 & +2.0 & $-$2.5 & +1.0 & +1.7 \\
\textit{95\% CI} & \textit{$\pm$1.5} & \textit{$\pm$4.9} & \textit{$\pm$2.8} & \textit{$\pm$2.7} & \textit{$\pm$11.6} \\
\bottomrule
\end{tabular}
\end{table}

\subsection{Reasoning Length}
\label{app:reasoning-length}

\begin{figure}[t]
\centering
\includegraphics[width=\columnwidth]{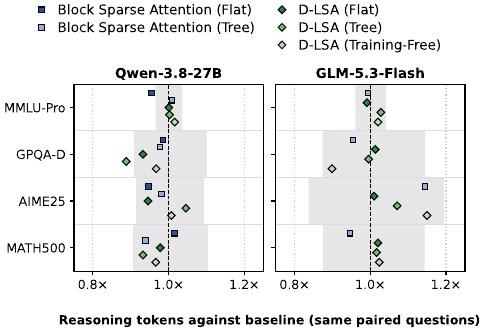}
\caption{\textbf{Reasoning tokens against baselines.}
Reasoning tokens of each method relative to the baseline of its model, on the same questions.
Grey bands mark the expected variation, and markers inside them show no measurable change in reasoning length.}
\label{fig:reasoning-length}
\end{figure}

Figure~\ref{fig:reasoning-length} compares the reasoning length of each method with the baseline of its model, question by question.
For each question, we average the reasoning tokens over its repeats and divide them by those of the baseline on the same question, and each marker is the geometric mean of these ratios over the benchmark.
The grey band of a benchmark is the median width of the 95\% bootstrap intervals of its methods, and a marker inside it shows no measurable change.
Every \dname{} design stays inside its band, except \dname{} (Tree) on GPQA-Diamond with \qwen{}, which reasons 11\% shorter.

\subsection{RULER}
\label{app:ruler}

Figure~\ref{fig:ruler} shows the 13 tasks of RULER from 4k to 10M tokens, grouped into retrieval, tracing, aggregation and question answering.
Single-needle retrieval stays at 95 or above for every method up to 5M tokens, except Block Sparse Attention (Tree), which falls to 88 and 71 at 1M tokens.
MultiKey 2, MultiKey 3 and CWE fall to at most 5 at 10M tokens for every method, including the baselines, and bound the RULER average of every method.
On \qwen{}, \dname{} (Flat) exceeds baseline Full Attention at 10M tokens mostly on MultiQuery and HotpotQA, at 92 and 35 against 72 and 20.
On \glm{}, \dname{} (Tree) falls to 60, 56 and 55 on MultiKey 1, MultiValue and VT at 10M tokens, against 100 for baseline sparse attention, and these three tasks account for most of its gap to \dname{} (Flat).

\subsection{PaperTrail}
\label{app:papertrail}

\begin{figure}[t]
\centering
\includegraphics[width=\columnwidth]{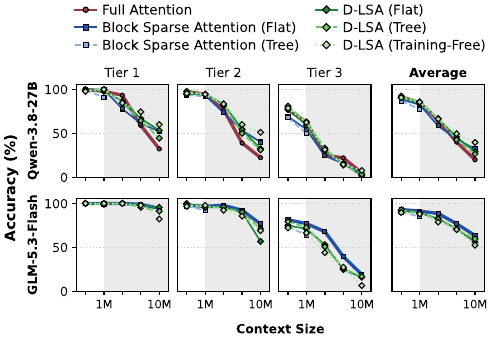}
\caption{\textbf{\dname{} accuracy on PaperTrail per tier.}
Tier 1, tier 2 and tier 3 questions, each averaged over the three domains and three repeats.}
\label{fig:papertrail}
\end{figure}

Figure~\ref{fig:papertrail} shows the three tiers of PaperTrail and their average from 512k to 10M tokens.
The harder the tier, the earlier it falls.
Tier 3 questions fall to at most 19.0 at 10M tokens for every method on both models, while tier 1 questions stay above 90 on \glm{}.
On \qwen{}, \dname{} exceeds baseline Full Attention from 5M tokens on tier 1 and tier 2 questions, and \dname{} (Flat) reaches 53.3 and 32.2 against 32.2 and 22.2 at 10M tokens.
On \glm{}, \dname{} trails baseline sparse attention mostly on tier 3 questions from 2M tokens, and \dname{} (Flat) also on tier 2 questions at 10M tokens, at 56.7 against 76.7.

Table~\ref{tab:papertrail} further reports the questions whose answer is not in the papers, at 512k tokens.
Every \dname{} design recognises these questions within 3.7 points of the baseline of its model, at 96.3 to 100.0 against 100.0 for \qwen{} and 96.3 for \glm{}.

\begin{table}[t]
\centering
\caption{\textbf{PaperTrail not-in-corpus questions and controls.}
Score (\%) on the questions whose answer is not in the papers at 512k tokens, for \boxfull~Full Attention, \boxblockflat~Block Sparse Attention (Flat) and \boxblocktree~(Tree), and \boxlsaflat~\dname{} (Flat), \boxlsatree~(Tree) and \boxlsafree~(Training-Free).
The controls ask the scored questions over parts of a sample, with \glm{} and \dname{} (Training-Free).}
\label{tab:papertrail}
\vskip 0.15in
\setlength{\tabcolsep}{6.6pt}
\ssmall
\begin{tabular}{@{}l!{\separator}ccc!{\separator}c@{}}
\toprule
 & Exoplanets & Long-context ML & QEC & Average \\
\midrule
\multicolumn{5}{@{}l}{\textbf{Not in corpus, \qwen{}}} \\
\addlinespace[1.5pt]
\boxfull{} Full Attention & 100.0 & 100.0 & 100.0 & 100.0 \\
\boxblockflat{} Block Sparse (Flat) & 100.0 & 100.0 & 83.3 & 94.4 \\
\boxblocktree{} Block Sparse (Tree) & 100.0 & 100.0 & 88.9 & 96.3 \\
\boxlsaflat{} \dname{} (Flat) & 100.0 & 100.0 & 88.9 & 96.3 \\
\boxlsatree{} \dname{} (Tree) & 100.0 & 94.4 & 94.4 & 96.3 \\
\boxlsafree{} \dname{} (Training-Free) & 100.0 & 94.4 & 100.0 & 98.1 \\
\addlinespace[3pt]
\multicolumn{5}{@{}l}{\textbf{Not in corpus, \glm{}}} \\
\addlinespace[1.5pt]
\boxblockflat{} Block Sparse (Flat) & 100.0 & 88.9 & 100.0 & 96.3 \\
\boxblocktree{} Block Sparse (Tree) & 100.0 & 88.9 & 94.4 & 94.4 \\
\boxlsaflat{} \dname{} (Flat) & 100.0 & 100.0 & 100.0 & 100.0 \\
\boxlsatree{} \dname{} (Tree) & 100.0 & 88.9 & 100.0 & 96.3 \\
\boxlsafree{} \dname{} (Training-Free) & 100.0 & 94.4 & 100.0 & 98.1 \\
\midrule
\multicolumn{5}{@{}l}{\textbf{Controls, \glm{}} \boxlsafree} \\
\addlinespace[1.5pt]
Question alone & 0.0 & 0.0 & 0.0 & 0.0 \\
Hosts without insertions & 0.0 & 0.0 & 0.2 & 0.1 \\
Fabricated + insertions & 100.0 & 100.0 & 100.0 & 100.0 \\
Background alone & 0.0 & 0.0 & 0.0 & 0.0 \\
\bottomrule
\end{tabular}
\end{table}

\subsection{Block Size}
\label{app:block-size}

\begin{figure}[t]
\centering
\includegraphics[width=\columnwidth]{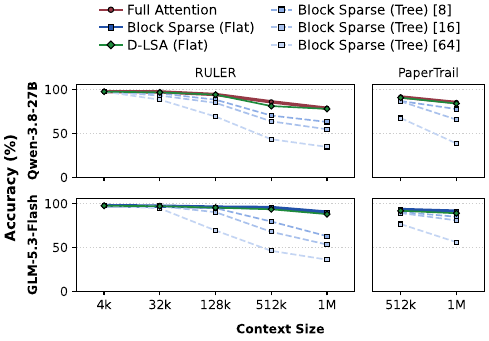}
\caption{\textbf{Block size of Block Sparse Attention (Tree).}
RULER and PaperTrail up to 1M tokens with blocks of 8, 16 and 64 keys ($n_{\mathrm{pool}}$), each selecting the same 4,608 keys, against the baseline of each model and \dname{} (Flat).}
\label{fig:block-size}
\end{figure}

Figure~\ref{fig:block-size} compares Block Sparse Attention (Tree) with blocks of 8, 16 and 64 keys, all selecting the same number of keys.
At short context, every block size matches the baseline.
As the context grows, accuracy falls earlier and further the larger the block, on both benchmarks and both models, as a larger block summarises more keys in one score and can hide a single relevant key.
Even with blocks of eight keys, the size we use in the main results, the tree baseline trails \dname{} (Flat) at long context.


\section{Performance}
\label{app:performance}

We first introduce the hardware, describe the kernels of the indexer (Appendix~\ref{app:kernels}), and how \dname{} integrates with different parallelism strategies (Appendix~\ref{app:parallelism}): Tensor Parallelism (TP) and Context Parallelism (CP).

\textbf{Hardware.}
We run on NVIDIA H200 (Hopper) and B300 (Blackwell Ultra) GPUs, eight per node (NVLink domain), with the details presented in Table~\ref{tab:hardware}.

\begin{table}[t]
\centering
\caption{\textbf{Hardware specs.}
Specifications of the GPUs we use, as given by NVIDIA~\citep{nvidia2026h200, nvidia2026hgx, nvidia2026blackwellultra, nvidia2026computecapabilities}.}
\label{tab:hardware}
\vskip 0.15in
\setlength{\tabcolsep}{8pt}
\ssmall
\begin{tabular}{@{}l!{\separator}rr@{}}
\toprule
 & H200 & B300 \\
\midrule
Architecture & Hopper (sm\_90) & Blackwell Ultra (sm\_103) \\
SMs & 132 & 148 \\
Shared memory per SM & 228\,KiB & 228\,KiB \\
Matrix instructions & WGMMA & tcgen05 \\
\midrule
Memory (HBM3e) & 141\,GB & 288\,GB \\
HBM bandwidth & 4.8\,TB/s & 8\,TB/s \\
NVLink per GPU & 900\,GB/s & 1.8\,TB/s \\
\midrule
bf16 tensor & 989\,Tflop/s & 2,250\,Tflop/s \\
fp32 & 67\,Tflop/s & 75\,Tflop/s \\
Ridge point (bf16) & 206\,flop/B & 281\,flop/B \\
\bottomrule
\end{tabular}
\end{table}

\textbf{Precision.}
The KV cache, the index keys and the bucket centroids are in bf16.
For the bucket centroids, which every decode step of the flat search reads in full, bf16 halves the bytes of fp32 at little cost to the selection.
In an offline study on \glm{},
the bf16 indexer selects roughly 20 of the 4,096 long-range keys of the flat search differently from an fp32 reference, and changes the attention mass that the selected keys cover by less than 0.0001.
Bucket centroids in fp8 would halve their bytes again, but would select about six times as many keys differently, so we keep them in bf16. We do not observe similar sensitivity for the node centroids, which we set to fp8 in \dname{} (Tree).

\textbf{Conventions.}
We run prefill over \emph{chunks} $\mathcal{C}$ of $\lvert \mathcal{C} \rvert = 32768$ consecutive queries, and decode over steps of one new query per request.
All sizes are per layer, and per chunk in prefill or per step in decode.
We run \dname{} with index keys and queries of $d^{(\mathrm{idx})} = 64$ dimensions, cluster blocks of $n_{\mathrm{cluster}} = 512$ keys, buckets of $\bar{n}_b = 8$ keys on average and at least $n_{\mathrm{pool}} = 4$.
Each query selects $p = 1024$ buckets, so that even buckets of $n_{\mathrm{pool}}$ keys hold its $k = 4096$ long-range keys, and the expansion fills $k$ from about the best half of them on average.
It also attends to $n_{\mathrm{sink}} = 4$ sink keys and a window of $n_w = n_{\mathrm{cluster}}$ keys.
On \glm{}, \dname{} (Flat) uses $k = 2048$ and $p = 512$ for the queries in the first 1M positions of the context (Table~\ref{tab:methods}), but for queries of a chunk at 10M tokens, the sizes above apply.
Therefore, at a context of $T = 10 \times 2^{20} \approx 10$M tokens, a layer holds $T / \bar{n}_b = 1310720$ buckets, whose centroids take $2 d^{(\mathrm{idx})} T / \bar{n}_b = 160$\,MiB.

\subsection{Kernels}
\label{app:kernels}

\begin{figure}[t]
\centering
\includegraphics[width=\columnwidth]{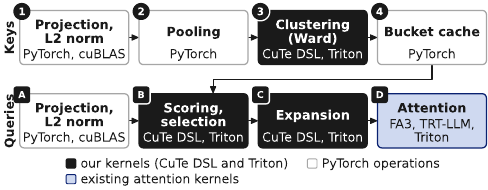}
\caption{\textbf{Implementation of the steps of \dname{}.} Badges as in Figure~\ref{fig:design}.}
\label{fig:kernels-map}
\end{figure}

We refer to the steps of \dname{} by their badges in Figure~\ref{fig:design}, and Figure~\ref{fig:kernels-map} shows how we implement each of them.
We write kernels for three steps of the indexer: \keystep{3}~clustering, \querystep{B}~scoring and selection of the buckets, and \querystep{C}~expansion of the selected buckets into the index lists $J_i$.
These kernels are written in CuTe DSL, the Python-based domain-specific language of CUTLASS~\citep{thakkar2023cutlass}, to target H200 and B300 specifically.
We also keep a Triton~\citep{tillet2019triton} version of each kernel, which is not tied to these two GPUs.
We give the shapes and times of the kernels on H200, and note where B300 differs.
For \querystep{D}~attention, we rely on pre-existing attention kernels, that attend over the index lists $J_i$ unchanged.
The only exception is CP (Appendix~\ref{app:parallelism}), for which we do add an attention kernel.

\keystep{3} \textbf{Clustering.}
The clustering does relatively little work: one Gram matrix per cluster block, over its pooled keys, takes about 0.13\,Gflop for a whole chunk, against 5.5\,Tflop for the bucket scan at 10M tokens.
Ward's method, however, merges buckets in rounds that run one after the other.
Each of these rounds scores every pair of buckets and merges the cheapest mutual pairs (Algorithm~\ref{alg:ward}, lines~\ref{line:ward-while}-\ref{line:ward-endwhile}), until $m$ buckets remain.
The overall runtime of the clustering step is therefore set by the latency of these rounds.
Our kernel cuts the time of each round by keeping its data on chip and avoiding round trips to the host between rounds.
More precisely, the kernel keeps the state of a cluster block in shared memory for all its rounds, and spreads its search for partners over a thread-block cluster of up to eight streaming multiprocessors (SMs), which exchange their results through distributed shared memory.
In \dname{} (Tree), each pairing that forms the node buckets (line~\ref{line:ward-node}) runs all its rounds in a single launch, whose thread blocks synchronise through counters in high-bandwidth memory (HBM), so the host neither launches nor reads anything between rounds.

\begin{figure}[t]
\centering
\IfFileExists{plots/kernels_threshold.pdf}{\includegraphics[width=\columnwidth]{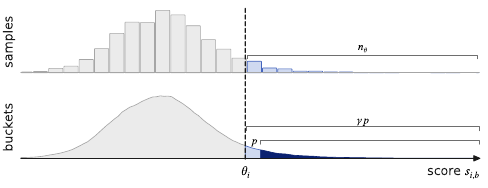}}{\fbox{\parbox{0.9\columnwidth}{\centering threshold figure (in progress)}}}
\caption{\textbf{Threshold estimate.} For one query $i$. Top: the sampled scores, with $\theta_i$ at or just below the $n_\theta$-th best. Bottom: the row, of which about $\gamma p$ buckets score at or above $\theta_i$ and are listed, the top-$p$ among them.}
\label{fig:kernels-threshold}
\end{figure}

\begin{figure*}[t]
\centering
\includegraphics[width=\textwidth]{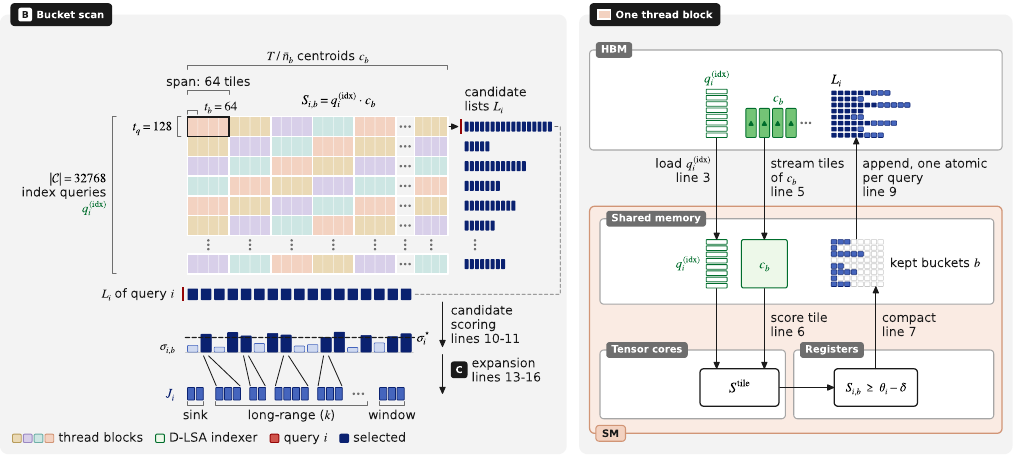}
\caption{\querystep{B} \textbf{Bucket scan on H200.} Left: the thread blocks (colours) over tiles of $t_q$ index queries and spans of centroid tiles, and below, candidate scoring and \querystep{C}~expansion for one query. Right: one thread block (Algorithm~\ref{alg:prefill-select}, lines~\ref{line:prefill-load}-\ref{line:prefill-append}).}
\label{fig:kernels-scan}
\end{figure*}

\querystep{B} \querystep{C} \textbf{Selection by threshold.}
\dname{} scores an index query $q_i^{(\mathrm{idx})}$ against the centroids $c_b$ of the buckets $\mathcal{B}_i$ of the cluster blocks before its own, and keeps its top-$p$ buckets.
Over the index queries of a chunk, the scores $s_{i,b} = q_i^{(\mathrm{idx})} \cdot c_b$ form a matrix with one row per query, so we call the scores of query $i$ its \emph{row}, in decode as well.
The row holds $N_i = \lvert \mathcal{B}_i \rvert$ scores, which grows with the position of the query, to about $T / \bar{n}_b$ at the end of a context of $T$ tokens.
In prefill and in decode alike, many thread blocks compute the scores of a row, each over its own \emph{span} of the row, so no thread block sees the whole row, and the selection cannot take its top-$p$ directly.
Instead, inspired by \citet{floyd1975expected} (who select from only the elements between two bounds estimated from a sample), before the scan, we set a threshold $\theta_i$ for each query, and a thread block compares each of its scores with $\theta_i$.
Then, it appends only the buckets whose score is at or above it to the \emph{list} $L_i$ of the query: one comparison per score, without seeing the rest of the row.
Only the buckets of $L_i$, a small multiple $\gamma p$ of them (see below), are then ranked.

We also evaluate an alternative inspired by the k-selection of FAISS~\citep{johnson2021billion}, which keeps the top-$p$ buckets of each span and merges them.
With this design, every thread block has to rank its span, rather than compare each score once.
Furthermore, the merge receives $p$ buckets from every span, rather than about $\gamma p$ in all. 
Overall, at 10M tokens in prefill, a row is split into 320 spans, which list 327,680 buckets, against about 1,536 in $L_i$.

We aim for about $\gamma p$ buckets of the row to score at or above $\theta_i$, with $\gamma > 1$: enough to hold the top-$p$ with room for error, and few enough to keep $L_i$ short.
A first pass, the \emph{threshold estimate}, sets $\theta_i$ from a sample of $n_s$ buckets of the row (Figure~\ref{fig:kernels-threshold}):
\begin{equation}
\begin{aligned}
\theta_i &\le \text{the $n_\theta$-th best sampled } s_{i,b} \\
&\approx \text{the $\gamma p$-th best } s_{i,b} \text{ of the row} \\
n_\theta &= \gamma \cdot p \cdot n_s / N_i
\end{aligned}
\label{eq:prefill-threshold}
\end{equation}
The sample takes about one bucket in every $N_i / n_s$ of the row, so its $n_\theta$-th best score ranks about $n_\theta N_i / n_s = \gamma p$-th in the row, and $\theta_i$ is placed at or just below it.
This estimate is noisy: for independent samples, the count of buckets above $\theta_i$ would typically differ from $\gamma p$ by about $\gamma p / \sqrt{n_\theta}$.
However, since neighbouring buckets are correlated, this is only a heuristic.
Prefill fixes $n_\theta$, so that its sample grows with the row, and decode fixes $n_s$, and derives $n_\theta$ for each request.

The noise affects only the cost of the selection.
The threshold sets how many buckets are listed, but not which are selected: when $L_i$ holds every bucket with $s_{i,b} \ge \theta_i$, and at least $p$ of them, the top-$p$ buckets of the row all reach $\theta_i$, and are thus all in $L_i$.
The estimate misses in two ways: $\theta_i$ comes out too high, and fewer than $p$ buckets score at or above it, or too low, and more score at or above it than $L_i$ has room for.
$\gamma$ leaves room for the noise on both sides, $\gamma p - p$ below and the rest of $L_i$ above.
A query that the estimate misses anyway falls back to its whole row.
In prefill, it scores all its buckets again and takes their top-$p$, the work the threshold avoids, but for one query rather than a whole chunk.
In decode, it takes the top-$p$ of the scores the scan wrote.
The fallback thus costs little as long as few queries take it.

Each bucket of $L_i$ is packed with its score into one 64-bit integer $\sigma_{i,b}$: the bits of $s_{i,b}$, in a form that keeps their order, in its upper 32 bits, and the bucket id, inverted, in its lower 32.
A larger $\sigma_{i,b}$ is thus a better score, or, at equal scores, a lower bucket id.
The \querystep{C}~\emph{expansion} finds, by radix select~\citep{alabi2012fast, zhang2023parallel} over the $\sigma_{i,b}$, the cut $\sigma_i^\star$ at which the buckets of $L_i$, taken best first, reach $k$ keys or $p$ buckets.
It writes into $J_i$ the keys of the buckets above $\sigma_i^\star$, in the order of $L_i$, then the first keys of the bucket at $\sigma_i^\star$ until $J_i$ holds $k$.
If these buckets hold fewer than $k$ keys outside the sink and the window, $J_i$ holds fewer.
These are the keys that an expansion of the buckets best first would select, in another order.
The attention does not depend on the order, so $L_i$ is never sorted.
Prefill and decode differ in where the scores go, how $\theta_i$ is set, and what a fallback costs.

\querystep{B} \querystep{C} \textbf{Flat prefill.}
In prefill, the flat search selects the buckets of all the queries of a chunk at once, in one matrix product on tensor cores, whose kernel compares each score with $\theta_i$ as soon as the score is computed, without writing it to memory.
We first show why the scores are not written, then how the product is spread over the GPU, and then the steps of Algorithm~\ref{alg:prefill-select}.

\begin{algorithm}[t]
\caption{Flat prefill selection of one chunk}
\label{alg:prefill-select}
\small
\begin{algorithmic}[1]
\REQUIRE \begin{tabular}[t]{@{}l@{}}index queries $q_i^{(\mathrm{idx})}$, $i \in \mathcal{C}$, and centroids $c_b$, in HBM \\ buckets $\mathcal{B}_i$ before the cluster block of each $i$ \\ sizes $p$, $k$, $n_{\mathrm{sink}}$, $n_w$, $n_\theta$, margin $\delta$, tile sizes $t_q$, $t_b$\end{tabular}
\ENSURE index lists $J_i$, rows of $n_{\mathrm{sink}}+k+n_w$ KV-cache slots
\algophase{Threshold estimate}
\begin{algoboxwide}{algogreen}{every tile of $t_q$ queries $\times$ part of the sample, in parallel}
\STATE\algolinelabel{line:prefill-sample} score a strided sample of the $c_b$ of $\mathcal{B}_i$, on tensor cores
\end{algoboxwide}
\STATE\algolinelabel{line:prefill-threshold} $\theta_i \gets$ a bin edge at or below the $n_\theta$-th best sampled score
\algophase{Bucket scan}
\begin{algoboxwide}{algogreen}{every tile of $t_q$ queries $\times$ span of centroid tiles, in parallel}
\STATE\algolinelabel{line:prefill-load} load the tile's $q_i^{(\mathrm{idx})}$ from HBM into shared memory
\FOR{each tile of $t_b$ centroids of the span}
\STATE\algolinelabel{line:prefill-load-centroids} load its $c_b$ from HBM into shared memory
\STATE\algolinelabel{line:prefill-product} $S_{i,b} \gets q_i^{(\mathrm{idx})} \cdot c_b$, on tensor cores \COMMENT{score tile}
\STATE\algolinelabel{line:prefill-compare} compact its $b$ with $S_{i,b} \ge \theta_i - \delta$ in shared memory
\ENDFOR
\STATE\algolinelabel{line:prefill-append} $L_i \gets L_i \cup \{\, b : S_{i,b} \ge \theta_i - \delta \,\}$, for the tile's $i$, in HBM
\end{algoboxwide}
\algophase{Candidate scoring}
\begin{algoboxwide}{algogreen}{every query $i$, in parallel}
\STATE\algolinelabel{line:prefill-rescore} $s_{i,b} \gets q_i^{(\mathrm{idx})} \cdot c_b$ in fp32, for $b \in L_i$
\STATE\algolinelabel{line:prefill-pack} $\sigma_{i,b} \gets (s_{i,b}, b)$ packed in one int64 \COMMENT{ties by bucket}
\end{algoboxwide}
\SHORTIF{\algolinelabel{line:prefill-fallback}$L_i$ overflowed, or fewer than $p$ of its $s_{i,b} \ge \theta_i$}{$L_i \gets \mathcal{B}_i$; redo lines~\ref{line:prefill-rescore}-\ref{line:prefill-pack}}
\algophase{Expansion}
\begin{algoboxwide}{algogreen}{every query $i$, in parallel}
\STATE\algolinelabel{line:prefill-radix} $\sigma_i^\star \gets \sigma_{i,b}$ where $L_i$, best first, has $k$ keys or $p$ buckets
\STATE\algolinelabel{line:prefill-write} $J_i \gets$ keys in $\mathcal{R}_i$ of $\{ b \in L_i : \sigma_{i,b} > \sigma_i^\star \}$, in listed order
\STATE\algolinelabel{line:prefill-cut} $J_i \gets J_i \cup$ first $k - \lvert J_i \rvert$ keys in $\mathcal{R}_i$ of the bucket at $\sigma_i^\star$
\STATE\algolinelabel{line:prefill-tail} $J_i \gets J_i \cup$ the sink and the window; pad $J_i$ with $-1$
\end{algoboxwide}
\end{algorithmic}
\end{algorithm}

With the $\lvert \mathcal{C} \rvert$ index queries of a chunk $\mathcal{C}$ as the rows of $Q_{\mathcal{C}}^{(\mathrm{idx})}$, and the $T / \bar{n}_b$ bucket centroids $c_b$ as the rows of $C$, the rows of the chunk form one score matrix
\begin{equation}
S = Q_{\mathcal{C}}^{(\mathrm{idx})} C^{\top}, \qquad S_{i,b} = q_i^{(\mathrm{idx})} \cdot c_b
\end{equation}
At 10M tokens, computing $S$ takes $F^{\mathrm{pre}}$ floating-point operations (flop) on at least $B_{\mathrm{read}}^{\mathrm{pre}}$ bytes of bf16 queries and centroids, and writing it takes $B_{\mathrm{scores}}^{\mathrm{pre}}$ bytes in fp32:
\begin{equation}
\begin{aligned}
F^{\mathrm{pre}} &= 2 \lvert \mathcal{C} \rvert \, d^{(\mathrm{idx})} \, T / \bar{n}_b \approx 5.5\,\text{Tflop} \\
B_{\mathrm{read}}^{\mathrm{pre}} &= 2 d^{(\mathrm{idx})} \big( T / \bar{n}_b + \lvert \mathcal{C} \rvert \big) \approx 164\,\text{MiB} \\
B_{\mathrm{scores}}^{\mathrm{pre}} &= 4 \lvert \mathcal{C} \rvert \, T / \bar{n}_b = 160\,\text{GiB}
\end{aligned}
\label{eq:prefill-flops}
\end{equation}
On an H200, at the peak of its tensor cores, computing $S$ would take at least $F^{\mathrm{pre}} / 989\,\text{Tflop/s} = 5.6$\,ms.
If every query and centroid were read from HBM only once, the products would have an arithmetic intensity of $F^{\mathrm{pre}} / B_{\mathrm{read}}^{\mathrm{pre}} \approx 32000$\,flop/B, two orders of magnitude above the ridge point of the H200. 
Under this ideal reuse, they can be bound by compute.

Writing $S$, however, is bound by memory, and would take at least $B_{\mathrm{scores}}^{\mathrm{pre}} / 4.8\,\text{TB/s} = 36$\,ms.
Both are lower bounds from the peaks of the GPU, and the second is more than six times the first at any context, as both grow with $T$.
At 9M tokens, $S$ alone would already exceed the 141\,GB of HBM of an H200.
At long context, we therefore never write $S$, and the selection writes only the lists $L_i$.

As in a matrix multiplication (GEMM), the product is \emph{tiled} (Figure~\ref{fig:kernels-scan}).
A thread block holds a tile of $t_q = 128$ index queries in shared memory, and streams the centroids through it in tiles of $t_b = 64$ (128 on B300).
It computes one $t_q \times t_b$ tile $S^{\mathrm{tile}}$ of $S$ at a time, on tensor cores, and keeps it on chip.
Each thread block takes such a tile of queries and a span of up to 64 tiles of centroids.
At 10M tokens, a row is thus split into 320 spans, and the chunk into $256 \times 320 = 81920$ thread blocks, while an H200 runs 528 at a time, 4 on each of its 132 SMs, which is enough to keep every SM busy.
At shorter contexts, the spans shrink, so that there are still enough thread blocks.

The bucket scan differs from a GEMM only in its epilogue.
Instead of writing each entry $S_{i,b}$ of $S^{\mathrm{tile}}$, the epilogue compares it with the threshold (lines~\ref{line:prefill-product}-\ref{line:prefill-compare}), and compacts the buckets at or above it in shared memory, so that each span appends them to $L_i$ in HBM with one atomic per query (line~\ref{line:prefill-append}).
With $d^{(\mathrm{idx})} = 64$, a tile takes few products, and this epilogue costs about as much as they do, so the two must overlap.
On H200, the thread block splits its warps by role: one warp loads the tiles of centroids into shared memory with the Tensor Memory Accelerator (TMA), in a pipeline of two stages, while a warpgroup multiplies them on tensor cores (WGMMA) into two accumulators in registers, and compares one with the threshold while the tensor cores fill the other.
On B300, the products accumulate in the tensor memory of the SM instead, from which the comparing threads load them.
Triton emits the same instructions, but chooses this schedule itself and gives no control of shared memory, which is why we write the scan in CuTe DSL.

At short context, below 32,768 buckets (about 256k tokens), the selection costs more than writing $S$, which then takes at most 4\,GiB, and we instead write $S$ with one matrix product in PyTorch and take the top-$p$ buckets of each row.

In prefill, we fix $n_\theta = 64$, so that the count above $\theta_i$ is off by about $\gamma p / 8$, and $\gamma = 1.5$, which sets the sample: every $(\gamma \cdot p / n_\theta) = 24$th bucket of the row, about 4\% of its scores at any context.
Algorithm~\ref{alg:prefill-select} then selects in four steps.
The \emph{threshold estimate} scores the sample twice on tensor cores, once for the mean and the best of its scores and once to count them in 64 bins between the two, binning them straight from the accumulators so that the scores of the sample are never written, and sets $\theta_i$ to the lower edge of the highest bin with at least $n_\theta$ samples at or above it (lines~\ref{line:prefill-sample}-\ref{line:prefill-threshold}).
The \emph{bucket scan} computes $S$ tile by tile, and lists the buckets whose score $S_{i,b}$ is at or above $\theta_i$ minus a margin $\delta$ (lines~\ref{line:prefill-load}-\ref{line:prefill-append}):
\begin{equation}
L_i = \{\, b \in \mathcal{B}_i : S_{i,b} \ge \theta_i - \delta \,\}
\label{eq:prefill-list}
\end{equation}
\emph{Candidate scoring} then re-scores the buckets of $L_i$, as vector search re-scores a shortlist~\citep{jegou2011searching}: it computes their score $s_{i,b}$ again in fp32, one dot product per bucket, and packs it into $\sigma_{i,b}$ (lines~\ref{line:prefill-rescore}-\ref{line:prefill-pack}).

The margin is there because the scan and candidate scoring, which multiply the same bf16 operands and both accumulate in fp32, accumulate in different orders, so $S_{i,b}$ and $s_{i,b}$ can differ: with unit-norm index queries and centroids and $d^{(\mathrm{idx})} = 64$, by less than $10^{-5}$.
The margin $\delta = 10^{-4}$ is ten times larger, so $L_i$ holds every bucket with $s_{i,b} \ge \theta_i$.
The list $L_i$ has room for $3p = 3072$ buckets, twice the target $\gamma p$.
A query whose list overflows, or holds fewer than $p$ buckets at or above $\theta_i$, falls back: it scores all its buckets again, and takes their top-$p$ (line~\ref{line:prefill-fallback}).
The \emph{expansion} writes $J_i$ (lines~\ref{line:prefill-radix}-\ref{line:prefill-tail}).

\textbf{CUDA graphs.}
A prefill chunk takes milliseconds, against which the launches of its kernels from the host cost little, and its selection reads the sizes of its lists back on the host, so we serve prefill without graphs.
A decode step of the selection, however, takes only tens of microseconds, about as long as its launches, so SGLang captures each decode step, our kernels included, in a CUDA graph, and replays it with one launch from the host.
A CUDA graph fixes the shapes of its kernels, so a graph sized for the longest context the server admits would launch the thread blocks of 10M tokens at every step, whatever the contexts of the batch.
We therefore capture one graph per range of context lengths, up to $1.125 \cdot 2^{e}$ tokens for $e = 17, \ldots, 23$ (from 144k to 9M), and one graph for the longest context, and each step replays the narrowest graph that covers the longest request of the batch.
The threshold estimate of the flat search samples across the width of the graph, so a request far shorter than its range would see too few samples and fall back to its full row.
The factor 1.125 avoids this for the common case: a request of a power-of-two length, with its answer, sits at about 89\% of its range.
The clustering of a complete cluster block runs outside the graph, before the replay, once every $n_{\mathrm{cluster}} = 512$ decoded tokens, and the decode times of Section~\ref{sec:performance} spread its cost over those tokens.

\begin{algorithm}[t]
\caption{Flat decode selection of one request}
\label{alg:decode-select}
\small
\begin{algorithmic}[1]
\REQUIRE \begin{tabular}[t]{@{}l@{}}index query $q_i^{(\mathrm{idx})}$ and centroids $c_b$ of $\mathcal{B}_i$, in HBM \\ the request's pages of $c_b$ \\ sizes $p$, $k$, $n_{\mathrm{sink}}$, $n_w$, $n_s$, $n_\theta$\end{tabular}
\ENSURE index list $J_i$, as in Algorithm~\ref{alg:prefill-select}
\algophase{Threshold estimate}
\begin{algoboxwide}{algogreen}{every part of the sample, in parallel}
\STATE\algolinelabel{line:decode-sample} score a strided sample of the $c_b$, on CUDA cores
\end{algoboxwide}
\STATE\algolinelabel{line:decode-threshold} $\theta_i \gets$ the $n_\theta$-th best sampled score, rounded down
\algophase{Bucket scan}
\begin{algoboxwide}{algogreen}{every span of the request's buckets, in parallel}
\STATE\algolinelabel{line:decode-read} read each $c_b$ once from HBM, through the request's pages
\STATE\algolinelabel{line:decode-score} $s_{i,b} \gets q_i^{(\mathrm{idx})} \cdot c_b$ in fp32, into shared memory and HBM
\STATE\algolinelabel{line:decode-wait} wait for $\theta_i$
\STATE\algolinelabel{line:decode-append} $L_i \gets L_i \cup \{\, \sigma_{i,b} : s_{i,b} \ge \theta_i \,\}$, for $b$ in the span, in HBM
\end{algoboxwide}
\algophase{Listing}
\STATE\algolinelabel{line:decode-order} gather the spans of $L_i$, in bucket order
\SHORTIF{\algolinelabel{line:decode-fallback}$L_i$ overflowed, or fewer than $p$ of its $s_{i,b} \ge \theta_i$}{$L_i \gets$ top-$p$ of the written $s_{i,b}$, as $\sigma_{i,b}$}
\algophase{Expansion}
\STATE\algolinelabel{line:decode-expand} $J_i \gets$ lines~\ref{line:prefill-radix}-\ref{line:prefill-tail} of Algorithm~\ref{alg:prefill-select} on $L_i$
\end{algorithmic}
\end{algorithm}

\querystep{B} \querystep{C} \textbf{Flat decode.}
In decode, each request adds one new query per step, whose buckets the flat search selects.
It does so in a single pass, which reads the centroids of the request once, scores them, and lists $L_i$ as it goes.
We first show why this one read sets its time, then how the pass is spread over the GPU, and then the steps of Algorithm~\ref{alg:decode-select}.

In decode, $S$ is a single row: the one index query $q_i^{(\mathrm{idx})}$ of a request scores each of its $T / \bar{n}_b$ centroids $c_b$ once per step.
With $\lvert \mathcal{C} \rvert = 1$ in Equation~\ref{eq:prefill-flops}, a step at 10M tokens takes $F^{\mathrm{dec}}$ flop on at least $B_{\mathrm{read}}^{\mathrm{dec}}$ bytes, and writing its row takes $B_{\mathrm{scores}}^{\mathrm{dec}}$ bytes in fp32:
\begin{equation}
\begin{aligned}
F^{\mathrm{dec}} &= 2 d^{(\mathrm{idx})} \, T / \bar{n}_b \approx 168\,\text{Mflop} \\
B_{\mathrm{read}}^{\mathrm{dec}} &= 2 d^{(\mathrm{idx})} \big( T / \bar{n}_b + 1 \big) \approx 160\,\text{MiB} \\
B_{\mathrm{scores}}^{\mathrm{dec}} &= 4 \, T / \bar{n}_b = 5\,\text{MiB}
\end{aligned}
\label{eq:decode-flops}
\end{equation}
Its arithmetic intensity $F^{\mathrm{dec}} / B_{\mathrm{read}}^{\mathrm{dec}} \approx 1$\,flop/B is far below the ridge point: the selection is bound by memory, and reading the centroids once would take at least $B_{\mathrm{read}}^{\mathrm{dec}} / 4.8\,\text{TB/s} = 35$\,\textmu s on H200.
Prefill, bound by compute, can afford to read the centroids more than once: every tile of queries reads them all, and candidate scoring reads those of $L_i$ again.
In decode, a second read would about double the bytes the scan reads, so Algorithm~\ref{alg:decode-select} selects within a single read of the centroids, besides the small sample of its threshold estimate.
It can, however, afford to write every $s_{i,b}$ to HBM, which adds only 3\% to the bytes it moves, so that its fallback takes the top-$p$ of the written scores, without reading the centroids again.
The threshold is thus not strictly needed in decode, but it saves a selection over all $N_i$ written scores of each request: the expansion ranks only the buckets of $L_i$.

With a single query, only the row is split: the scan takes spans of 256 buckets, one per thread block.
The centroids of a request lie in pages of $B_{\mathrm{cache}}^{(\mathrm{idx})}$, as its keys lie in pages of the KV cache, and each thread block reads the $c_b$ of its span through the page table of the request.
The threshold estimate and the scan score on the CUDA cores, with the same dot product: at 1\,flop/B the scan is bound by HBM, not by arithmetic, and gains nothing from tensor cores.
A request whose row fits in its list takes every bucket.

Each CUDA graph samples a fixed number $n_s$ of buckets, from 2,048 to 16,384, in runs of 8 neighbouring buckets of a page, spread evenly over the row.
A graph that covers longer contexts thus samples a smaller share of the row, with a smaller $n_\theta$ and a noisier estimate, so the target $\gamma p$ grows with the graph, from $1.5p$ to $3p$.
We set $\gamma$ for each graph by hand, so that $L_i$ still holds at least $p$ buckets, and fits in its buffer, when the count misses $\gamma p$ by four times the deviation above, $4\gamma p / \sqrt{n_\theta}$.
Algorithm~\ref{alg:decode-select} runs in four launches, followed by the expansion, and has no candidate scoring.
The \emph{threshold estimate} takes two launches: one scores the sample, and one finds its $n_\theta$-th best score by radix select over the bits of the scores, and rounds it down (lines~\ref{line:decode-sample}-\ref{line:decode-threshold}).
The \emph{bucket scan} writes the $s_{i,b}$ of each span, and appends the buckets with $s_{i,b} \ge \theta_i$ to $L_i$, already packed as $\sigma_{i,b}$ (lines~\ref{line:decode-read}-\ref{line:decode-append}).
It compares the very scores that rank the buckets, so it needs no margin.
The \emph{listing} gathers into $L_i$ the spans that the thread blocks appended, in bucket order (line~\ref{line:decode-order}), and the \emph{expansion} is that of Algorithm~\ref{alg:prefill-select} (line~\ref{line:decode-expand}).
The list $L_i$ has room for 2,048 to 8,192 buckets, by graph, and a request whose list overflows, or holds fewer than $p$ buckets, falls back to the top-$p$ of the scores the scan wrote (line~\ref{line:decode-fallback}).
Programmatic dependent launch~\citep{nvidia2026pdl}, a feature of Hopper and Blackwell GPUs, lets a kernel start while the one before it still runs, and wait for its results only where it reads them (line~\ref{line:decode-wait}).
The scan thus reads and scores the centroids while $\theta_i$ is still being computed.

\querystep{B} \textbf{Tree prefill.}
In prefill, the beam search (Algorithm~\ref{alg:beam}) starts at the lowest level with at most $n_{\mathrm{start}} = 32768$ node buckets, a value we chose by a sweep of accuracy against time.
It keeps a beam of $p_\ell = 2048$ node buckets on every level.
The descent starts from every node bucket of this level (line~\ref{line:beam-start}), which one matrix product in cuBLAS scores in full.
Each level of the descent (lines~\ref{line:beam-while}-\ref{line:beam-endwhile}) then runs in one kernel (five or six launches in Triton), besides a matrix product in cuBLAS that scores the open node buckets of the level.
In this kernel, a thread block holds the scores of one query in shared memory, one 32-bit word per candidate, up to the 227\,KiB a thread block can take.
It scores the children of the node buckets kept on the level above (line~\ref{line:beam-score}).
It then finds the $p_\ell$-th score by a radix select of four passes, each of which counts one byte of the words in a histogram of 256 bins (line~\ref{line:beam-keep}).
Finally, it writes the children of the node buckets it keeps as the candidates $C$ of the next level (line~\ref{line:beam-children}).
Above about 24k node buckets, keeping the beam of the start level this way would leave a single thread block per SM, so it takes two launches instead: a listing of the node buckets above a sampled threshold, as in the flat search, and the keep of the beam.
At level 0 (lines~\ref{line:beam-level0}-\ref{line:beam-topp}), the thread block writes the top-$p$ buckets as packed scores $\sigma_{i,b}$.
The expansion of Algorithm~\ref{alg:prefill-select} (lines~\ref{line:prefill-radix}-\ref{line:prefill-tail}) reads them as it reads $L_i$.

\querystep{B} \textbf{Tree decode.}
In decode, the new queries of all the requests of a batch descend together, through the beam kernels of the prefill, but over a copy of the hierarchy at fixed addresses, and at shapes set by the longest context, so that a CUDA graph can replay the step.
Each request adds one query per step, but the descent still runs its levels one after another, so its time barely changes with the number of requests, while the flat search scans every bucket once per request.
Every query scores the levels from the top level of the longest context down, and holds no candidates above its own start level.
A step takes 7 to 13 kernels from 128k to 10M tokens, as the number of levels grows with the context.

\keystep{1} \keystep{2} \keystep{4} \querystep{A} \textbf{Other steps.}
The remaining steps of \dname{} are the projections and the L2 normalisation of the index keys $k_j^{(\mathrm{idx})}$ and queries $q_i^{(\mathrm{idx})}$, the pooling, and the tables of $B_{\mathrm{cache}}^{(\mathrm{idx})}$.
They consist of matrix products, reductions and a sort, which run as standard PyTorch operations, with the products through cuBLAS.
These steps cost little next to the scoring.
On \glm{}, for example, each of the two projections of a chunk of 32,768 tokens from the 512 dimensions of the latent takes $2 \cdot 32768 \cdot 512 \cdot 64$ flop, 4.3\,Gflop for both.
This is under 1\% of the scoring of the chunk at contexts of 1M tokens and beyond (Equation~\ref{eq:prefill-flops}).

\querystep{C} \querystep{D} \textbf{Attention kernels.}
\dname{} attends through existing kernels, which only need to read the index lists in a layout of their own.
The expansion writes the index list $J_i$ of every query $i$ as a row of $n_{\mathrm{sink}} + k + n_w = 4612$ int32 entries.
Each entry is the slot in the KV cache of one key $j \in J_i$, and the row is padded with $-1$ when the query has fewer keys to attend to.
The index lists of a chunk $\mathcal{C}$ thus form one matrix of $\lvert \mathcal{C} \rvert$ rows.
On \glm{}, $J_i$ takes the place of the index list of the model's own indexer, and the sparse attention of SGLang attends over it unchanged.
On \qwen{}, we pass every query $i$ to a paged attention kernel as a sequence of its own: its row is the page table, each page holds one slot, and the sequence length is $\lvert J_i \rvert$.
Table~\ref{tab:attention-kernels} lists the attention kernels we use, from FlashAttention-3 (FA3)~\citep{shah2024flashattention} and its successor FA4, FlashMLA~\citep{flashmla2025}, the TRT-LLM kernels of FlashInfer~\citep{ye2025flashinfer} and Triton.
Each is the default of SGLang~\citep{zheng2024sglang} for its model and GPU, except where that default cannot run.
For instance, on B300, \dname{} on \qwen{} attends through the Triton kernel of SGLang, as FlashAttention supports pages of a single slot only on Hopper.
With CP8, \glm{} decodes through a split-KV sparse kernel that we add to SGLang (Appendix~\ref{app:parallelism}), and on B300 it prefills through a TileLang kernel~\citep{wang2025tilelang}.
With decode context parallelism (DCP8, Appendix~\ref{app:parallelism}) on H200, Full Attention and Block Sparse Attention (Flat) on \qwen{} decode through Triton kernels.

\begin{table}[t]
\centering
\caption{\textbf{Attention kernels.}
Kernel through which each method attends, by model, phase and GPU, at TP8 or on one GPU.
\boxfull~Full Attention, \boxblockflat~Block Sparse Attention (Flat), \boxlsaflat~\boxlsatree~\dname{} (Flat and Tree).
Kernels in italics are non-default SGLang options.}
\label{tab:attention-kernels}
\vskip 0.15in
\setlength{\tabcolsep}{4pt}
\ssmall
\begin{tabular*}{\columnwidth}{@{\extracolsep{\fill}}l!{\separator}ccc!{\separator}ccc@{}}
\toprule
 & \multicolumn{3}{c!{\separator}}{H200} & \multicolumn{3}{c}{B300} \\
\cmidrule(lr){2-4} \cmidrule(l){5-7}
 & \boxfull & \boxblockflat & \boxlsaflat\,\boxlsatree & \boxfull & \boxblockflat & \boxlsaflat\,\boxlsatree \\
\midrule
\multicolumn{7}{@{}l}{\textbf{\glm{}}} \\
\addlinespace[1.5pt]
Prefill & FA3 & FA3 & FA3 & TRT-LLM & TRT-LLM & TRT-LLM \\
Decode & \textit{FlashMLA} & FA3 & FA3 & \textit{Triton} & TRT-LLM & TRT-LLM \\
\midrule
\multicolumn{7}{@{}l}{\textbf{\qwen{}}} \\
\addlinespace[1.5pt]
Prefill & FA3 & Triton & FA3 & TRT-LLM & Triton & \textit{Triton} \\
Decode & FA3 & Triton + FA4 & FA3 & TRT-LLM & Triton + TRT-LLM & \textit{Triton} \\
\bottomrule
\end{tabular*}
\end{table}

\textbf{\nname{} training.}
The kernels above serve \dname{} at inference.
\nname{} also needs kernels for training, for its forward and its backward pass, with which a layer is up to 1.6$\times$ faster than with Full Attention at 128k tokens (Section~\ref{sec:native}).

A few changes stem from the architectural differences of \nname{} compared to \dname{}.
A training step processes whole sequences, of up to 128k tokens in our runs, with no KV cache and no decode, so every query of a sequence selects at once, as in a prefill chunk as long as the sequence.
The keys change at every step as the attention projections learn, so every step clusters all keys again, where \dname{} clusters each cluster block once and keeps its buckets in the cache.
The selection clusters and scores the attention's own normalised keys and queries, of dimension 128, and takes the best score of the 8 query heads (Equation~\ref{eq:native-score}).
At 128k tokens, a row holds at most 16,384 buckets, from which a query selects $p = 512$ for $k = 2048$ keys, next to a window of 512 keys and 4 sink keys.
Finally, every step runs a backward pass, which \dname{} never does at long context.

Two of these differences shape the kernels.
As the rows are short, the selection places $\theta_i$ from the whole row rather than from a sample: a first pass over the row counts its scores in a histogram of 256 bins, and a second lists the buckets from the bin of the $p$-th score up, both on tensor cores, and the top-$p$ of the list are then taken by their scores.
The backward pass is the main addition.
As \nname{} stops the gradient through the selection, the clustering and the scoring of the centroids need no backward pass, and the only new backward kernel is that of the sparse attention.
For one query head, with logits $z_{i,j} = q_i \cdot k_j / \sqrt{d_h}$, its forward pass, as in any attention kernel, runs over the queries, each query $i$ attending to the keys it selected and writing only its own output:
\begin{equation}
o_i = \sum_{j \in J_i} \alpha_{i,j} v_j, \qquad \alpha_{i,j} = \mathrm{softmax}_{j \in J_i}\left(z_{i,j}\right)
\end{equation}
Its backward pass cannot run over the queries in the same way, as each key $j$ and its value receive a gradient of the NTP loss $\mathcal{L}$ from every query that selected the key:
\begin{equation}
\frac{\partial \mathcal{L}}{\partial k_j} = \sum_{i \in I_j} \frac{\partial \mathcal{L}}{\partial z_{i,j}} \frac{q_i}{\sqrt{d_h}}, \qquad
\frac{\partial \mathcal{L}}{\partial v_j} = \sum_{i \in I_j} \alpha_{i,j} \frac{\partial \mathcal{L}}{\partial o_i}
\label{eq:nlsa-backward}
\end{equation}
where $I_j = \{\, i : j \in J_i \,\}$, and the queries of $I_j$ lie anywhere in the sequence at or after $j$.
As the 8 query heads share one key-value head, $\partial \mathcal{L} / \partial k_j$ and $\partial \mathcal{L} / \partial v_j$ also sum over the heads.
Run over the queries, the backward pass would have many thread blocks add into the gradient of the same key, with an atomic addition for each share.
The backward kernels instead run over the keys, as the backward pass of FlashAttention~\citep{dao2022flashattention} does for dense attention (Algorithm~\ref{alg:nlsa-backward}).
With a sparse selection, this first requires inverting the selection, listing the queries of each group of keys (lines~\ref{line:nlsa-count}-\ref{line:nlsa-invert}).
A thread block then holds the gradients of a group of keys in registers, adds the shares of these queries into them (lines~\ref{line:nlsa-hold}-\ref{line:nlsa-accumulate}), and writes them with few atomic additions (line~\ref{line:nlsa-write}).
The gradient of each query sums over its own $J_i$, so a separate pass gathers it as in the forward pass, computing $\partial \mathcal{L} / \partial z_{i,j}$ again (line~\ref{line:nlsa-dq}).
To invert it, the backward pass needs the selection of every query, the $k$ int32 long-range indices that the forward pass keeps, as the sink and the window follow from the positions: 1\,GiB at 128k tokens, about the gap between the peak memory of \nname{} and that of Full Attention.
Evaluating and implementing a port of these kernels to CuTe DSL, as for \dname{}, is left to future work.

With these kernels, Section~\ref{sec:native} times one attention layer of \nname{} on one H200, at batch 1 in bf16, against the same layer with causal attention in FA3.
Each time covers the forward and the backward pass of the whole layer, from its input projections to its output projection, with the clustering and the selection.

\begin{algorithm}[t]
\caption{Sparse attention backward of \nname{}, for one query head}
\label{alg:nlsa-backward}
\small
\begin{algorithmic}[1]
\REQUIRE \begin{tabular}[t]{@{}l@{}}$q_i$, $k_j$, $v_j$, $o_i$ and $\mathrm{d}o_i$, in HBM \\ log-sum-exp $\lambda_i$ and the $k$ long-range indices of $J_i$, \\ kept by the forward pass\end{tabular}
\ENSURE $\mathrm{d}q_i$, $\mathrm{d}k_j$, $\mathrm{d}v_j$, writing $\mathrm{d}x$ for $\partial \mathcal{L} / \partial x$
\algophase{Inversion}
\STATE\algolinelabel{line:nlsa-count} count, for each group $G$ of keys, the queries $i$ with $J_i \cap G \neq \emptyset$
\STATE\algolinelabel{line:nlsa-invert} $I_G \gets \bigcup_{j \in G} I_j$, laid out by a prefix sum over the counts
\algophase{Key and value gradients}
\begin{algoboxwide}{algogreen}{every query $i$, in parallel}
\STATE\algolinelabel{line:nlsa-delta} $D_i \gets \mathrm{d}o_i \cdot o_i$
\end{algoboxwide}
\begin{algoboxwide}{algogreen}{every group $G$ of keys, in parallel}
\STATE\algolinelabel{line:nlsa-hold} hold $k_j$, $v_j$, $j \in G$, in registers; $\mathrm{d}k_j, \mathrm{d}v_j \gets 0$
\FOR{each query $i \in I_G$ and key $j \in J_i \cap G$}
\STATE\algolinelabel{line:nlsa-recompute} $z_{i,j} \gets q_i \cdot k_j / \sqrt{d_h}$; \ $\alpha_{i,j} \gets \exp(z_{i,j} - \lambda_i)$
\STATE\algolinelabel{line:nlsa-dv} $\mathrm{d}v_j \gets \mathrm{d}v_j + \alpha_{i,j} \, \mathrm{d}o_i$
\STATE\algolinelabel{line:nlsa-dz} $\mathrm{d}z_{i,j} \gets \alpha_{i,j} \big( \mathrm{d}o_i \cdot v_j - D_i \big)$
\STATE\algolinelabel{line:nlsa-accumulate} $\mathrm{d}k_j \gets \mathrm{d}k_j + \mathrm{d}z_{i,j} \, q_i / \sqrt{d_h}$ \COMMENT{Equation~\ref{eq:nlsa-backward}}
\ENDFOR
\STATE\algolinelabel{line:nlsa-write} add $\mathrm{d}k_j$, $\mathrm{d}v_j$ into zeroed HBM, with few atomic additions
\end{algoboxwide}
\algophase{Query gradients}
\begin{algoboxwide}{algogreen}{every query $i$, in parallel}
\STATE\algolinelabel{line:nlsa-dq} $\mathrm{d}q_i \gets \sum_{j \in J_i} \mathrm{d}z_{i,j} \, k_j / \sqrt{d_h}$ \COMMENT{$\mathrm{d}z_{i,j}$ computed again}
\end{algoboxwide}
\end{algorithmic}
\end{algorithm}

\begin{table}[t]
\centering
\caption{\textbf{Indexer kernels.}
Time of each step of the indexer of \dname{} on one H200, with its CuTe DSL and its Triton kernels, and the regime of its main CuTe DSL kernel.}
\label{tab:kernels}
\vskip 0.15in
\setlength{\tabcolsep}{6pt}
\ssmall
\begin{tabular}{@{}l!{\separator}rrr!{\separator}l@{}}
\toprule
Step & CuTe DSL (ms) & Triton (ms) & Speedup & Regime \\
\midrule
\multicolumn{5}{@{}l}{\textit{Flat prefill}} \\
Threshold estimate & 2.02 & 8.85 & 4.38$\times$ & compute (28\%) \\
Bucket scan & 11.39 & 24.82 & 2.18$\times$ & compute (49\%) \\
Candidate scoring & 1.63 & 1.66 & 1.02$\times$ & memory (85\%) \\
Expansion & 2.37 & 4.14 & 1.75$\times$ & selection \\
\textbf{Total} & \textbf{17.41} & \textbf{39.48} & \textbf{2.27$\times$} & \\
\midrule
\multicolumn{5}{@{}l}{\textit{Tree prefill}} \\
Start level & 5.28 & 5.80 & 1.10$\times$ & selection \\
One beam level & 6.25 & 7.85 & 1.26$\times$ & selection \\
Level 0 (top-$p$) & 9.13 & 9.59 & 1.05$\times$ & memory (66\%) \\
\midrule
\multicolumn{5}{@{}l}{\textit{Clustering}} \\
Ward (buckets) & 0.043 & 0.056 & 1.30$\times$ & latency \\
Node buckets & 0.405 & 1.387 & 3.42$\times$ & latency \\
\midrule
\multicolumn{5}{@{}l}{\textit{Flat decode}} \\
Decode select & 0.071 & 0.153 & 2.15$\times$ & memory (75\%) \\
\bottomrule
\end{tabular}
\end{table}

\begin{table}[t]
\centering
\caption{\textbf{Tree search in decode.}
Time of the selection of one layer for one decode step, from the scoring of the centroids to the index lists, in \textmu s, with the flat and the tree search, on one H200, replayed from a CUDA graph.
The faster search is in bold.}
\label{tab:tree-decode}
\vskip 0.15in
\setlength{\tabcolsep}{4.5pt}
\ssmall
\begin{tabular}{@{}c!{\separator}cc!{\separator}cc!{\separator}cc@{}}
\toprule
 & \multicolumn{2}{c}{\textbf{128k}} & \multicolumn{2}{c}{\textbf{1M}} & \multicolumn{2}{c}{\textbf{10M}} \\
\cmidrule(lr){2-3} \cmidrule(lr){4-5} \cmidrule(l){6-7}
Requests & Flat & Tree & Flat & Tree & Flat & Tree \\
\midrule
1 & \textbf{30} & 38 & \textbf{35} & 49 & 86 & \textbf{68} \\
4 & \textbf{32} & 40 & 56 & \textbf{54} & 211 & \textbf{81} \\
16 & \textbf{42} & 47 & 109 & \textbf{82} & 700 & \textbf{125} \\
\bottomrule
\end{tabular}
\end{table}

\textbf{Measurements.}
Table~\ref{tab:kernels} times each step on one H200 with Nsight Compute, in a single profile with the clocks unlocked, with its CuTe DSL and its Triton kernels.
The inputs are synthetic, at the shapes we serve: a prefill chunk of 32,768 queries at 10M tokens, a beam of $p_\ell = 2048$ over node buckets of $n_{b,\ell} = 4$, decode with one request at 10M tokens, the Ward clustering of the 64 cluster blocks of one prefill chunk, and the clustering of one node block of $n_{\mathrm{cluster},\ell} = 4096$ buckets into node buckets.
The times are those of the kernels alone, without the PyTorch launches between them.
The regime of a step is where its main CuTe DSL kernel sits: bound by compute or by memory, with its share of the bf16 peak or of the HBM bandwidth of Table~\ref{tab:hardware}, bound by neither as a radix select over packed integer keys (selection), or limited by its serial rounds (latency).
The bucket scan, the largest step of the flat prefill, reaches 49\% of the bf16 peak and uses 1\% of the HBM bandwidth, so the tiles of queries reread the centroids mostly from the L2 cache.
On real inputs, the index queries of 12 held-out documents of 1M tokens and 3 books of 10M tokens on \glm{}, over all its sparse layers, $L_i$ holds 1,679 buckets on average at 1M tokens and 1,737 at 10M, never overflows, and falls back, with fewer than $p$ buckets above $\theta_i$, for $5 \times 10^{-5}$ of the queries at 1M and $10^{-3}$ at 10M.
In decode, $3 \times 10^{-4}$ of the steps fall back at 1M, mostly as their lists overflow, and none at 10M.
The decode scan reads at 3.58\,TB/s, 75\% of the HBM bandwidth.
At 10M tokens, the CuTe DSL kernels are faster than the Triton ones in every step, from 1.02$\times$ for candidate scoring, which is bound by the HBM bandwidth in both languages, to 4.4$\times$ for the threshold estimate.
On B300, they are faster by up to 4.3$\times$ in every step but level 0 of the tree search, where the two languages tie.
Of the 22\,ms that CuTe DSL saves on a prefill chunk at 10M tokens, 13.4\,ms come from the bucket scan, mostly from overlapping its epilogue with its products, and 6.8\,ms from the threshold estimate, whose Triton version writes the scores of its sample to HBM and reads them twice.
The Triton versions of other steps take more launches or more passes: the node-bucket clustering takes three launches per round and reads two flags on the host every eight rounds (3.4$\times$), and the expansion finds its cut one bit per pass rather than one byte (1.75$\times$).
Table~\ref{tab:kernels} sums the kernel times of the four launches of the decode select (the sample, the threshold, the scan and the listing), 71\,\textmu s at 10M tokens.
Replayed from a CUDA graph, where programmatic dependent launch overlaps them, they take 14.7\,\textmu s at 1M tokens and 60.7\,\textmu s at 10M, against 36.4\,\textmu s and 146.6\,\textmu s in Triton.
With the expansion and the other kernels of the selection, a flat decode step selects in 35\,\textmu s at 1M tokens and 86\,\textmu s at 10M, and Table~\ref{tab:tree-decode} compares it with the tree search.
With one request, the flat search is faster up to about 5M tokens.
With 4 and 16 requests, the tree search is faster from about 1M and 256k tokens, and with 16 requests it is 5.6$\times$ faster at 10M tokens.
As we evaluate one request at a time, every result of \dname{} (Tree) decodes with the flat search, and only its prefill uses the tree.
At 1M tokens, the four steps of the flat prefill take 5.2\,ms in CuTe DSL against 9.4\,ms in Triton.
For a whole layer of \glm{} on H200, CuTe DSL makes a prefill chunk at 10M tokens 1.35$\times$ faster than Triton.

\subsection{Parallelism}
\label{app:parallelism}

\begin{figure*}[t]
\centering
\includegraphics[width=\textwidth]{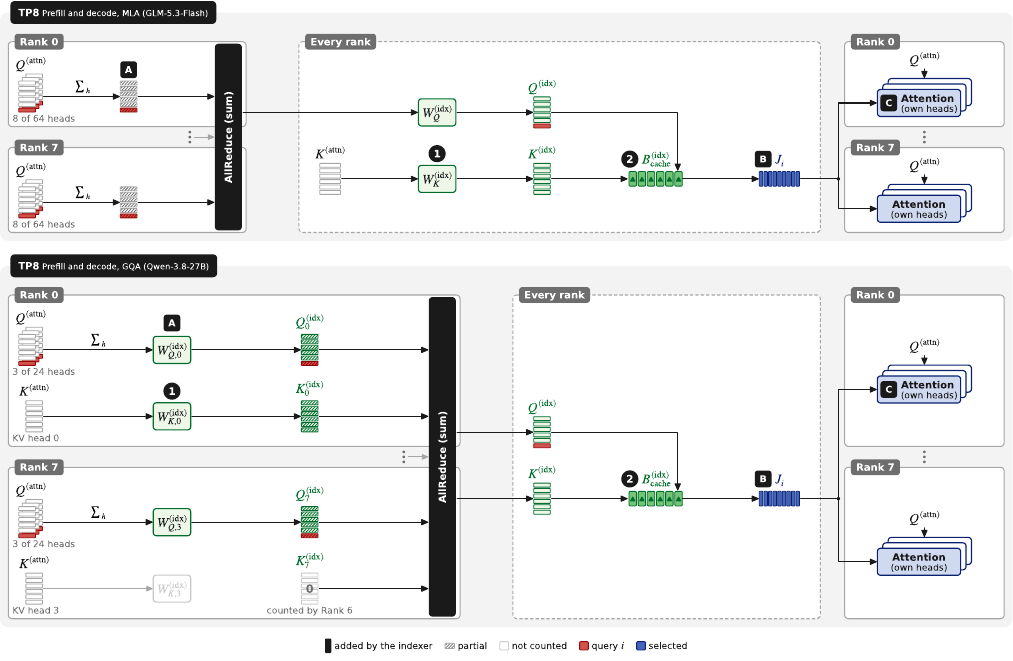}
\caption{\textbf{\dname{} with tensor parallelism (TP8).} Top: MLA (\glm{}). Bottom: GQA (\qwen{}). Indexer keys: \keystep{1}~projection (from the replicated latents with MLA, a partial per KV head with GQA), \keystep{2}~bucket cache (the same on every rank). Indexer queries: \querystep{A}~sum over the rank's query heads and projection (a partial, summed across ranks by one \coll{AllReduce}, before the projection with MLA), \querystep{B}~selecting the index list (the same on every rank), \querystep{C}~attention over the rank's own query heads.}
\label{fig:parallelism-tp}
\end{figure*}

We describe our integration of \dname{} with tensor (TP) and context parallelism (CP) in SGLang, for multi-head latent attention (MLA), as in \glm{}, and for grouped-query attention (GQA), as in \qwen{}.
$N$ denotes the number of ranks of a parallel group, and $r$ one of them.
We study setups with $N = 8$ (size of our NVLink domain).

\name{} is orthogonal to expert and pipeline parallelism: the former splits only the experts, and the latter keeps the index of a layer co-located with its attention on the same rank, so in this section we do not describe these two mesh configurations, but focus on TP and CP instead.

For both TP and CP, every rank holds the whole bucket cache $B_{\mathrm{cache}}^{(\mathrm{idx})}$, which trades memory for communication (Section~\ref{sec:lsa}).
Table~\ref{tab:parallelism} summarises the communication that the indexer adds for the two mesh configurations. We discuss these in more detail below.

\begin{table}[t]
\centering
\caption{\textbf{Communication added by the indexer.}
Collectives that the indexer of \dname{} adds to one layer for TP alone, PCP and DCP, with MLA (\glm{}) and GQA (\qwen{}), beyond those of the attention.
Sizes are per rank, and decode frequencies per request.}
\label{tab:parallelism}
\vskip 0.15in
\setlength{\tabcolsep}{2.8pt}
\ssmall
\begin{tabularx}{\columnwidth}{@{}ll!{\separator}>{\raggedright\arraybackslash}X!{\separator}ll@{}}
\toprule
 & Phase & Collective & Size & Frequency \\
\midrule
TP, MLA & both & \coll{AllReduce} of query head sums & 2\,KiB / token & every chunk and step \\
TP, GQA & both & \coll{AllReduce} of partial index queries and keys & 512\,B / token & every chunk and step \\
PCP, MLA & prefill & - & - & - \\
DCP & decode & \coll{AllReduce} of index keys & 64\,KiB / block & every $n_{\mathrm{cluster}}$ steps \\
\bottomrule
\end{tabularx}
\end{table}

\textbf{Tensor parallelism.}
We first describe TP alone, without CP, as illustrated in Figure~\ref{fig:parallelism-tp}. 
For TP, work gets split by head: 
\querystep{A}~rank $r$ holds a set $\mathcal{Q}_r$ of $n_h/N$ query heads, and projects them into a partial index query $q_{i,r}^{(\mathrm{idx})}$:
\begin{equation}
q_{i,r}^{(\mathrm{idx})} = \sum_{h \in \mathcal{Q}_r} q_{i,h}^{(\mathrm{attn})} W_Q^{(\mathrm{idx})}
\label{eq:tp-query}
\end{equation}
With MLA, the ranks sum the query heads $q_{i,h}^{(\mathrm{attn})}$ on the latent instead, and apply $W_Q^{(\mathrm{idx})}$ to the sum after the \coll{AllReduce} below, which is equivalent by linearity.
\keystep{1}~With MLA, the latent $k_j^{(\mathrm{attn})}$ is shared by all heads, thus replicated on every rank.
With GQA, the KV heads are split across ranks too, and an index key $k_j^{(\mathrm{idx})}$ is projected from all of them, so rank $r$ projects the keys $k_{j,g}^{(\mathrm{attn})}$ of the set $\mathcal{K}_r$ of KV heads it contributes into a partial index key $k_{j,r}^{(\mathrm{idx})}$:
\begin{equation}
k_{j,r}^{(\mathrm{idx})} = \sum_{g \in \mathcal{K}_r} k_{j,g}^{(\mathrm{attn})} W_{K,g}^{(\mathrm{idx})}
\label{eq:tp-key}
\end{equation}
where $W_{K,g}^{(\mathrm{idx})}$ are the rows of $W_K^{(\mathrm{idx})}$ that project KV head $g$.
Likewise, query head $h$ uses only the rows $W_{Q,g(h)}^{(\mathrm{idx})}$ of the KV head $g(h)$ it attends, in place of $W_Q^{(\mathrm{idx})}$ in Equation~\ref{eq:tp-query}.

For TP8, the four KV heads of \qwen{} are each replicated on two ranks: both contribute their own query heads, but only one counts the KV head in its $\mathcal{K}_r$, and the other contributes $k_{j,r}^{(\mathrm{idx})} = 0$.
One \coll{AllReduce} sums the partials, in fp32, into the projections of Equation~\ref{eq:projection}, before their normalisation:
\begin{equation}
q_i^{(\mathrm{idx})} = \sum_{r} q_{i,r}^{(\mathrm{idx})}, \qquad k_j^{(\mathrm{idx})} = \sum_{r} k_{j,r}^{(\mathrm{idx})}
\label{eq:par-sum}
\end{equation}
\querystep{B}~Every rank thus holds the same index query $q_i^{(\mathrm{idx})}$ and selects the same $J_i$ from its copy of $B_{\mathrm{cache}}^{(\mathrm{idx})}$.
\querystep{C}~It attends over $J_i$ with its own query heads $\mathcal{Q}_r$.
\keystep{2}~The clustering takes the index keys $k_j^{(\mathrm{idx})}$ of a cluster block only once all its $n_{\mathrm{cluster}}$ keys have arrived, over several chunks or decode steps.
With MLA, a rank recomputes them from the latents $k_j^{(\mathrm{attn})}$ in its KV cache, so they are never stored.
With GQA, a rank holds only some KV heads, so it cannot recompute them alone.
Instead, it stores $k_j^{(\mathrm{idx})}$ once summed: the \coll{AllReduce} of Equation~\ref{eq:par-sum} carries the partial key $k_{j,r}^{(\mathrm{idx})}$ of a token together with the token's own partial query, so storing it costs memory, $2 d^{(\mathrm{idx})} = 128$\,B per token, but no communication.
Recomputing it would save that memory, at the cost of a second \coll{AllReduce} of the fp32 partial keys, 256\,B per token, once each block is complete.

With MLA, the \coll{AllReduce} carries the 512 dimensions of the latent, 2\,KiB.
Since the KV cache is replicated, $B_{\mathrm{cache}}^{(\mathrm{idx})}$ adds the same 24.5\,B per token to the 1\,KiB of KV cache as without parallelism.

With GQA, the \coll{AllReduce} carries the partial index query and key, 512\,B.
A rank of \qwen{} holds one of its four KV heads, 1\,KiB, to which the indexer adds 24.5\,B of bucket cache and $2 d^{(\mathrm{idx})} = 128$\,B of stored index key.

\begin{figure*}[t]
\centering
\includegraphics[width=\textwidth]{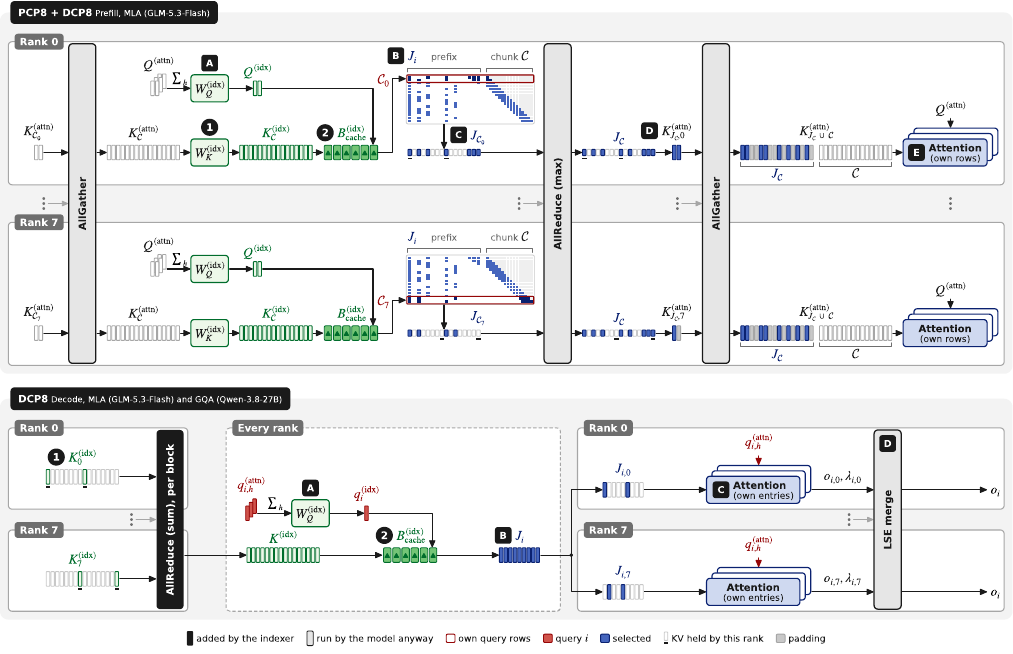}
\caption{\textbf{\dname{} with context parallelism (CP8).} Top: prefill of \glm{} (PCP8 + DCP8), each rank holding a contiguous span of the query rows. Indexer keys: \keystep{1}~projection of the whole chunk (on every rank), \keystep{2}~bucket cache (the same on every rank). Indexer queries: \querystep{A}~projection of the rank's own rows, \querystep{B}~selecting their index lists (other ranks' rows lighter), \querystep{C}~marking the selected prefix rows (a 1\,B mask, merged across ranks), \querystep{D}~gathering those rows from the ranks that hold them (padded), \querystep{E}~attention of the rank's own rows. Bottom: decode of \glm{} (PCP8 + DCP8) and \qwen{} (TP8 + DCP8). Indexer keys: \keystep{1}~projection of a complete cluster block (summed across ranks, once per block), \keystep{2}~bucket cache. Indexer queries: \querystep{A}~projection, \querystep{B}~selecting the index list (the same on every rank), \querystep{C}~attention over the entries the rank holds, \querystep{D}~merging the partial outputs of the ranks (by log-sum-exp).}
\label{fig:parallelism-cp}
\end{figure*}

\textbf{Context parallelism.}
CP splits work by token in two ways~\citep{liu2024ring, yang2025context}, which we describe in the following, and illustrate in Figure~\ref{fig:parallelism-cp}.

Decode context parallelism (DCP), also called KV parallelism~\citep{bhatia2025helix}, splits the KV cache across the ranks, in both prefill and decode, so that each rank stores only $1/N$ of it, which lets a context of millions of tokens fit in HBM, and reads only its share in each decode step.
SGLang deals the tokens out round-robin: token $j$ goes to rank $r_j$, its slot in the KV cache modulo $N$, and that rank holds key $k_j$ and thus computes index key $k_j^{(\mathrm{idx})}$.

Prefill context parallelism (PCP) also splits the query rows of a chunk $\mathcal{C}$, giving rank $r$ the rows $\mathcal{C}_r$, so that it selects the index lists $J_i$ of its own rows $i \in \mathcal{C}_r$ only.
The index query is not split: with PCP, every rank holds in prefill all $n_h$ query heads $q_{i,h}^{(\mathrm{attn})}$ of its rows, and in decode (and with GQA also in prefill), the ranks gather them, as the attention does anyway, so every rank computes $q_i^{(\mathrm{idx})}$ itself.

We extend SGLang with PCP for \glm{}, which leaves us free to choose its layout, for both the base model and \name{}.
We make each $\mathcal{C}_r$ one contiguous span of $\mathcal{C}$, which lets the linear-attention layers pass their recurrent state along it.
\name{} does not depend on this split, but it is a natural fit: as every query attends a fixed budget of keys, a contiguous split already balances the sparse attention across ranks, whereas dense causal attention commonly uses a zigzag or striped split to balance its causal work~\citep{brandon2023striped, yang2025context}.

With MLA, we use both PCP and DCP (Figure~\ref{fig:parallelism-cp}, top), with no TP in the prefill attention, since SGLang was already developing PCP for \glm{}, while it had none for \qwen{}.
This also suits \name{}: PCP splits the prefill by query rows rather than by heads.
\querystep{A}~Every rank thus holds all query heads of its rows and computes their index queries $q_i^{(\mathrm{idx})}$, $i \in \mathcal{C}_r$, alone, without the \coll{AllReduce} of Equation~\ref{eq:par-sum} that TP needs.
For PCP, the attention already gathers the latents $K_{\mathcal{C}}^{(\mathrm{attn})}$ of the whole chunk.
\keystep{1}~Every rank thus projects the index keys $K_{\mathcal{C}}^{(\mathrm{idx})}$ of the whole chunk itself, and \keystep{2}~clusters them into the same $B_{\mathrm{cache}}^{(\mathrm{idx})}$.
The indexer thus adds no communication, as long as each chunk starts on a cluster block, as chunks of $32768 = 64\, n_{\mathrm{cluster}}$ tokens do.
\querystep{B}~Each rank selects $J_i$ for its rows $i \in \mathcal{C}_r$, but DCP splits the KV cache of the prefix across ranks, so most of the selected keys $j \in J_i$ live on other ranks, $r_j \neq r$.
The sparse attention, of the base model as of \name{}, gathers only those rows.
\querystep{C}~Every rank marks the prefix rows that its lists $J_i$, $i \in \mathcal{C}_r$, select in a mask $J_{\mathcal{C}_r}$ of one byte per prefix row, and one \coll{AllReduce} with the maximum gives every rank their union $J_{\mathcal{C}}$ over the chunk, less the chunk itself, whose rows every rank already holds:
\begin{equation}
J_{\mathcal{C}} = \bigcup_{i \in \mathcal{C}} J_i \setminus \mathcal{C}
\label{eq:cp-gather}
\end{equation}
\querystep{D}~Each rank then contributes the latents $K_{J_{\mathcal{C}},r}^{(\mathrm{attn})}$ of the rows of $J_{\mathcal{C}}$ it holds, padded to the largest count over the ranks, to one \coll{AllGather}, which gives every rank all of $J_{\mathcal{C}}$.
Every rank lays them out with the rows of the chunk in one buffer, $K_{J_{\mathcal{C}} \cup \mathcal{C}}^{(\mathrm{attn})}$, and remaps $J_i$, $i \in \mathcal{C}_r$, onto it.
\querystep{E}~The attention kernel then runs unchanged on the remapped lists.
Without the mask, every rank would gather the whole prefix, 1\,KiB per row, or 10\,GiB per layer at a prefix of $T = 10$M tokens, while the mask costs 1\,B per row.
Empirically, in our runs of \glm{} on a 10M-token prompt, the union $J_{\mathcal{C}}$ of the index lists of all 32,768 queries of a late chunk covers only about 30\% of the prefix, as nearby queries select largely the same keys, so the \coll{AllGather} moves, and every rank holds, about 30\% of the prefix rather than all of it, up to the padding.

With GQA, we run attention with TP8 and add DCP8, without PCP: TP splits the prefill by head and computes the keys of the whole chunk on every rank, and DCP splits the KV cache, which is what lets a context of 10M tokens fit on one node.
DCP8 replaces the split of the KV heads: every rank holds all four KV heads of the tokens it owns, $r_j = r$, and gathers the query heads of the other ranks before the indexer and the attention, in prefill and decode.
It thus needs neither the partial sums of Equation~\ref{eq:par-sum} nor a stored $k_j^{(\mathrm{idx})}$: like MLA, it recomputes the index keys of its own tokens from its KV cache.

In decode (Figure~\ref{fig:parallelism-cp}, bottom), each new token $j$ is written to its rank $r_j$, so the keys of a cluster block are spread across all ranks.
\keystep{1}~Once its $n_{\mathrm{cluster}}$ keys are complete, every rank needs all their index keys $k_j^{(\mathrm{idx})}$, but can compute only those with $r_j = r$.
Each rank writes them into an $n_{\mathrm{cluster}} \times d^{(\mathrm{idx})}$ buffer, one row per token of the block in order, with zeros in the rows of the other ranks, and one \coll{AllReduce} sums the buffers, 64\,KiB per block in bf16, or 128\,B per token.
As every row is non-zero on one rank only, the sum gathers the keys of the block in order, where an \coll{AllGather} would group them by rank.
\keystep{2}~Every rank then clusters the same keys into the same buckets of $B_{\mathrm{cache}}^{(\mathrm{idx})}$.
Until then, the keys of the block are covered by the window of the $n_w$ most recent keys, which is selected by position, so they need no index key.
This \coll{AllReduce} runs before the CUDA graph of the decode step is replayed.
In decode the attention splits the query heads across the ranks, by TP with GQA and, with MLA, by a split of the MLA heads that we add to SGLang, so every rank gathers all $n_h$ query heads $q_{i,h}^{(\mathrm{attn})}$ of every request, as the attention does anyway.
\querystep{A}~Every rank thus projects the same index query $q_i^{(\mathrm{idx})}$, \querystep{B}~selects the same $J_i$, and \querystep{C}~attends over the entries of $J_i$ it holds,
\begin{equation}
J_{i,r} = \{\, j \in J_i : r_j = r \,\}
\label{eq:cp-split}
\end{equation}
into a partial output $o_{i,r}$, softmax-normalised over $J_{i,r}$, with $\lambda_{i,r}$ the log-sum-exp of its logits.
\querystep{D}~The ranks merge the partial outputs of each query head by their log-sum-exp~\citep{dao2023flashdecoding}:
\begin{equation}
o_i = \sum_{r} e^{\lambda_{i,r} - \lambda_i}\, o_{i,r}, \qquad \lambda_i = \log \sum_{r} e^{\lambda_{i,r}}
\label{eq:cp-merge}
\end{equation}
As the $J_{i,r}$ partition $J_i$, $o_i$ is the attention over $J_i$, up to floating-point rounding.

\textbf{Memory.}
Replicating $B_{\mathrm{cache}}^{(\mathrm{idx})}$ trades memory for communication.
With $\beta_{\mathrm{KV}}$ and $\beta_B$ the bytes per token of the KV cache and of the bucket cache, the persistent storage of the KV cache and the flat index, per layer and rank, is approximately
\begin{equation}
T \left( \frac{\beta_{\mathrm{KV}}}{N} + \beta_B \right)
\label{eq:cp-memory}
\end{equation}
where $\beta_B \approx 24.5$\,B, and $\beta_{\mathrm{KV}} = 1024$\,B for \glm{} and 4,096\,B for \qwen{}.
At $T = 10$M with DCP8, the bucket cache adds 246\,MiB as measured, $T \beta_B$ plus padding rows, to the $T \beta_{\mathrm{KV}} / N = 1.25$\,GiB share of the KV cache of a layer of \glm{} (+19\%), and 5\% on \qwen{}.
An index per rank over its own keys would avoid the \coll{AllReduce} of the keys, but its buckets and its $J_i$ would differ from those of a single GPU.

\end{document}